\def\year{2022}\relax
\documentclass[letterpaper]{article} 
\usepackage[pdftex]{graphicx}
\usepackage{aaai22}  
\usepackage{times}  
\usepackage{helvet}  
\usepackage{courier}  
\usepackage[hyphens]{url}  
\usepackage{graphicx} 

\usepackage{natbib}  
\usepackage{caption} 
\DeclareCaptionStyle{ruled}{labelfont=normalfont,labelsep=colon,strut=off} 
\usepackage{xspace}
\usepackage{bbding}
\usepackage{pifont}
\usepackage{amssymb}

\usepackage{microtype}
\usepackage{algpseudocode}
\usepackage[linesnumbered,ruled,vlined]{algorithm2e}
\usepackage{booktabs}
\usepackage{rotating}
\usepackage{lipsum}
\usepackage{hhline}
\usepackage{extarrows}
\usepackage{booktabs} 
\usepackage{xspace}
\usepackage{subfig}
\usepackage{wasysym}
\usepackage{graphicx}
\usepackage{amsfonts} %
\usepackage{xcolor}
\usepackage{CJKutf8}
\usepackage{multirow}
\usepackage{multicol}
\usepackage{xspace}
\usepackage{subfig}
\usepackage{amsmath}
\usepackage{float}
\usepackage{hyphenat}
\usepackage{eufrak}
\usepackage{amssymb}
\usepackage{bbm}

\usepackage{newfloat}
\usepackage{listings}
\floatstyle{ruled}
\newfloat{listing}{tb}{lst}{}
\floatname{listing}{Listing}

\title{Hyperbolic Disentangled Representation for Fine-Grained Aspect Extraction}

\author{
    Chang-You Tai\textsuperscript{\rm 1},
    Ming-Yao Li\textsuperscript{\rm 1},
    Lun-Wei Ku\textsuperscript{\rm 1}
}
\affiliations{
    \textsuperscript{\rm 1}Academia Sinica, Taipei, Taiwan\\
    johnnyjana730@gmail.com, 	xuauul@gmail.com, lwku@iis.sinica.edu.tw
}

\newcommand{\system}{HDAE\xspace}

\begin{document}
\vspace{-5.0pc}
\maketitle

\vspace{-5.0pc}

\begin{abstract}




Automatic identification of salient aspects from user reviews is especially
useful for opinion analysis. There has been significant progress in utilizing
weakly supervised approaches, which require only a small set of seed words for
training aspect classifiers. However, there is always room for
improvement. First, no weakly supervised approaches fully utilize
latent hierarchies between words. Second, each seed word's representation
should have different latent semantics and be distinct when it represents a different
aspect. In this paper we propose HDAE, a hyperbolic disentangled
aspect extractor in which a hyperbolic aspect classifier captures
words' latent hierarchies, and an aspect-disentangled representation models the
distinct latent semantics of each seed word. Compared to
previous baselines, \system achieves average F1 performance gains of 18.2\% and
24.1\% on Amazon product review and restaurant review datasets, respectively.
In addition, the embedding visualization experience demonstrates that \system is a
more effective approach to leveraging seed words. An ablation study and a case
study further attest the effectiveness of the proposed components.

\end{abstract}


\vspace{-1.0pc}

\section{Introduction}
 
 
 
Researchers have begun to focus on aspect extraction, the
automatic detection of fine-grained segments with predefined
aspects~\cite{hu2004mining, doi:10.2200/S00416ED1V01Y201204HLT016,
pontiki2016semeval}, due to its potential for
downstream tasks. For example, aspect extraction benefits users and customers when
searching through review segments for aspects of interest on the Internet.
Aspect extraction is also crucial for document
summarization~\cite{angelidis2018summarizing}, recommendation
justification~\cite{ni2019justifying}, and review-based
recommendation~\cite{chin2018anr}.

Aspect extraction research can be divided into 
supervised approaches, unsupervised approaches, and weakly supervised
approaches.\footnote{We touch on unsupervised and supervised approaches in
the related work section.} Among these, many studies have been conducted on weakly supervised
approaches~\cite{DBLP:journals/corr/abs-1909-00415,
angelidis2018summarizing, zhuang2020joint} since they 
allow the model to be trained without substantial human-labeled
data. For example, \citet{angelidis2018summarizing} initialize
fine-grained aspect representations using only a small number of descriptive
keywords, or seed words, to identify highly salient opinions in review
segments. Also, \citet{DBLP:journals/corr/abs-1909-00415} propose a
student-teacher framework that more effectively leverages seed words by using a
bag-of-words classifier teacher.

\begin{figure}[tb]   
\vspace{-0.5 pc}
\centering
\includegraphics[scale=0.24, trim={0 0 0 0}]{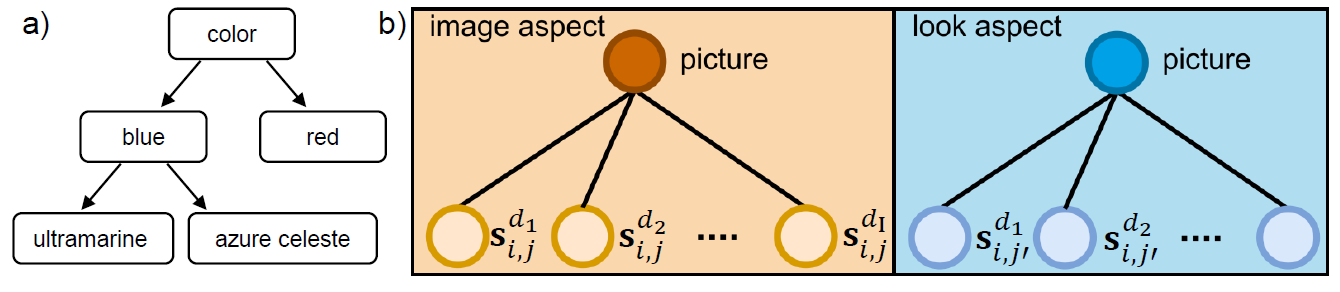}
\vspace{-1.0 pc}
\caption{a)~Seed word \textit{color} and its hypernym pairs. b)~An illustration
of latent semantics under seed word \textit{picture}. For example, in the TV
domain's \textit{image} aspect, \textit{pixel of picture}
$\textbf{s}^{d_1}_{i,j}$ and \textit{screen picture} $\textbf{s}^{d_2}_{i,j}$
exist, wherease in the boot domain's \textit{look} aspect, 
\textit{cute picture} $\textbf{s}^{d_1}_{i,j'}$ and \textit{attractive picture}
$\textbf{s}^{d_2}_{i,j'}$ exist.}
\label{fig:intro}
\vspace{-1.5 pc}
\end{figure}


However, there is room for improvement in such seed word based methods. First, they neglect to consider the latent hierarchies between words,
and it is assumed that capturing latent hierarchies between words will further
improve seed word based methods on aspect inference, for instance by better
identifying and organizing seed words and their hypernym
pairs~\cite{DBLP:journals/corr/abs-2010-01898, DBLP:journals/corr/abs-1906-02505}. For
example, as shown in Fig.~\ref{fig:intro}(a), the general seed word
\textit{color} near the top can be used to find the more specific words
\textit{blue} or \textit{green} in the middle, after which even more
specific words can be found such as \textit{ultramarine} or \textit{azure celeste}. If seed
words or their hypernym pairs exist in one review segment, the model can infer
that it is of the corresponding aspect.

To allow the model to fully capture latent hierarchies between words, we introduce hyperbolic
space~\cite{nickel2017poincar,murty2018hierarchical,xu2018neural,
DBLP:journals/corr/abs-1906-02505,lopez-strube-2020-fully}.  Compared to
Euclidean space, hyperbolic space effectively encodes hierarchical structure
information~\cite{nickel2017poincar}, the latent hierarchies between words in
this paper. In particular, when embedding tree-like structures, compared to the
volume in Euclidean space, which leads to high distortion
embeddings~\cite{DBLP:journals/corr/abs-1804-03329,sarkar2011low}, volume
in hyperbolic space grows exponentially and can embed trees with arbitrarily
low distortion~\cite{sarkar2011low, nickel2017poincar}. By virtue of such a
hierarchy, a seed
word based model can better identify and utilize seed words and their hypernym
words and thus achieve better aspect inference in hyperbolic space.

Second, existing seed word-based approaches model each seed word representation
in a uniform manner while neglecting the fact that each seed word should have different
latent semantics when conducting aspect extraction. For example, for the Amazon
product review dataset~\cite{angelidis2018summarizing}, in the TV
domain's \textit{picture} aspect, the latent semantics under the seed word
\textit{picture} can be \textit{pixel of the picture}, \textit{screen
picture}, or \textit{HD picture}, as shown in Fig.~\ref{fig:intro}(b). It is
essential to select the most relevant latent semantics of the seed word when using the
seed word \textit{picture} to infer review aspects of segments. Furthermore,
as shown in Fig.~\ref{fig:intro}(b), the latent semantics of the seed word should be
different in different aspects: this is also neglected by the
current uniform representation.  Such a uniform approach to modeling seed words tends
to result in sub-optimal representations.

Thus, we propose \system, a hyperbolic disentangled aspect extractor
which captures words' latent hierarchies and  disentangles the latent semantics of each seed word. First, we propose a hyperbolic aspect
classifier, using a hyperbolic distance function to calculate the relationship
between
the segment vector and the aspect representation generated from the seed word. Second, we
introduce an aspect disentanglement module to model each seed word's latent
semantics and then generate an aspect-refined representation of each review segment
by selecting the most relevant latent semantics. In addition, we propose
aspect-aware regularization to model each latent semantic meaning under its aspect
scope while encouraging the independence of different latent semantic meanings. We conduct
experiments on two datasets, demonstrating that \system achieves better
aspect inference, which is further substantiated by embedding
visualizations. We also provide two case studies to investigate \system's aspect inference ability compared with
baselines without fully capturing words' latent hierarchies and the
interpretability of the seed words’ disentangled latent semantics.


We summarize our contributions: first, we propose a novel hyperbolic
disentangled aspect extractor. To the best of our knowledge, this is the first
work to investigate how to leverage hyperbolic components and disentangled
representations for weakly supervised
approaches to aspect extraction.\footnote{The codes is at https://github.com/johnnyjana730/HDAE/} Second, we propose a hyperbolic aspect classifier
which captures word's latent hierarchies and generates associations between the
review segment and aspects of interest. Third, we introduce the aspect
disentanglement module and aspect-aware latent semantic regularization to model
the latent semantic meaning of each seed word. Experiments and a case study
demonstrate the effect of the proposed methods for aspect extraction.





\vspace{-0.5pc}

\section{Related Work} 





\textbf{Aspect Extraction} In addition to weakly supervised approaches,
there are also supervised approaches and unsupervised approaches. 
Supervised neural networks achieve better performance than traditional rule-based 
approaches by viewing aspect extraction as a sequence
labeling problem which can be tackled with hidden Markov
models~\cite{jin2009opinionminer}, conditional random
fields~\cite{yang2012extracting, mitchell2013open}, or
recurrent neural networks~\cite{wang2016recursive, liu2015fine}.
However, supervised approaches require large amounts of labeled data for
training. Unsupervised approaches, in contrast, do not use annotated data. Early
examples are latent Dirichlet allocation (LDA)-based
methods~\cite{chen2014aspect, garcia2018w2vlda,
shi2018short}. Recently, neural network-based methods~\cite{iyyer2016feuding,
srivastava2017autoencoding, he2017unsupervised, luo2019unsupervised,
shi2020simple} have shown remarkable performance and have outperformed LDA-based
methods. However, unsupervised approaches are not effective when used
directly for aspect extraction~\cite{DBLP:journals/corr/abs-1909-00415,
angelidis2018summarizing, DBLP:journals/corr/abs-2004-13580}. For example,
many-to-one mapping or high-resolution selective mapping is required by
\citet{he2017unsupervised} and \citet{shi2020simple} to manually associate the
model-inferred aspect with gold-standard aspects.

\textbf{Hyperbolic representations} have been used to model complex
networks~\cite{Krioukov_2010, nickel2017poincar,
DBLP:journals/corr/abs-1806-03417, DBLP:journals/corr/TayLH17a,
DBLP:journals/corr/abs-1805-09786,
DBLP:journals/corr/abs-2010-01898} and have proven more suitable than
Euclidean space in representing hierarchical 
data~\cite{sala2018representation, nickel2017poincar}. For example, 
\citet{lopez-strube-2020-fully} introduce hyperbolic representations to capture
latent hierarchies arising from the class distribution for multi-class
multi-label classification. \citet{aly2019every} use Poincar\'{e}
embeddings to improve existing methods for domain-specific taxonomy induction.
\citet{le-etal-2019-inferring} propose utilizing hyperbolic representations to
infer missing hypernymy relations. \citet{HGCFtranhyperml} show that points
in hyperbolic space can be more concentrated while maintaining the desired
separation and revealing nuanced differences. To our knowledge, this is
the first work to apply
hyperbolic representations to weakly supervised approaches for aspect
extraction.




\textbf{Disentangled representations} improve model performance by
identifying and disentangling latent explanatory factors in
the observed data~\cite{DBLP:journals/corr/abs-1206-5538} and have shown their
success in the NLP domain~\cite{shen2017style, zhao2018adversarially,
chen2019variational, DBLP:journals/corr/HuYLSX17}. For instance,
\citet{DBLP:journals/corr/HuYLSX17} propose disentangled representations with
designated semantic structure, which generates sentences with dynamically
specified attributes. \citet{osti_10220295} derive disentangled representations
which separate the distinct and informative factors of variations to improve
content-based detection. Disentangled representation has been
successively applied to the recommendation~\cite{DBLP:journals/corr/abs-1910-14238,
ma2019disentangled, hu-etal-2020-graph} and
computer vision~\cite{Liu_Wang_Wu_Xiao_2020,
dupont2018learning} domains. For
example, \citet{wang2020disentangled} model diverse relationships
and disentangle user intents to achieve better-performing representations. To
our knowledge, this is the first work to apply disentangled representations to
weakly supervised approaches for aspect extraction.

\vspace{-1.0 pc}

\section{Preliminaries}




\noindent \textbf{Problem formulation} The goal of aspect extraction is to
predict an aspect category $a_i \in A_\mathrm{C} = \{a_j\}^{\mathrm{K}}_{j=1}$
given a review segment (e.g., sentence, clause) $x^s =
\{x_1,x_2,...,x_\mathrm{T}\}$ from a specific domain $d_\mathrm{C}$ (e.g.,
laptop bags, TVs), where the review segments are created by splitting each
review in the corpus;  $x_i$ is the word index in the segment; $a_i$ is an
aspect and $A_\mathrm{C}$ refers to the aspect set pertaining to 
domain~$d_\mathrm{C}$; $\mathrm{K}$ is 
the number of total aspects   
and $\mathrm{T}$ is the segment's length. For every aspect $a_i \in A_\mathrm{C}$,
a small number of seed words $[s_{i,1},s_{i,2},...,s_{i,N}]$ are provided
during training. The classifier predicts $\mathrm{K}$ aspect
probabilities $\mathrm{p}^{a}_s = \langle
\mathrm{p}^{a_1}_s,...,\mathrm{p}^{a_\mathrm{K}}_s\rangle$ given a test review
segment $x^s$ and the seed words.


\noindent \textbf{Hyperbolic Geometry}
We introduce two hyperbolic models:\footnote{For more details; see
\citet{robbin2011introduction}.} the Poincar\'{e} ball model and the Klein model.

\noindent \textbf{The Poincar\'{e} ball model} is defined as a Riemannian manifold
$\mathcal{P}^{n}$ = ($\beta$ ,$g^{\beta}_{\mathrm{\textbf{x}}}$), where
$\beta^{n} = \{\mathrm{\textbf{x}} \in \mathbb{R}^{n}: ||\mathrm{\textbf{x}}||
< 1\}$ is an open unit ball, with the metric tensor
$g^{\beta}_{\mathrm{\textbf{x}}} = \lambda^{2}_{\mathrm{\textbf{x}}}g^{E}$,
where $\lambda_{\mathrm{\textbf{x}}} = \frac{2}{1-||\mathrm{\textbf{x}}||^2}$;
$g^{E}$ is the Euclidean metric tensor. The distance on the manifold is defined as
\vspace{-0.2 pc}
\begin{align}
	d_\mathcal{P}(\mathrm{\textbf{x}},\mathrm{\textbf{y}}) = \mathrm{arcosh}
	\begin{pmatrix}  1 + 2
	\frac{||\mathrm{\textbf{x}}-\mathrm{\textbf{y}}||^{2}}{(1-||\mathrm{\textbf{x}}||^{2})(1-||\mathrm{\textbf{y}}||^{2})}
	\end{pmatrix}.
\end{align}




\noindent \textbf{The Klein model} is given by $\mathcal{K}^{n} = \{\textbf{x} \in
\mathbb{R}^{n} : ||\textbf{x}|| < 1 \}$ and is often used for aggregation since
the Einstein midpoint~\cite{DBLP:journals/corr/abs-1805-09786} can be easily
computed in the Klein model. Formally, a point in the Klein model can be obtained
from Poincar\'{e} coordinates by
\vspace{-0.2 pc}
\begin{align}
   \mathcal{P}^{n} \to \mathcal{K}^{n} : \pi_{\mathcal{P}\to\mathcal{K}}(\textbf{x}_\mathcal{K}) = \frac{2\textbf{x}_\mathcal{P}}{1 + ||\textbf{x}_\mathcal{P}||^{2}}
\end{align}
and the backward transition formulas
\vspace{-0.2 pc}
\begin{align}
   \mathcal{K}^{n} \to \mathcal{P}^{n} : \pi_{\mathcal{K}\to\mathcal{P}}(\textbf{x}_\mathcal{P}) = \frac{\textbf{x}_\mathcal{K}}{1 + \sqrt{1 - ||\textbf{x}_\mathcal{K}||^{2}}}.
\end{align}

\vspace{-0.2 pc}


For the Poincar\'{e} ball model, the exponential map, from tangent space to
hyperboloid manifold, $\mathrm{exp}_{\textbf{x}}$:
$\mathcal{T}_{\textbf{x}}\mathcal{P}$ $\to$ $\mathcal{P}$, and the logarithmic map,
from hyperboloid manifold to tangent space, $\mathrm{log}_{\textbf{x}}$:
$\mathcal{P} \to \mathcal{T}_{\textbf{x}}\mathcal{P}$, can be found 
in~\citet{DBLP:journals/corr/abs-1910-12892}. For simplicity, we denote
$d^{\mathrm{exp}}_\mathcal{P}$ as the hyperbolic distance of the tangent space vector
after applying the exponential map:
%
\begin{align}
   d^{\mathrm{exp}}_\mathcal{P}(\mathrm{\textbf{x}},\mathrm{\textbf{y}}) =  d_\mathcal{P}({\mathrm{exp}_{0}(\textbf{x}}),\mathrm{exp}_{0}(\mathrm{\textbf{y}})).
\end{align}

\vspace{-0.5 pc}

\section{Methodology}

\subsection{Euclidean Aspect Extractor}


Our work builds on the seed word based model developed
by~\citeauthor{angelidis2018summarizing}. We describe the method,
including segment representation generation and the aspect classifier.


\textbf{Segment Representation} For each review segment $x^s =
\{x_1,x_2,...,x_\mathrm{T}\}$, the segment representation $\textbf{v}_s$ is
generated by a weighted sum of an individual word:
%
%
\begin{align}
\label{eqn:segment_repr}
   \textbf{v}_s = \sum^{n}_{i=1} & c_{i}\textbf{v}_{x_i} \\ 
   c_i = \frac{\exp(u_i)}{\sum^{n}_{j=1} \exp(u_j)} &;  u_i = \textbf{v}^{\top}_{x_i} \cdot \textbf{M} \cdot \textbf{v}'_s,
\end{align}
where $\textbf{v}_{x_i}$ is the vector of the $i$-th word $x_i$;
$\textbf{v}'_s$ is average of the segment's word vector; and \textbf{M} $\in
\mathbb{R}^{d \times d}$ denotes the attention matrix.

\textbf{Euclidean Aspect Classifier}  To predict a probability
distribution over $\mathrm{K}$ aspects, the vector~$\textbf{v}_s$ is fed to
a hidden classification layer followed by the softmax function:
%
\begin{equation}
\label{eqn:euclideanaspectclassifier}
   \mathrm{p}^{a}_s = \mathrm{softmax}(\textbf{W}\textbf{v}_s + \textbf{b}),
\end{equation}
where $\textbf{W}$ and $\textbf{b}$ are trainable parameters. To focus on the
aspect of interest, for each aspect
$a_i$, which has seed words $[s_{i,1},s_{i,2},...,s_{i,N}]$, the model 
generates the aspect vector $\textbf{a}_i$ by using the labeled aspect seed
words:
%
\begin{align}
\label{eqn:aspect_calculated}
   \textbf{a}_i = \sum^{N}_{j} \textbf{z}_{i,j}\textbf{s}_{i,j}
   ; \textbf{A} = [\textbf{a}^{\top}_i;...;\textbf{a}^{\top}_\mathrm{K}],
\end{align}
%
where \textbf{A} $\in \mathbb{R}^{\mathrm{K} \times d}$ denotes the aspect
matrix; and $\textbf{s}_{i,j}$ denotes the $j$-th seed word representation; 
  the weight vectors $\textbf{z}_{i,j}$  
are determined by the method mentioned in
\citet{angelidis2018summarizing}; and $N$ is the number of seed words.
Then, the segment reconstructed vector $\textbf{r}_s$ is generated based on the
aspect vector:
\vspace{-1.0 pc}
\begin{align}
  \textbf{r}_s &= \textbf{A}^{\top}\mathrm{p}^{\mathrm{asp}}_s.
\end{align}

To optimize the performance, the model is trained by reconstruction loss, which maximizes
the distance between inner product  $\textbf{r}_s\textbf{v}_s$ and
$\textbf{r}_s\textbf{v}_{n}$, where $\textbf{v}_{n_i}$ is the vector of a
randomly sampled negative segment.
%
\begin{align}
  \label{eqn:reconstruction_loss}
  J_{r}(\theta) = \sum_{x^s \in C}&\sum^{k_n}_{i=1}\mathrm{max}(0,1- \textbf{r}_s\textbf{v}_s +  \textbf{r}_s\textbf{v}_{n_i}),
\end{align}




\subsection{Hyperbolic Disentangled Aspect Extractor}

Here, we present \system, 
a hyperbolic aspect classifier with 
an aspect disentanglement module proposed to
model multiple latent semantic meanings for each seed word according to its aspect
category.





\subsubsection{Hyperbolic Aspect Classifier}
To infer the review segment's aspect probability $\mathrm{p}^{\mathrm{a}_i}_s$ in
hyperbolic space, we follow \citet{DBLP:journals/corr/abs-1905-09791} in using the
hyperbolic distance function and biases to calculate the relationship between
segment vector $\textbf{v}_s$ and aspect representation $\textbf{a}_i$ as

\vspace{-1.0 pc}

\begin{align}
    \label{eqn:aspectinferringhyper}
   \mathrm{p}^{a_i}_s = -d^{\mathrm{exp}}_\mathcal{P}(\textbf{v}_{s}, \textbf{a}_{i})^{2} + b_{v} + b_{a_{i}}.
\end{align}


Then, to generate the reconstructed embedding $\textbf{r}_s$, the Einstein midpoint
is used to aggregate hyperbolic aspect weights, with a simple form in the
Klein disk model:

\vspace{-0.6 pc}

\begin{align}
\label{eqn:reconstruct_repr}
    \textbf{r}_s = \mathrm{log}_{0}&(\pi_{\mathcal{K}\to\mathcal{P}}(\sum_{a_{i}\in A_\mathrm{C}} \frac{k_i\gamma(\textbf{a}^{\mathcal{K}}_i)}{\sum_{j}k_{j}\gamma(\textbf{a}^{\mathcal{K}}_{j})}
    \textbf{a}^{\mathcal{K}}_i)) \\
    k_i &= \mathrm{exp}(\beta \mathrm{p}^{a_i}_s - c),
\end{align}

\noindent where $\textbf{a}^{\mathcal{K}}_i$ =
$\pi_{\mathcal{P}\to\mathcal{K}}(\textbf{a}^{\mathcal{P}}_i)$;
$\textbf{a}^{\mathcal{P}}_i$ denotes the Poincar\'{e} aspect embedding;
$\textbf{a}^{\mathcal{P}}_i =
\mathrm{exp}_{0}(\textbf{a}_i)$; $\beta$ and $c$ are set parameters; 
and Lorentz factors $\gamma(t) = \frac{1}{(1-||t||^{2})^{1/2}}$.




\subsubsection{Aspect Disentanglement Module}


To generate multiple latent semantic meanings for each seed word, we 
propose a disentangled semantic representation. Then, we present aspect-aware
regularization, which models latent semantic vectors for each seed word, after
which we discuss refined seed word representation.

  


\noindent \textbf{Disentangled Semantic Representation}
For aspect~$a_i$, we devise a representation function to output a disentangled semantic
vector $\textbf{s}^{d}_{i,j}$ for the $j$-th seed word $s_{i,j}$,
which is composed of $\mathrm{I}$ independent components:

\vspace{-0.5 pc}
\begin{equation}
    \textbf{s}^{d}_{i,j} = (\textbf{s}^{d_1}_{i,j}, \textbf{s}^{d_2}_{i,j},
    \textbf{s}^{d_3}_{i,j},
    ...,\textbf{s}^{d_\mathrm{I}}_{i,j}),
\end{equation}

\noindent where disentangled semantic vector $\textbf{s}^{d_k}_{i,j}$ is generated by
adding a standard Gaussian random variable to the original seed word
representation $\textbf{s}_{i,j}$.

\begin{figure}[htb]
\vspace{-0.5 pc}
\centering
\includegraphics[scale=0.21, trim={10 0 0 0}]{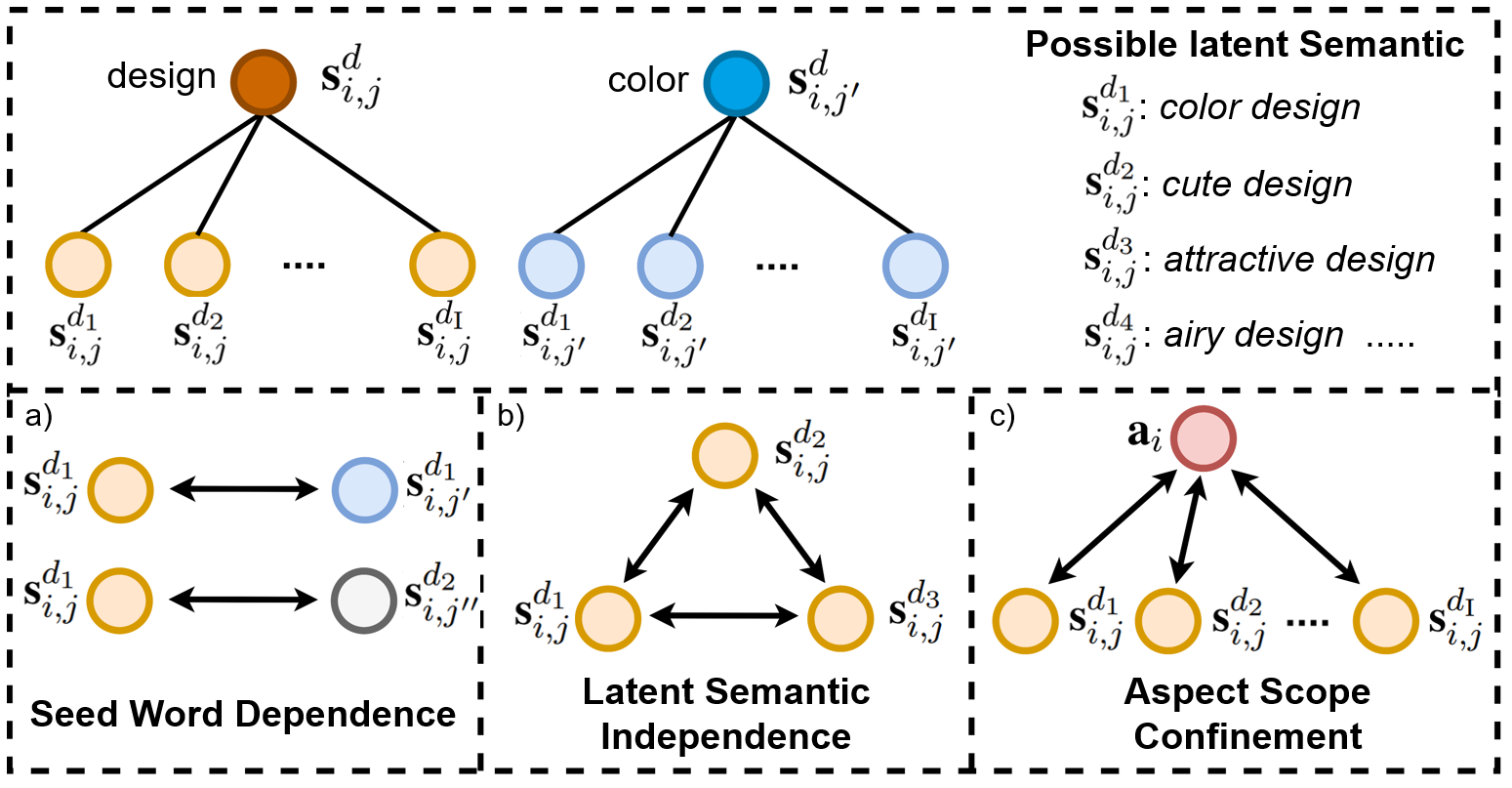}
\vspace{-1.5 pc}
\caption{The proposed aspect disentanglement module generates 
disentangled semantic representations for each seed word and models latent 
semantics using (a)~seed word dependence, (b)~latent semantic independence, 
and (c)~aspect scope confinement.}
\label{fig:distance}
\end{figure}

\noindent \textbf{Aspect-Aware Regularization} This models the latent
semantic representation of each seed word according to its aspect category and has three
objectives, as shown in Fig.~\ref{fig:distance}: (a)~seed word
dependence, (b)~latent semantic independence, and (c)~aspect scope confinement,
which are controlled by latent semantic modeling distances $d_1, d_2,$ and 
$d_3$.  


\noindent \textbf{Seed Word Dependence} The interdependence between seed word
pairs sheds light on the modeling of the seed word's latent semantics within the
scope of its aspect. For example, for seed word  \textit{design} in the
boot domain's \textit{look} aspect, the latent semantic meaning,
which facilitates fine-grained aspect inference, can be \textit{color design},
\textit{design style}, \textit{cute design}, and \textit{attractive design}.
The desired latent semantic meaning can be modeled by narrowing the gap between either
the latent semantic meaning of \textit{design} and the latent semantic meanings of other seed words,
such as \textit{color}, \textit{style}, \textit{cute}, and \textit{attractive}
in the same \textit{look} aspect. Likewise, in the TV domain's \textit{service}
aspect, latent semantic meanings \textit{shipping service}, \textit{replacement
service}, and \textit{delivery service} can be generated by minimizing the distance
between either the latent semantic meaning of \textit{service} and that of
\textit{shipping}, \textit{replacement}, and \textit{delivery}, which are seed
words in the same aspect.

To model the interdependence of seed word pairs, we use the 
hyperbolic distance function $d_\mathcal{P}()$ to achieve fine-grained
relationship modeling, since hyperbolic space offers the ability to not
only preserve hierarchical (tree-like)
information~\cite{nickel2017poincar,zhang2020hype,
DBLP:journals/corr/abs-1805-09786,DBLP:journals/corr/abs-1910-12933} but also
nuanced differences (to better group them)~\cite{HGCFtranhyperml,
tai2021knowledge} and outperforms Euclidean counterparts in various kinds of
data~\cite{zhang2020hype,
DBLP:journals/corr/abs-1805-09786,DBLP:journals/corr/abs-1910-12933,DBLP:journals/corr/abs-2005-00545,HGCFtranhyperml,
tai2021knowledge}. Thus, it is assumed that with more space (hyperbolic space)
to organize points, the model can divide disentangled representations and
better group them. Given seed word pairs such as
\textit{design} $s_{i,j}$ and \textit{color} $s_{i,j'}$ in the specific
aspect, we require at least one latent semantic pair distance to be
close enough:

\vspace{-1.5 pc}
\begin{equation}
\begin{split}
     \mathrm{sim}(s_{i,j},s_{i,j'}) = \mathrm{argmin}\{ \\ d^{\mathrm{exp}}_\mathcal{P}(\textbf{s}^{d_{k}}_{i,j},\textbf{s}^{d_{k'}}_{i,j'})|\textbf{s}^{d_{k}}_{i,j} \in \textbf{s}^{d}_{i,j}, \textbf{s}^{d_{k'}}_{i,j'} \in \textbf{s}^{d}_{i,j'}\} 
\end{split}
\end{equation}

\vspace{-0.6 pc}
\begin{align}
\begin{split}
  J_{d_1}(\theta) = \sum_{a_i \in A_\mathrm{C}}&\sum^{N}_{j=1}\sum^{N}_{j'=j+1}\mathrm{max}(0,
  \\
  \label{eqn:inter_dependence}
  \mathrm{sim}(s_{i,j},s_{i,j'}) - d_1),
\end{split}
\end{align}

\vspace{-0.2pc}

\noindent where $\mathrm{sim}$ outputs the minimal distance from all possible seed
word latent semantic meaning pairs; $d_1$ is the inter seed word alignment distance,
which maintains two latent semantic meanings within a certain distance. Intuitively, for
different aspect word pairs, the alignment score should be different, as
in~\citet{wang2020disentangled}.  For example, in
the boot domain's \textit{look} aspect, the seed word dependence between
\textit{design} and  \textit{color} should be more significant than
\textit{design} and \textit{going}. We leave this to future work.


\noindent \textbf{Latent Semantic Independence} Latent semantic meanings should be
distinct from each other. 
Independent latent semantic meanings reduce redundancy and confusion in
aspect inference. To achieve this, we maintain the distance between the seed word's
latent semantic meanings.

\vspace{-0.5 pc}

\begin{align}
\begin{split}
  J_{d_2}(\theta)=\sum_{a_i \in A_\mathrm{C}}\sum^{N}_{j}\sum^{\mathrm{I}}_{k=1}\sum^{\mathrm{I}}_{k' =k+1}\mathrm{max} \\ (0, 
  d_2 -d^{\mathrm{exp}}_\mathcal{P}(\textbf{s}^{d_{k}}_{i,j},\textbf{s}^{d_{k'}}_{i,j})),
\end{split}
\end{align}


\noindent where $d_2$ is the latent semantic distance. 

\noindent \textbf{Aspect Scope  Confinement} For each seed word, all latent
semantic meanings should be limited in terms of aspect scope. For example, in the
boot domain's \textit{color} aspect, all latent semantic meanings of seed
word \textit{style} should refer to \textit{color's style}. However, in
the \textit{look} aspect, all latent semantic meanings of the same seed word \textit{style}
should refer to \textit{outlook style}. To thus limit all
latent meanings of a seed word to its aspect scope, we introduce another
regularization:

\vspace{-0.5 pc}

\begin{align}
  \label{eqn:aspect_scope}
  J_{d_3}(\theta) = \sum_{a_i \in A_\mathrm{C}}\sum^{N}_{j}\sum^{\mathrm{I}}_{k=1}\mathrm{max}(0,d^{\mathrm{exp}}_\mathcal{P}(\textbf{s}^{d_k}_{i,j},\textbf{a}_i)-d_3),
\end{align}



\noindent where $d_3$ is the aspect scope confinement distance and $\textbf{a}_i$ is
the aspect representation from Eq.~\ref{eqn:aspect_calculated}.
Note compared to seed word dependence and Eq.~\ref{eqn:inter_dependence},
which focuses on dependence between seed word pairs, 
Eq.~\ref{eqn:aspect_scope} ensures  all latent semantic meanings are
modeled within the specific aspect scope.


\noindent \textbf{Refined Seed Word Representation} This constructs refined seed
representations based on its latent semantics. For each seed word, the latent
semantics should be independent from each other; only one latent
semantic meaning should be used to find the aspect relevant content. For example, for the
boot domain's \textit{look} aspect, possible latent semantics of
seed word  \textit{style} include \textit{cute style}, \textit{casual
style}, or \textit{attractive style}; as we can see these latent semantics are
of different meanings, and combining them together may led to a sub-optimal
seed word representation.
Also, according to each review
segment, the most relevant latent semantic meaning should be selected when predicting
its aspect distribution. Thus, we introduce
the Gumbel softmax~\cite{jang2017categorical}, a differentiable softmax function
for generating discrete variables, and use this to generate the desired refined
seed word representation $\textbf{s}^{r}_{i,j}$ according to each segment
$\textbf{v}_s$:

\vspace{-1.0 pc}

\begin{equation}
\label{eqn:reconstruct_refinded}
     \textbf{s}^{r}_{i,j} = \sum^{\mathrm{I}}_{k=1} g_k \textbf{s}^{d_k}_{i,j}, \; g_k = \frac{c_k}{\sum_{k'}c_{k'}},
\end{equation}


\begin{equation}
      c_k =\mathrm{exp}(\frac{-d_{_\mathcal{P}}(\textbf{v}_{s},\textbf{s}^{d_k}_{i,j})}{\tau})
\end{equation}


\noindent where $\tau$, the temperature parameter, controls the extent to which the
output becomes a one-hot vector. With the refined seed word representation
$\textbf{s}^{r}_{i,j}$, the aspect representation can be generated by
Eq.~\ref{eqn:aspect_calculated}.











\newcommand\mycommfont[1]{\footnotesize\ttfamily\textcolor{blue}{#1}}
\SetKwInput{KwInput}{Input}                
\SetKwInput{KwOutput}{Output}              
\vspace{-0.5 pc}
\begin{algorithm}[h]
\small
\DontPrintSemicolon
\KwInput{review segments $S = \{s~|~s\in \mathrm{C} \}$, aspect seed words}
\SetKwFunction{FMain}{ }
\SetKwProg{Fn}{Begin}{}{}
  \textbf{Initialize} \system parameter with pre-trained word vector\\
  \ForEach{epoch}{
    \For{$x^s \in S$}{
      Generate segment embedding $\textbf{v}_s$  \hspace*{\fill}(Eq~~ \ref{eqn:segment_repr}) \\
      \For{$i\leftarrow 1$ \KwTo $\mathrm{K}$}{
        Generate refined aspect seed word vector $\textbf{s}^r_{i, j}$  \hspace*{\fill}(Eq~\ref{eqn:reconstruct_refinded}) \\
        Calculate aspect embedding $\textbf{a}_{i}$  \hspace*{\fill}(Eq~~ \ref{eqn:aspect_calculated}) \\
        Generate aspect probability $\mathrm{p}^{a_i}_s$  \hspace*{\fill}(Eq~\ref{eqn:aspectinferringhyper}) \\
      }
      Generate reconstructed embedding $\textbf{r}_s$  \hspace*{\fill}(Eq~\ref{eqn:reconstruct_repr}) \\
      Calculate objective $J$  \hspace*{\fill}(Eq~\ref{eqn:all_objective}) \\
    }
    Update parameters by Adam optimizer
  }
\caption{\system Learning}
\label{alg:aspec_training}
\vspace{-0.5 pc}

\end{algorithm}

\subsection{Learning Algorithm}

%
The formal description of the above aspect inference process is presented in
Algorithm~\ref{alg:aspec_training}. To train
\system, we rely on the previously introduced reconstruction loss $J_{r}$
(Eq.~\ref{eqn:reconstruction_loss}). Since the reconstruction objective only
provides a weak training signal~\citep{angelidis2018summarizing}, the
distillation objective $J_{d}$ from
the teacher~\citep{DBLP:journals/corr/abs-1909-00415} is used to provide an
additional training signal. Also, the disentangled modeling objectives
$J_{d_1}$, $J_{d_2}$, and $J_{d_3}$ are used to model each latent semantic meaning
according to its aspect category. Thus, the overall objective is

\vspace{-1.2 pc}
\begin{equation}
\label{eqn:all_objective}
       J(\theta) = J_{r}(\theta) + \lambda J_{d}(\theta) + J_{d_1}(\theta) + J_{d_2}(\theta) + J_{d_3}(\theta).
\end{equation}

\noindent The $\lambda$ controls the influence of the distillation objective loss.

\vspace{-0.5 pc}

\section{Experiments and Results}



\noindent \textbf{Datasets} We used Amazon product reviews
from the \textsc{OpoSum} dataset~\citep{angelidis2018summarizing} and restaurant
reviews from the SemEval-2016 Aspect-based Sentiment Analysis
task~\citep{pontiki2016semeval}. The Amazon product reviews cover six
domains, ranging from laptop bags (Bags), bluetooth headsets (B/T), boots,
keyboards (KBs), and televisions (TVs) to vacuums (VCs). The restaurant reviews
dataset covers six languages: English (En), Spanish (Sp),
French (Fr), Russian (Ru), Dutch (Du), and Turkish (Tur). During training, seed
words are provided but not segment aspect labels. Details are provided in the appendix.



\noindent \textbf{Baseline} \textbf{LDA-Anchors}~\cite{lund2017tandem}, an interactive topic model
which utilizes seed words as ``anchors'' to identify the segment
aspect. \textbf{ABAE}~\cite{he2017neural}, an unsupervised
method which adopts reconstruction loss to make the reconstructed embedding
similar to a segment vector. This requires a manual mapping between the 
model-inferred aspect and gold-standard aspects. \textbf{SSCL}~\cite{shi2020simple},
an unsupervised method that uses a contrastive learning algorithm and
knowledge distillation for aspect inference. For manual mapping, the
high-resolution selective mapping (HRSMap) is
used. \textbf{MATE*}~\cite{angelidis2018summarizing}, a
seed-based weakly supervised method which generates predefined aspect
representations by seed word vector. This can be trained by an extra multitask
training objective (MT).\footnote{\noindent MT cannot be applied in restaurant reviews
since it requires datasets from different domains but the same
language.} \textbf{TS-*}~\cite{DBLP:journals/corr/abs-1909-00415}, a seed based
weakly supervised method which adopts a teacher-student iterative co-training
framework, where the teacher (TS-Teacher) is a bag-of-words classifier based on
seed words and the student uses the attention-weighted average of word2vec
embeddings (TS-ATT). \textbf{Gold-*}, supervised models trained using
ground truth aspect labels, only available for restaurant reviews, and not
directly comparable with other weakly supervised
baselines~\cite{DBLP:journals/corr/abs-1909-00415}.

Note that for SSCL and TS, the BERT model also can be used as the
encoder (SSCL-BT, TS-BT). The results of the compared models are
obtained from the corresponding published papers. We also report our
re-implemented version of SSCL-BT*. We do not provide the ABAE and SSCL results
for restaurant reviews for non-English datasets, since this requires
domain knowledge for manual aspect mapping.\footnote{\noindent We report ABAE and SSCL
results for EN restaurant reviews in the appendix in our arxiv version.}

\noindent \textbf{Implementation Details} For \system and other models, detailed hyper-parameter settings are given in the appendix.


\begin{table}[t]
\small
\centering
\scalebox{0.9}{
\begin{tabular}{ccccccc}
\hline
\multicolumn{1}{|c|}{\multirow{1}{*}{}} 
& \multicolumn{6}{c|}{Product review domain}
\\
\multicolumn{1}{|l|}{\multirow{1}{*}{Model}} &
\multicolumn{1}{c}{Bags} &
\multicolumn{1}{c}{KBs} &
\multicolumn{1}{c}{Boots} &
\multicolumn{1}{c}{B/T} &
\multicolumn{1}{c}{TVs} &
\multicolumn{1}{c|}{VCs} \\
\hline \multicolumn{1}{|l|}{{LDA-Anchors}} & \multicolumn{1}{c}{{33.5}} &
\multicolumn{1}{c}{{34.7}} &
\multicolumn{1}{c}{{31.7}} &
\multicolumn{1}{c}{{38.4}} &
\multicolumn{1}{c}{{29.8}} &
\multicolumn{1}{c|}{{30.1}} \\
\multicolumn{1}{|l|}{{ABAE}} & 
\multicolumn{1}{c}{{38.1}} &
\multicolumn{1}{c}{{38.6}} &
\multicolumn{1}{c}{{35.2}} &
\multicolumn{1}{c}{{37.6}} &
\multicolumn{1}{c}{{39.5}} &
\multicolumn{1}{c|}{{38.1}} \\
\multicolumn{1}{|l|}{{SSCL}} & 
\multicolumn{1}{c}{{61.0}} &
\multicolumn{1}{c}{{60.6}} &
\multicolumn{1}{c}{{57.3}} &
\multicolumn{1}{c}{{65.2}} &
\multicolumn{1}{c}{{64.6}} &
\multicolumn{1}{c|}{{57.2}} \\
\multicolumn{1}{|l|}{{SSCL-BT}} & 
\multicolumn{1}{c}{{65.5}} &
\multicolumn{1}{c}{{62.3}} &
\multicolumn{1}{c}{{60.4}} &
\multicolumn{1}{c}{{69.5}} &
\multicolumn{1}{c}{{67.0}} &
\multicolumn{1}{c|}{{61.0}} \\
\multicolumn{1}{|l|}{{SSCL-BT*}} & 
\multicolumn{1}{c}{{56.5}} &
\multicolumn{1}{c}{{61.7}} &
\multicolumn{1}{c}{{41.5}} &
\multicolumn{1}{c}{{51.4}} &
\multicolumn{1}{c}{{58.2}} &
\multicolumn{1}{c|}{{52.4}} \\
\hline \multicolumn{1}{|l|}{{MATE}} & 
\multicolumn{1}{c}{{46.2}} &
\multicolumn{1}{c}{{43.5}} &
\multicolumn{1}{c}{{45.6}} &
\multicolumn{1}{c}{{52.2}} &
\multicolumn{1}{c}{{48.8}} &
\multicolumn{1}{c|}{{42.3}} \\
\multicolumn{1}{|l|}{{MATE-MT}} & 
\multicolumn{1}{c}{{48.6}} &
\multicolumn{1}{c}{{45.3}} &
\multicolumn{1}{c}{{46.4}} &
\multicolumn{1}{c}{{54.5}} &
\multicolumn{1}{c}{{51.8}} &
\multicolumn{1}{c|}{{47.7}} \\
\multicolumn{1}{|l|}{{TS-Teacher}} &
\multicolumn{1}{c}{{59.3}} &
\multicolumn{1}{c}{{58.2}} &
\multicolumn{1}{c}{{50.6}} &
\multicolumn{1}{c}{{63.3}} &
\multicolumn{1}{c}{{61.0}} &
\multicolumn{1}{c|}{{58.4}} \\
\multicolumn{1}{|l|}{{TS-ATT}} &
\multicolumn{1}{c}{{58.7}} &
\multicolumn{1}{c}{{57.0}} &
\multicolumn{1}{c}{{52.6}} &
\multicolumn{1}{c}{{67.6}} &
\multicolumn{1}{c}{{63.2}} &
\multicolumn{1}{c|}{{58.8}} \\
\multicolumn{1}{|l|}{{TS-BT}} & 
\multicolumn{1}{c}{{59.1}} &
\multicolumn{1}{c}{{59.0}} &
\multicolumn{1}{c}{{53.9}} &
\multicolumn{1}{c}{{65.8}} &
\multicolumn{1}{c}{{66.1}} &
\multicolumn{1}{c|}{{61.0}} \\
\hline
\multicolumn{1}{|l|}{{\system}} &
\multicolumn{1}{c}{{{\textbf{68.8}}}} &
\multicolumn{1}{c}{{{\textbf{72.2}}}} &
\multicolumn{1}{c}{{{\textbf{64.0}}}} &
\multicolumn{1}{c}{{{\textbf{72.0}}}} & 
\multicolumn{1}{c}{{{\textbf{71.2}}}} &
\multicolumn{1}{c|}{{{\textbf{66.9}}}} \\
\hline 
\end{tabular}}
\caption{Micro-averaged F1 for 9-class EDU-level aspect detection in product reviews}
\label{tb:amazon_review_result}
\end{table}




\begin{table}[t]
\small
\centering
\scalebox{0.9}{
\begin{tabular}{ccccccc}
\hline
\multicolumn{1}{|c|}{\multirow{1}{*}{}} 
& \multicolumn{6}{c|}{Restaurant review domain}
\\
\multicolumn{1}{|l|}{\multirow{1}{*}{Model}} &
\multicolumn{1}{c}{En}&
\multicolumn{1}{c}{Sp}&
\multicolumn{1}{c}{Fr}&
\multicolumn{1}{c}{Ru}&
\multicolumn{1}{c}{Du}&
\multicolumn{1}{c|}{Tur} \\
\hline \multicolumn{1}{|l|}{{W2V-Gold}} & \multicolumn{1}{c}{{58.8}} &
\multicolumn{1}{c}{{50.4}} &
\multicolumn{1}{c}{{50.4}} &
\multicolumn{1}{c}{{69.3}} &
\multicolumn{1}{c}{{51.4}} &
\multicolumn{1}{c|}{{55.7}} \\
\multicolumn{1}{|l|}{{BERT-Gold}} & \multicolumn{1}{c}{{63.1}} &
\multicolumn{1}{c}{{51.6}} &
\multicolumn{1}{c}{{50.6}} &
\multicolumn{1}{c}{{64.6}} &
\multicolumn{1}{c}{{53.5}} &
\multicolumn{1}{c|}{{55.3}} \\
\multicolumn{1}{|l|}{{\system-Gold}} & \multicolumn{1}{c}{{\textbf{70.5}}} &
\multicolumn{1}{c}{{\textbf{72.5}}} &
\multicolumn{1}{c}{{\textbf{65.4}}} &
\multicolumn{1}{c}{{\textbf{67.9}}} &
\multicolumn{1}{c}{{\textbf{73.8}}} &
\multicolumn{1}{c|}{{\textbf{65.4}}} \\
\hline \multicolumn{1}{|l|}{{LDA-Anchors}} & \multicolumn{1}{c}{{28.5}} &
\multicolumn{1}{c}{{17.7}} &
\multicolumn{1}{c}{{13.1}} &
\multicolumn{1}{c}{{14.8}} &
\multicolumn{1}{c}{{25.9}} &
\multicolumn{1}{c|}{{27.7}} \\
\multicolumn{1}{|l|}{{MATE}} &
\multicolumn{1}{c}{{41.0}} &
\multicolumn{1}{c}{{24.9}} &
\multicolumn{1}{c}{{25.8}} &
\multicolumn{1}{c}{{18.4}} &
\multicolumn{1}{c}{{36.1}} &
\multicolumn{1}{c|}{{39.0}} \\ \multicolumn{1}{|l|}{{MATE-UW}} & \multicolumn{1}{c}{{40.3}} &
\multicolumn{1}{c}{{18.3}} &
\multicolumn{1}{c}{{27.8}} &
\multicolumn{1}{c}{{21.8}} &
\multicolumn{1}{c}{{31.5}} &
\multicolumn{1}{c|}{{25.2}} \\
\multicolumn{1}{|l|}{{TS-Teacher}} & \multicolumn{1}{c}{{44.9}} &
\multicolumn{1}{c}{{41.8}} &
\multicolumn{1}{c}{{34.1}} &
\multicolumn{1}{c}{{54.4}} &
\multicolumn{1}{c}{{40.7}} &
\multicolumn{1}{c|}{{30.2}}  \\
\multicolumn{1}{|l|}{{TS-ATT}} & \multicolumn{1}{c}{{47.8}} &
\multicolumn{1}{c}{{41.7}} &
\multicolumn{1}{c}{{32.4}} &
\multicolumn{1}{c}{{59.0}} &
\multicolumn{1}{c}{{42.1}} &
\multicolumn{1}{c|}{{42.3}} \\
\multicolumn{1}{|l|}{{TS-BT}} & \multicolumn{1}{c}{{51.8}} &
\multicolumn{1}{c}{{42.0}} &
\multicolumn{1}{c}{{39.2}} &
\multicolumn{1}{c}{{58.0}} &
\multicolumn{1}{c}{{43.0}} &
\multicolumn{1}{c|}{{45.0}} \\
\multicolumn{1}{|l|}{{\system}} &
\multicolumn{1}{c}{{\textbf{57.9}}} &
\multicolumn{1}{c}{{\textbf{65.7}}} &
\multicolumn{1}{c}{{\textbf{48.6}}} &
\multicolumn{1}{c}{{\textbf{62.9}}} &
\multicolumn{1}{c}{{\textbf{57.2}}} &
\multicolumn{1}{c|}{\textbf{50.8}} \\
\hline 
\end{tabular}}
\caption{Micro-averaged F1 for 12-class sentence-level aspect detection in restaurant reviews}
\label{tb:restaurant_revreview}
\vspace{-1.5 pc}
\end{table}

\subsection{Experimental Results}

\noindent \textbf{Overall Inference Performance} 
Tables~\ref{tb:amazon_review_result} and~\ref{tb:restaurant_revreview} show the
results for aspect extraction on both datasets. We observe that \system
achieves superior performance. For example, in Amazon product reviews, compared
to TS-W2V, \system yields F1 performance gains of 16.0\%, 8.1\%, 31.9\%,
24.7\%, 11.3\%, and 17.4\% on Bags, KBs, Boots, B/T, TVs, and VCs,
respectively; similar trends are observed in the restaurant review dataset.
Moreover, the reduction in the parameter size of \system is also remarkable:
97.8\% versus TS-BT.\footnote{The parameter sizes of \system and TS-BT are 2.5M and
109.5M.} These results demonstrate the effectiveness of the proposed hyperbolic
disentangled aspect extractor (\system). 

We also observe the weakly unsupervised approaches MATE* and TS-*
significantly outperform the unsupervised approaches LDA-Anchors and ABAE,
suggesting the effectiveness of seed words. Note our reproduced
SSCL-BT* does not consistently outperform MATE, perhaps because
SSCL-BT relies heavily on the quality of initial k-means centroids since
poorly initialized centroids may cause model-inferred aspects after training to
lack good coverage for gold-standard aspects, and thus make manual mapping more
difficult. %

Compared to fully supervised (*-Gold)
models, \system dramatically reduces the performance gap between weakly supervised
approaches and fully supervised approaches and even outperforms in Spanish and 
Dutch restaurant reviews. Moreover, we investigate \system's
performance when given ground-truth aspect labels denoted by \system-Gold
in Table~\ref{tb:restaurant_revreview}. The result shows that \system-Gold
outperforms both W2V-Gold and BERT-Gold, showing the effectiveness of the
proposed hyperbolic disentangled based approach. Note that we vary the ratio of
ground-truth aspect labels and compare model performance for different
label ratios; these result are provided in the appendix.





\begin{figure}[htb]
\centering
\vspace{-1.0 pc}
\includegraphics[scale=0.29, trim={40 0 0 0}]{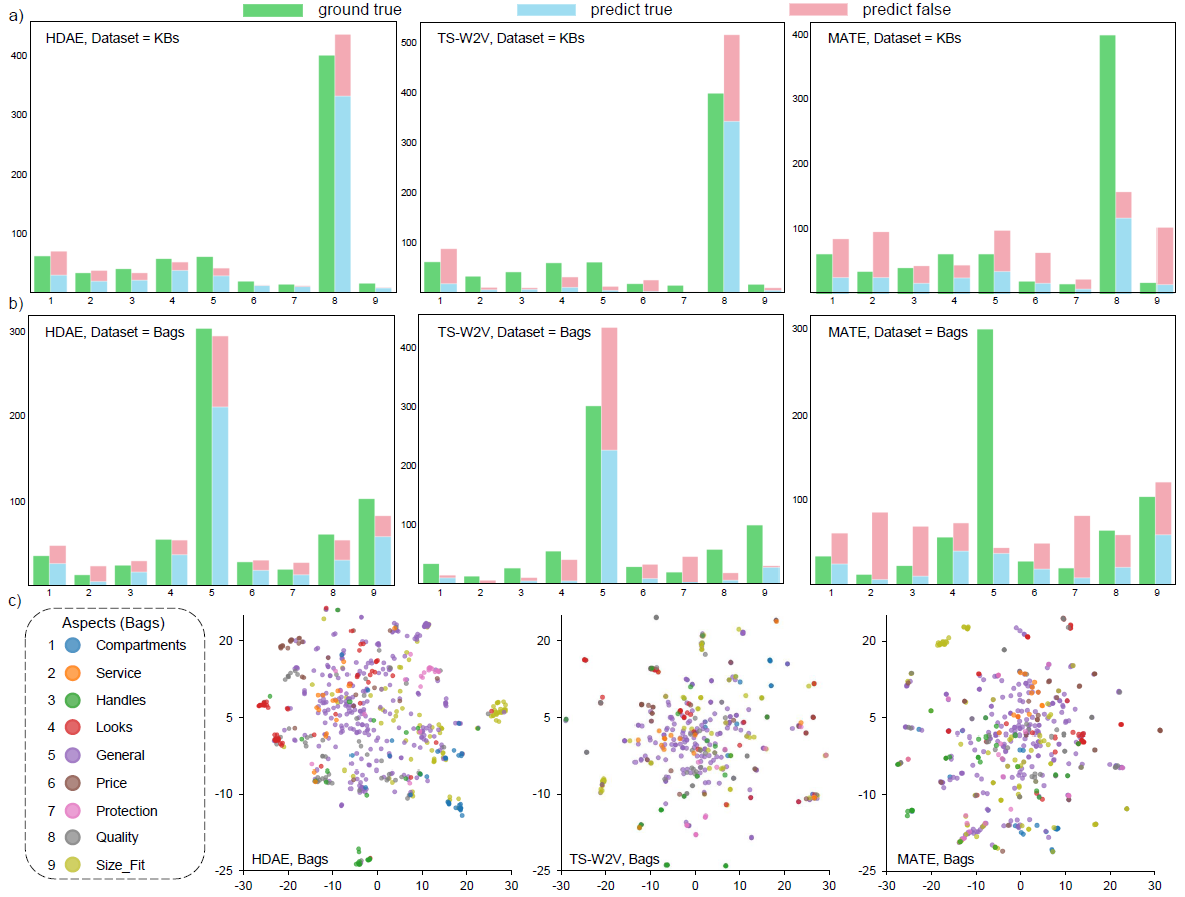}
\vspace{-1.5 pc}
\caption{Inference performance per aspect of \system, TS-W2V, and MATE on 
the a)~KBs and b)~Bags datasets. On Bags, we use t-SNE to compare
c)~the embedding of each model, where the different colors represent
different aspects.}
\label{fig:aspect_infer_performance_for_each_asp}
\end{figure}


\noindent \textbf{Inference Performance per Aspect} Here we investigate 
the abilities that seed word based approaches infer on different aspects, shown in
Fig.~\ref{fig:aspect_infer_performance_for_each_asp}(a) and~(b). First, we
observe that TS-W2V puts a greater focus on the general aspect, possibly because 
the teacher always predicts review segments as general aspect if no
seed word appears. Second, compared to MATE and TS-W2V, \system yields
better inference performance in almost all aspects without putting excessive
bias on certain predictions, showing better aspect inference ability. To
further investigate the performance on Bags,\footnote{For other
datasets, refer to the appendix in our arxiv version.} we compared sentence vectors $\textbf{v}_s$
of each model\footnote{For \system, the hyperbolic sentence vector
$\mathrm{exp}_{0}$($\textbf{v}_s$) is used.} by using t-SNE to visualize
vectors, as shown in Fig.~\ref{fig:aspect_infer_performance_for_each_asp}(c).
We find that a well-differentiated sentence vector benefits the model's aspect
inference ability. First, compared to \system and MATE, less-separated vectors
are found for TS-W2V, which correlates to the fact that TS-W2V performs poorly
in every aspect except the general aspect. Second, we observe for both the
\textit{looks} (red) and \textit{size fit} (yellow) aspects, differentiated
vector clusters are found in \system and MATE, which correlates to the good
accuracy in both aspects. Third, compared to MATE, the differentiated vector
clusters of the \textit{handles} (green), \textit{protection} (pink),
\textit{price} (brown), and \textit{compartments} (blue) aspects explain
\system's better inference ability in those aspects.

\vspace{-0.5 pc}

%



\begin{table}[t]
\fontsize{9}{10}\selectfont
\small
\centering
{
\begin{tabular}{|c|c|c|c|c|c|c|c|c|}
\hline \multicolumn{1}{|c|}{\multirow{1}{*}{Ablation}} &
\multicolumn{1}{|c|}{Bag} &
\multicolumn{1}{|c|}{KBs} &
\multicolumn{1}{|c|}{B/T} & \multicolumn{1}{|c|}{Boots} &
\multicolumn{1}{|c|}{TV} &
\multicolumn{1}{|c|}{VCs} 
\\
\hline 
\multicolumn{1}{|l|}{HDAE} &
\multicolumn{1}{|c|}{68.8} &
\multicolumn{1}{|c|}{72.2} &
\multicolumn{1}{|c|}{72.0} &
\multicolumn{1}{|c|}{64.0} &
\multicolumn{1}{|c|}{71.2} & \multicolumn{1}{|c|}{66.9} 
\\
\multicolumn{1}{|l|}{HDAE ($\lambda$ = 0)} &
\multicolumn{1}{|c|}{{67.3}} &
\multicolumn{1}{|c|}{65.6} &
\multicolumn{1}{|c|}{{70.1}} &
\multicolumn{1}{|c|}{{60.5}} &
\multicolumn{1}{|c|}{54.1} & \multicolumn{1}{|c|}{59.1} 
\\
\multicolumn{1}{|l|}{MATE} &
\multicolumn{1}{|c|}{{46.2}} &
\multicolumn{1}{|c|}{43.5} &
\multicolumn{1}{|c|}{{52.2}} &
\multicolumn{1}{|c|}{{45.6}} &
\multicolumn{1}{|c|}{48.8} & \multicolumn{1}{|c|}{42.3} 
\\
\hline 
\end{tabular}}
\vspace{-0.5 pc}
\caption{\system ablation study. The $\lambda$ is the ratio of distillation objective loss. When $\lambda$ is 0, the distillation objective $J_d$ is not used.}
\vspace{-1.5 pc}
\label{tb:RMPR_ablation2}
\end{table}

\subsection{Ablation Study and Parameter Sensitivity}

To verify the effectiveness of the proposed components, we conducted an
ablation study for \system, as shown in Table~\ref{tb:RMPR_ablation}. After
removing the hyperbolic aspect classifier (3) and aspect disentangle module (4), we
observe drops in performance, indicating the effect of the proposed components.
Note that (4), which only contains the hyperbolic aspect classifier, outperforms
the MATE-* and TS-* models, showing that the proposed hyperbolic aspect
classifier effectively leverages the seed words. Furthermore, removing the
hyperbolic distance function for the disentangled aspect representation, shown by
(2), degrades performance, suggesting that the hyperbolic distance function is
indispensable for modeling the latent semantic meanings of seed words.

\begin{figure}[htb]
\vspace{-1.0 pc}
\centering
\includegraphics[scale=0.19, trim={40 0 0 0}]{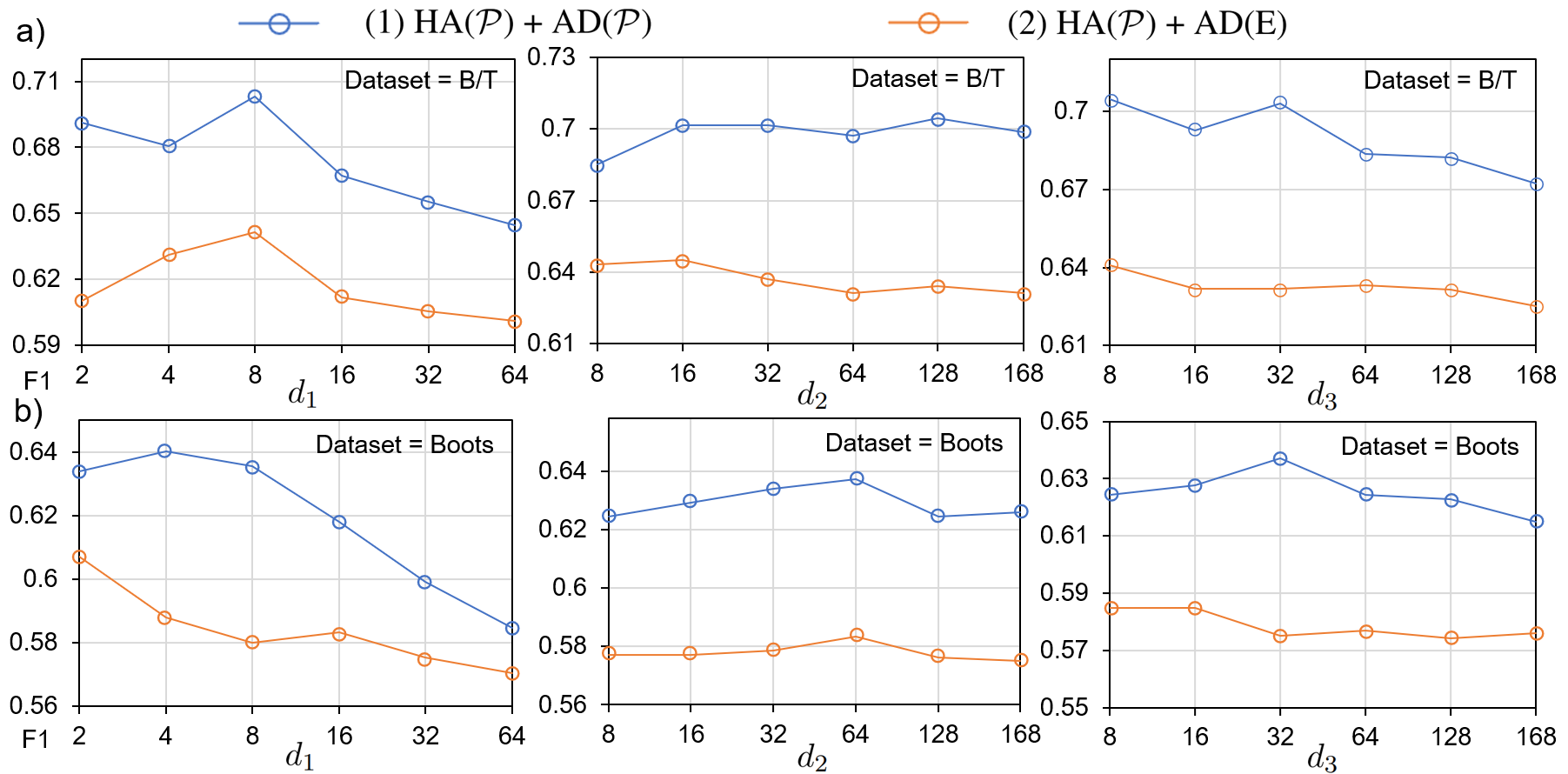}
\vspace{-1.5 pc}
\caption{Micro-averaged F1 scores of (1) and (2) with different $d_1$, $d_2$,
and $d_3$ on a) B/T and b) Boots datasets}
\label{fig:HDAE_d1d2d3_para}
\vspace{-0.5 pc}
\end{figure}


Then, we investigated the sensitivity of latent semantic modeling distance $d_1, d_2,$ and
$d_3$\footnote{For other parameters, please refer to the appendix in our arxiv version.} on (1) and (2),
as shown in Fig.~\ref{fig:HDAE_d1d2d3_para}. We offer the following observations.
First, both (1) and (2) achieve the best results when a small $d_1$, e.g,
$d_1 \leq 8$, is set, demonstrating the importance of narrowing the gap between
seed word pairs when modeling latent semantic meanings. Also, (1) and~(2) 
both perform better when a large $d_2$, e.g, $d_2 \geq 64$, is set, verifying the
importance of independence modeling. Last, (1) and~(2) achieve the best
performance when $d_3$ is set to around 8 to 32, perhaps due to the strong
regularization on each latent semantic meaning introduced when $d_3$ is too large.


\begin{table}[t]
\fontsize{9}{10}\selectfont
\small
\scalebox{0.9}{
\noindent\begin{tabular}{|c|c|c|c|c|}
\hline 
\multicolumn{2}{|l}{\multirow{1}{*}{a) The keyboard works very well.}} &
\multicolumn{1}{c|}{GT: General} 
\\
\multicolumn{3}{|l|}{\multirow{1}{*}{Seed Words: think, recommend, purchase, using, unit, star, microsoft}}
\\
\hline
\multicolumn{1}{|c|}{\system: General \color{blue} \Checkmark} &
\multicolumn{1}{|c|}{{MATE: General \color{blue} \Checkmark}} &
\multicolumn{1}{c|}{{TS-W2V: General \color{blue} \Checkmark}} 
\\
\hline 
\multicolumn{2}{|l}{\multirow{1}{*}{b) The {\color{red}color} is nice, more light {\color{red}blue}.}} &
\multicolumn{1}{c|}{GT: Color} 
\\
\multicolumn{3}{|l|}{\multirow{1}{*}{Seed Words: {\color{red}color}, love, style, unbelievably, gorgeous, {\color{red}blue}}}
\\
\hline
\multicolumn{1}{|c|}{\system: Color \color{blue} \Checkmark} &
\multicolumn{1}{|c|}{{MATE: Color \color{blue} \Checkmark}} &
\multicolumn{1}{c|}{{TS-W2V: Color \color{blue} \Checkmark}} 
\\
\hline 
\multicolumn{2}{|l}{\multirow{1}{*}{c) What I received is a {\color{red}grayish brown} shoe.}} &
\multicolumn{1}{c|}{GT: Color} 
\\
\multicolumn{3}{|l|}{\multirow{1}{*}{Seed Words: {\color{red}color}, love, style, unbelievably, gorgeous, blue}} 
\\
\hline
\multicolumn{1}{|c|}{\system: Color \color{blue} \Checkmark} &
\multicolumn{1}{|c|}{{MATE: General \color{red}\XSolidBrush}} &
\multicolumn{1}{c|}{{TS-W2V: Price \color{red}\XSolidBrush}} 
\\
\hline 
\multicolumn{2}{|l}{\multirow{1}{*}{d) It's not {\color{red}leather}.}} &
\multicolumn{1}{c|}{GT: Quality } 
\\
\multicolumn{3}{|l|}{\multirow{1}{*}{Seed Words: quality, {\color{red}material}, handle, poor, broke, durable, month}} 
\\
\hline
\multicolumn{1}{|c|}{\system: Quality \color{blue} \Checkmark} &
\multicolumn{1}{|c|}{{MATE: Handles \color{red}\XSolidBrush}} &
\multicolumn{1}{|c|}{{TS-W2V: General \color{red}\XSolidBrush}}  

\\
\hline 
\multicolumn{2}{|l}{\multirow{1}{*}{e) On the other hand, find it to be too {\color{red}stiff}.}} &
\multicolumn{1}{c|}{GT: Comfort} 
\\
\multicolumn{3}{|l|}{\multirow{1}{*}{Seed Words: feel, comfortable, mushy, key, like, {\color{red}difficult}}} 
\\
\hline
\multicolumn{1}{|c|}{\system: Comfort \color{blue} \Checkmark} &
\multicolumn{1}{|c|}{{MATE: Function \color{red}\XSolidBrush}} &
\multicolumn{1}{|c|}{{TS-W2V: General \color{red}\XSolidBrush}} 
\\
\hline 
\end{tabular}}
\caption{Comparison of predictions on sample Product review segments between \system, MATE, and MATE. For each review segment, the ground truth (GT) aspect and its corresponding seed words are provided.}
\vspace{-1.5 pc}
\label{tb:case_st}
\end{table}





\vspace{-0.5 pc}
\subsection{Case study}

To more closely investigate the aspect inference ability of \system, we compare the
predictions made by \system, MATE, and TS-W2V, the results of which are shown in
Table~\ref{tb:case_st}. For the example in Table~\ref{tb:case_st}(b), we
see that the review segment contains keywords such as \textit{color} and
\textit{blue} which are explicitly captured in aspect seed words. All
models correctly infer and review the segment's aspect. However, for
cases in Table~\ref{tb:case_st}(c,d,e), the reviews' segments do not
explicitly match their aspect seed words 
  but instead match the hyponymic relations (is-a) present between seed words and review segments.   
For example, there are hierarchical
relations such as \textit{grayish} brown is a \textit{color}, \textit{leather} is a \textit{material},
and \textit{stiff} is a type of \textit{difficult} for cases in
Table~\ref{tb:case_st}(c,d,e). We find only \system correctly
recognizes the review segments' aspects. We thus conclude \system captures
and utilizes hyponymic relations (is-a) present
between seed words and review segments, deriving reasonable aspect inference
for each review segment and thus achieving better performance. Analogous
behavior is observed for other cases in the appendix. 

\begin{figure}[tb]
\vspace{-1.0 pc}
\centering
\includegraphics[scale=0.18, trim={20 0 0 0}]{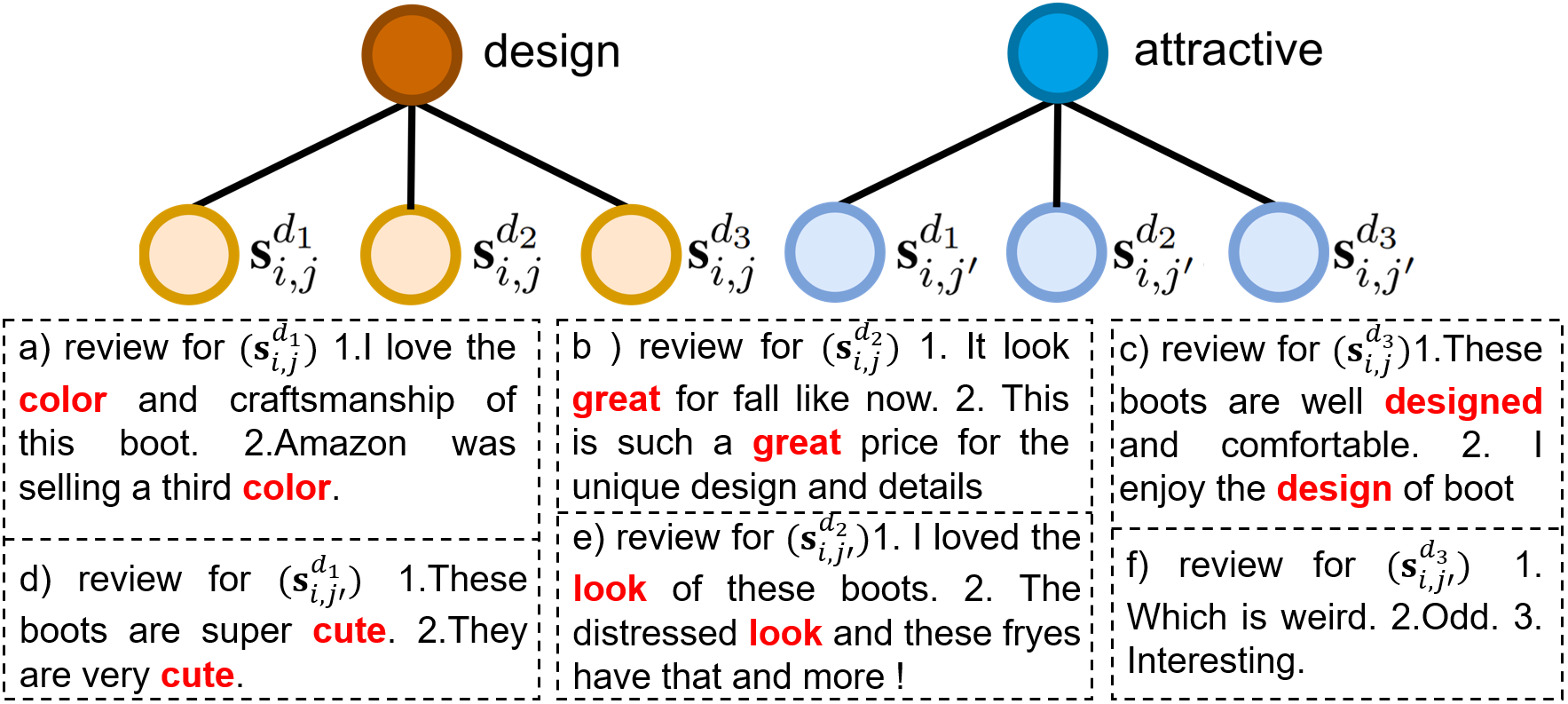}
\vspace{-0.5 pc}
\caption{Interpretability of latent semantic meanigs of seed words. Best viewed in
color.}
\label{fig:case_study}
\vspace{-1.0 pc}
\end{figure}




To explore the interpretability of the seed words' latent semantic meanings, we conducted a
case study in which we randomly selected review segments from the boot
domain's \textit{look} aspect and investigated its association with each aspect
of latent semantic meaning. Figure~\ref{fig:case_study} shows the review segments
captured by each seed word's latent semantic meaning: we find that each aspect's
latent semantics focus on a distinct type of review segment. For example, for
the seed word \textit{design}, the latent semantic meaning $\textbf{s}^{d_1}_{i,j}$
focuses on segments with \textit{color} information, whereas
$\textbf{s}^{d_2}_{i,j}$ focuses on segments with the
\textit{great} keyword. Likewise, for the seed word \textit{attractive}, the latent
semantic meaning $\textbf{s}^{d_1}_{i,j'}$ focuses on segments with
\textit{cute} information, whereas $\textbf{s}^{d_3}_{i,j'}$ focuses on 
segments with \textit{unattractive} information. These results demonstrate that the
proposed aspect disentanglement module assists \system in modeling different
latent semantics for each seed word. Also, \system finds the most
relevant latent semantic meanings for each review segment, explaining the improvements in
the aspect inference ability.

\vspace{-0.5pc}


\section{Conclusions and Future Work}

We present \system, a hyperbolic disentangled aspect extractor which
includes a hyperbolic aspect classifier and an aspect disentanglement module. On two
datasets, \system, with its 97.8\% reductions in parameter size versus TS-BT,
shows superior aspect inference ability, further substantiated by an embedding
visualization. The effect of the proposed components is proved by an ablation
study, a parameter sensitivity study, and a case study. 

In the future, we plan to explore the proposed module on other aspect-based
sentiment analysis (ABSA) subtasks. We would also like to further improve
the performance of the proposed components, for instance by setting up alignment scores for
different aspect word pairs when modeling seed word dependence.

\section*{Acknowledgement}
This research is partially supported by Ministry of Science and  Technology,  Taiwan  under  the  project  contract 110-2221-E-001-001- and 110-2634-F-002-051.

\bibliography{aaai22}

\section{APPENDIX}

We provide more information on datasets (Section 1.1) and implementation details (Section 1.2). Besides, we report more detailed experimental results (Section 1.3), and parameter sensitivity analysis (Section 1.4).

\subsection{Datasets}

During training, segment aspect labels (9-class for product reviews and 12-class for restaurant reviews) are not available but provided during validation and test. For instance, in the Laptop Bags domain, the review segments' aspects could be Compartments, Customer Service, Looks, or Price.  Note that a general aspect is assigned if the segment doesn't discuss any specific aspects. For each domain, we train our model on a training set only with seed words $G$ via the teacher. For the aspect seed words, we follow~\citet{angelidis2018summarizing, DBLP:journals/corr/abs-1909-00415} to use the same 30 seed words for two datasets. Besides, for two datasets, we follow~\citet{angelidis2018summarizing, DBLP:journals/corr/abs-1909-00415} to do data preprocessing, such as removing stop-words. 
 
 
 
 For the Amazon review dataset, the reviews of each domain are already segmented by~\cite{angelidis2018summarizing}, where they use a Rhetorical Structure Theory parser~\cite{feng-hirst-2012-text} to segment reviews into elementary discourse units (EDUs). Across domains, the average numbers of training, validation, and test segments are around 1 million, 700 segments, respectively. For restaurant reviews, the reviews of each language are already segmented into sentences.  Across languages, the average number of training and test segments is around 2500 and 800 segments, respectively.

\begin{table}[t]
\small
\centering
\scalebox{1.0}{
\begin{tabular}{ccccccc}
\hline
\multicolumn{1}{|c|}{\multirow{1}{*}{}} 
& \multicolumn{6}{c|}{Product Review Domain}
\\
\multicolumn{1}{|c|}{\multirow{1}{*}{}} &
\multicolumn{1}{c}{Bags}&
\multicolumn{1}{c}{KBs}&
\multicolumn{1}{c}{Boots}&
\multicolumn{1}{c}{B/T}&
\multicolumn{1}{c}{TVs}&
\multicolumn{1}{c|}{VCs}\\
\hline \multicolumn{1}{|c|}{{$\beta$}} & 
\multicolumn{1}{c}{{0.01}} &
\multicolumn{1}{c}{{0.02}} &
\multicolumn{1}{c}{{0.01}} &
\multicolumn{1}{c}{{0.02}} &
\multicolumn{1}{c}{{0.02}} &
\multicolumn{1}{c|}{{0.02}} \\
\multicolumn{1}{|c|}{{I}} & 
\multicolumn{1}{c}{{6}} &
\multicolumn{1}{c}{{4}} &
\multicolumn{1}{c}{{6}} &
\multicolumn{1}{c}{{8}} &
\multicolumn{1}{c}{{6}} &
\multicolumn{1}{c|}{{4}} \\
\multicolumn{1}{|c|}{{$\tau$}} & 
\multicolumn{1}{c}{{1e-3}} &
\multicolumn{1}{c}{{1e-5}} &
\multicolumn{1}{c}{{1e-5}} &
\multicolumn{1}{c}{{1e-4}} &
\multicolumn{1}{c}{{1e-3}} &
\multicolumn{1}{c|}{{1e-6}} \\
\multicolumn{1}{|c|}{$\lambda$} & 
\multicolumn{1}{c}{{5}} &
\multicolumn{1}{c}{{3000}} &
\multicolumn{1}{c}{{10}} &
\multicolumn{1}{c}{{5}} &
\multicolumn{1}{c}{{10}} &
\multicolumn{1}{c|}{{3000}} \\
\multicolumn{1}{|c|}{{$d_1$}} &
\multicolumn{1}{c}{{2}} &
\multicolumn{1}{c}{{128}} &
\multicolumn{1}{c}{{16}} &
\multicolumn{1}{c}{{16}} &
\multicolumn{1}{c}{{16}} &
\multicolumn{1}{c|}{{128}} \\
\multicolumn{1}{|c|}{{$d_2$}} &
\multicolumn{1}{c}{{128}} &
\multicolumn{1}{c}{{32}} &
\multicolumn{1}{c}{{16}} &
\multicolumn{1}{c}{{32}} &
\multicolumn{1}{c}{{16}} &
\multicolumn{1}{c|}{{4}} \\
\multicolumn{1}{|c|}{{$d_3$}} & 
\multicolumn{1}{c}{{8}} &
\multicolumn{1}{c}{{64}} &
\multicolumn{1}{c}{{8}} &
\multicolumn{1}{c}{{16}} &
\multicolumn{1}{c}{{16}} &
\multicolumn{1}{c|}{{64}} \\
\hline 
\end{tabular}}
\caption{Hyper-parameter settings for the product review.}
\label{tb:para_ProductReviewDomain}
\vspace{-0.9 pc}
\end{table}

\begin{table}[t]
\small
\centering
\scalebox{1.0}{
\begin{tabular}{ccccccc}
\hline
\multicolumn{1}{|c|}{\multirow{1}{*}{}} 
& \multicolumn{6}{c|}{Restaurant Review Domain}
\\
\multicolumn{1}{|c|}{\multirow{1}{*}{}} &
\multicolumn{1}{c}{En}&
\multicolumn{1}{c}{Sp}&
\multicolumn{1}{c}{Fr}&
\multicolumn{1}{c}{Ru}&
\multicolumn{1}{c}{Du}&
\multicolumn{1}{c|}{Tur}\\
\hline \multicolumn{1}{|c|}{{$\beta$}} & 
\multicolumn{1}{c}{{0.01}} &
\multicolumn{1}{c}{{0.02}} &
\multicolumn{1}{c}{{0.02}} &
\multicolumn{1}{c}{{0.02}} &
\multicolumn{1}{c}{{0.01}} &
\multicolumn{1}{c|}{{0.01}} \\
\multicolumn{1}{|c|}{{I}} & 
\multicolumn{1}{c}{{4}} &
\multicolumn{1}{c}{{4}} &
\multicolumn{1}{c}{{6}} &
\multicolumn{1}{c}{{2}} &
\multicolumn{1}{c}{{4}} &
\multicolumn{1}{c|}{{2}} \\
\multicolumn{1}{|c|}{{$\tau$}} & 
\multicolumn{1}{c}{{1e-3}} &
\multicolumn{1}{c}{{1e-5}} &
\multicolumn{1}{c}{{1e-6}} &
\multicolumn{1}{c}{{1e-3}} &
\multicolumn{1}{c}{{1e-3}} &
\multicolumn{1}{c|}{{1e-5}} \\
\multicolumn{1}{|c|}{$\lambda$} & \multicolumn{1}{c}{{100}} &
\multicolumn{1}{c}{{100}} &
\multicolumn{1}{c}{{10}} &
\multicolumn{1}{c}{{1000}} &
\multicolumn{1}{c}{{100}} &
\multicolumn{1}{c|}{{1}} \\
\multicolumn{1}{|c|}{{$d_1$}} &
\multicolumn{1}{c}{{16}} &
\multicolumn{1}{c}{{16}} &
\multicolumn{1}{c}{{32}} &
\multicolumn{1}{c}{{16}} &
\multicolumn{1}{c}{{32}} &
\multicolumn{1}{c|}{{2}} \\
\multicolumn{1}{|c|}{{$d_2$}} &
\multicolumn{1}{c}{{32}} &
\multicolumn{1}{c}{{64}} &
\multicolumn{1}{c}{{16}} &
\multicolumn{1}{c}{{32}} &
\multicolumn{1}{c}{{32}} &
\multicolumn{1}{c|}{{32}} \\
\multicolumn{1}{|c|}{{$d_3$}} & 
\multicolumn{1}{c}{{128}} &
\multicolumn{1}{c}{{128}} &
\multicolumn{1}{c}{{64}} &
\multicolumn{1}{c}{{32}} &
\multicolumn{1}{c}{{64}} &
\multicolumn{1}{c|}{{32}} \\
\hline 
\end{tabular}
}
\caption{Hyper-parameter settings for the restaurant review datasets.}
\label{tb:para_RestaurantReviewDomain}
\vspace{-0.9 pc}
\end{table}

\subsection{Implementation details}

For \system, the details hyper-parameter settings are given in Table~\ref{tb:para_ProductReviewDomain} and~\ref{tb:para_RestaurantReviewDomain}, which are determined by optimizing on a validation set. We also provide the parameter sensitivity experiment of latent  semantic  modeling  distance $d_1$, $d_2$, and $d_3$, Grumbel-Softmax temperature $\tau$, ratio of distillation objective $\lambda$, number of latent semantic $\mathrm{I}$ in section~\ref{sec:parameter_senetivity}. The total number of negative examples $k_n$ was set to 10. We followed
the procedure in~\citet{angelidis2018summarizing} to set the
200-dimensional word embeddings for the Amazon product reviews and the
300-dimensional multilingual word2vec embeddings
from~\citet{DBLP:journals/corr/RuderGB16} for restaurant reviews. For all models, the same 30 seed words were set per aspect. For \system, model parameters are optimized by using the Adam optimizer~\cite{2014arXiv1412.6980K}. For setting distillation objective, the teacher, a bag-of-word classifier, is implement, and we use iterative co-training to update each seed word's predictive quality. For TS-*, we report the result from iterative co-training, and in each round, we divide the learning rate by 10. For SSCL-BT*, we use code provided in\footnote{https://github.com/tshi04/AspDecSSCL} and conduct aspect mapping after training the teacher model. The smooth factor $\lambda$ is set to 0.5 and temperature is set to 1. For all models, the learning rate was selected from [$2\times {10}^{-4}$, $1\times {10}^{-6}$, $5\times {10}^{-7}$, $5\times {10}^{-8}$, $1\times {10}^{-8}$]. Other hyperparameters were optimized according to validation results. For each model, we repeat each experiment 5 times and report the average test performance with the parameter configuration that achieves the best validation performance. 


\begin{table}[t]
\small
\centering
\scalebox{0.9}{
\begin{tabular}{ccccccc}
\hline
\multicolumn{1}{|c|}{\multirow{1}{*}{}} 
& \multicolumn{6}{c|}{Restaurant Review Domain}
\\
\multicolumn{1}{|c|}{\multirow{1}{*}{Model}} &
\multicolumn{1}{c}{En}&
\multicolumn{1}{c}{Sp}&
\multicolumn{1}{c}{Fr}&
\multicolumn{1}{c}{Ru}&
\multicolumn{1}{c}{Du}&
\multicolumn{1}{c|}{Tur} \\
\hline  \multicolumn{1}{|c|}{{ABAE}} & 
\multicolumn{1}{c}{{35.8}} &
\multicolumn{1}{c}{{-}} &
\multicolumn{1}{c}{{-}} &
\multicolumn{1}{c}{{-}} &
\multicolumn{1}{c}{{-}} &
\multicolumn{1}{c|}{{-}} \\
\multicolumn{1}{|c|}{{SSCL-BT*}} & 
\multicolumn{1}{c}{{47.3}} &
\multicolumn{1}{c}{{-}} &
\multicolumn{1}{c}{{-}} &
\multicolumn{1}{c}{{-}} &
\multicolumn{1}{c}{{-}} &
\multicolumn{1}{c|}{{-}} \\
\multicolumn{1}{|c|}{{LDA-AR}} & \multicolumn{1}{c}{{28.5}} &
\multicolumn{1}{c}{{17.7}} &
\multicolumn{1}{c}{{13.1}} &
\multicolumn{1}{c}{{14.8}} &
\multicolumn{1}{c}{{25.9}} &
\multicolumn{1}{c|}{{27.7}} \\
\multicolumn{1}{|c|}{{MATE}} &
\multicolumn{1}{c}{{41.0}} &
\multicolumn{1}{c}{{24.9}} &
\multicolumn{1}{c}{{17.8}} &
\multicolumn{1}{c}{{18.4}} &
\multicolumn{1}{c}{{36.1}} &
\multicolumn{1}{c|}{{39.0}} \\ \multicolumn{1}{|c|}{{MATE-UW}} & \multicolumn{1}{c}{{40.3}} &
\multicolumn{1}{c}{{18.3}} &
\multicolumn{1}{c}{{19.2}} &
\multicolumn{1}{c}{{21.8}} &
\multicolumn{1}{c}{{31.5}} &
\multicolumn{1}{c|}{{25.2}} \\
\multicolumn{1}{|c|}{{TS-Teacher}} & \multicolumn{1}{c}{{44.9}} &
\multicolumn{1}{c}{{41.8}} &
\multicolumn{1}{c}{{34.1}} &
\multicolumn{1}{c}{{54.4}} &
\multicolumn{1}{c}{{40.7}} &
\multicolumn{1}{c|}{{30.2}}  \\
\multicolumn{1}{|c|}{{TS-ATT}} & \multicolumn{1}{c}{{47.8}} &
\multicolumn{1}{c}{{41.7}} &
\multicolumn{1}{c}{{32.4}} &
\multicolumn{1}{c}{{59.0}} &
\multicolumn{1}{c}{{42.1}} &
\multicolumn{1}{c|}{{42.3}} \\
\multicolumn{1}{|c|}{{TS-BT}} & \multicolumn{1}{c}{{51.8}} &
\multicolumn{1}{c}{{42.0}} &
\multicolumn{1}{c}{{39.2}} &
\multicolumn{1}{c}{{58.0}} &
\multicolumn{1}{c}{{43.0}} &
\multicolumn{1}{c|}{{45.0}} \\
\multicolumn{1}{|c|}{{\system}} &
\multicolumn{1}{c}{{\textbf{57.9}}} &
\multicolumn{1}{c}{{\textbf{65.7}}} &
\multicolumn{1}{c}{{\textbf{48.6}}} &
\multicolumn{1}{c}{{\textbf{62.9}}} &
\multicolumn{1}{c}{{\textbf{57.2}}} &
\multicolumn{1}{c|}{\textbf{50.8}} \\
\hline 
\end{tabular}}
\caption{Micro-averaged F1 reported for 12-class sentence-level aspect detection in restaurant reviews.}
\label{tb:restaurant_revreview_tmp}
\end{table}

\subsection{Experimental Results} we provide the performance the unsupervised based method ABAE and SSCL on English Restaurant review, as shown in Table~\ref{tb:restaurant_revreview_tmp}

\begin{figure*}[htb]
\centering
\includegraphics[scale=0.35, trim={0 0 0 0}]{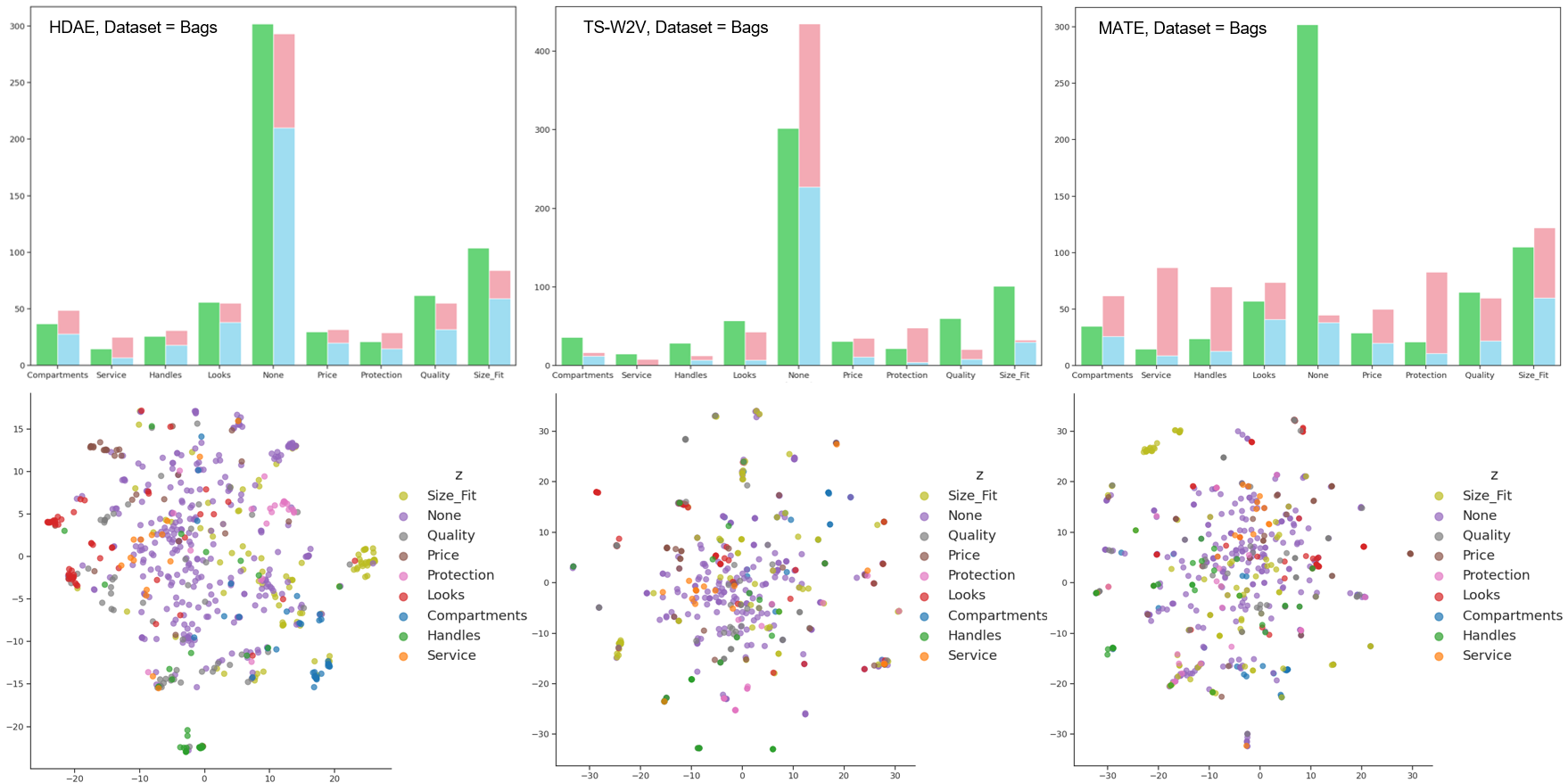}
\caption{Inference performance per aspect of \system, TS-W2V, and MATE on Bags dataset. The following figure is segment vector t-SNE visualization of each model, where the different color of point represent different aspect.}
\label{fig:bag_pre_emb}
\end{figure*}

\begin{figure*}[htb]
\centering
\includegraphics[scale=0.35, trim={0 0 0 0}]{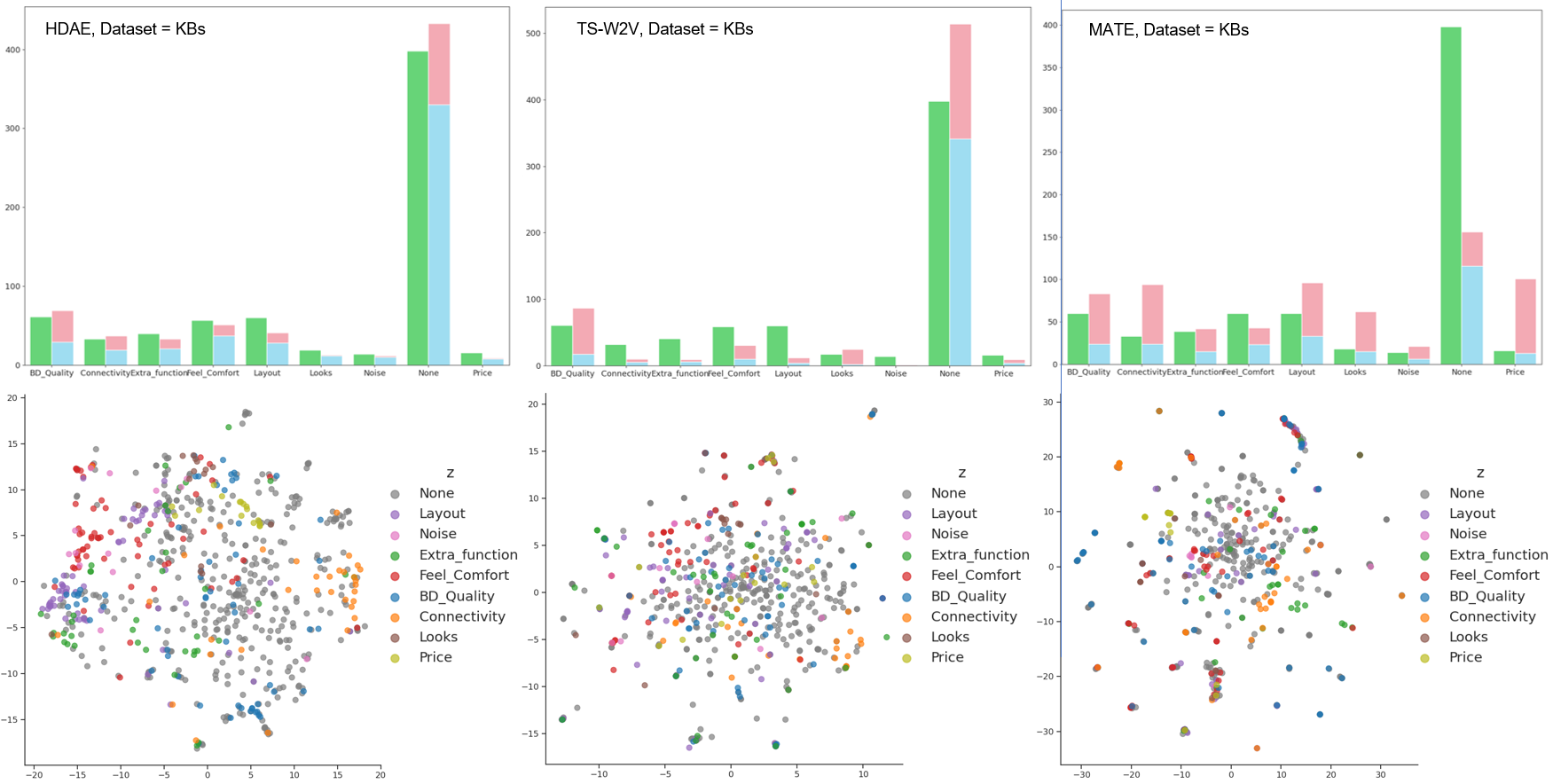}
\caption{Inference performance per aspect of \system, TS-W2V, and MATE on Keyboards dataset. The following figure is segment vector t-SNE visualization of each model, where the different color of point represent different aspect.}
\label{fig:kbs_pre_emb}
\end{figure*}

\begin{figure*}[htb]
\centering
\includegraphics[scale=0.35, trim={0 0 0 0}]{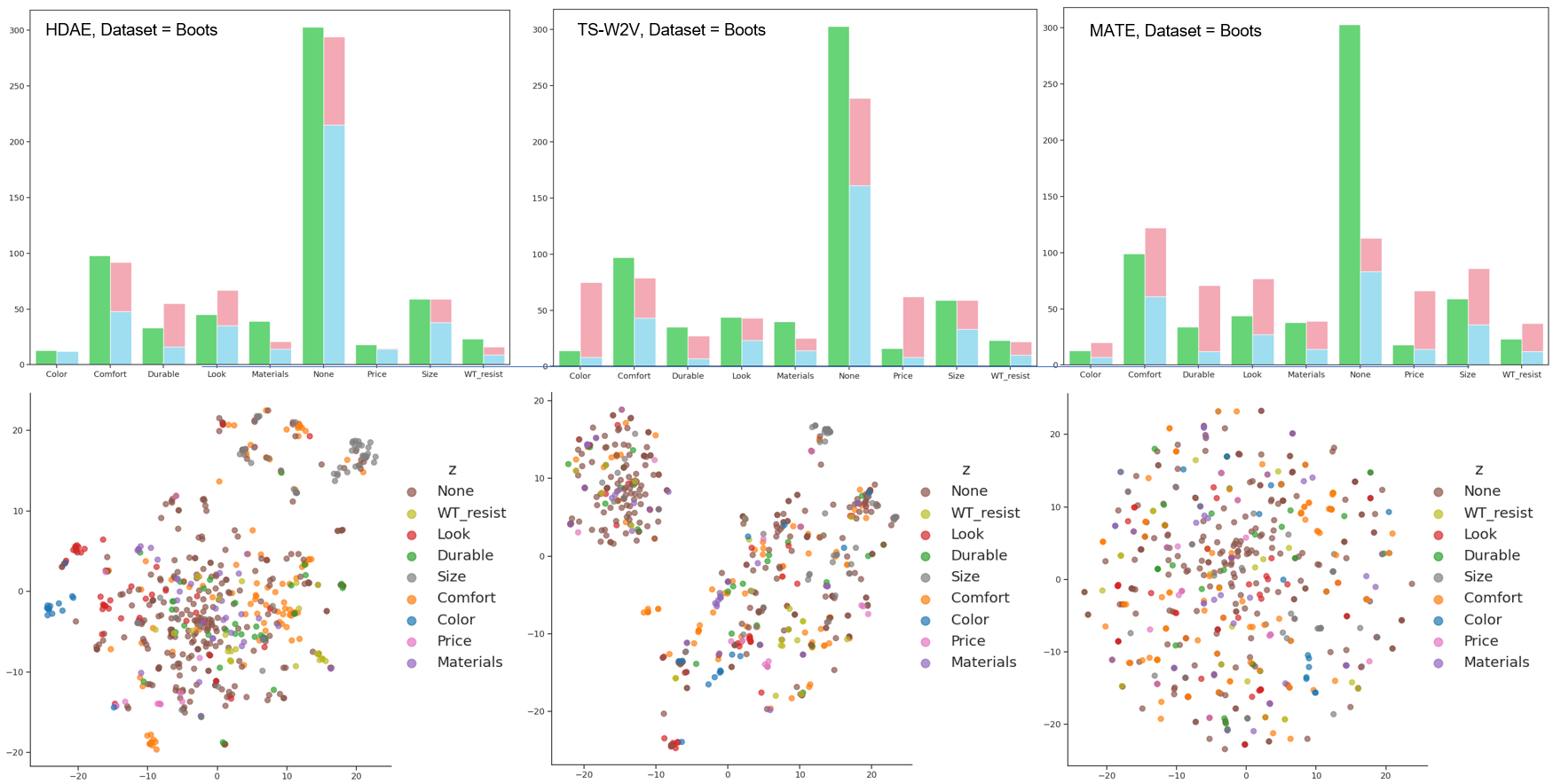}
\caption{Inference performance per aspect of \system, TS-W2V, and MATE on Boots dataset. The following figure is segment vector t-SNE visualization of each model, where the different color of point represent different aspect.}
\label{fig:boots_pre_emb}
\end{figure*}

\begin{figure*}[htb]
\centering
\includegraphics[scale=0.35, trim={0 0 0 0}]{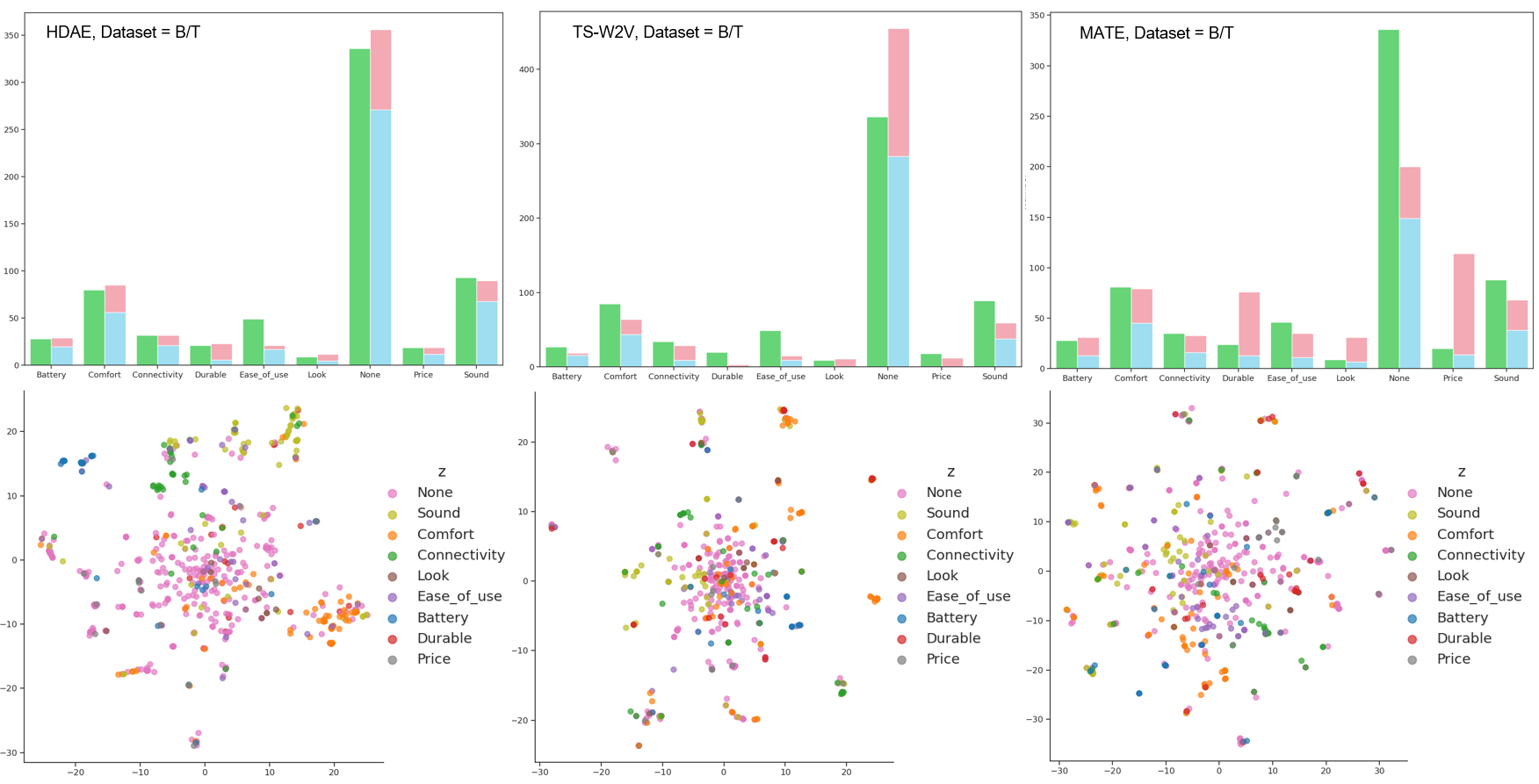}
\caption{Inference performance per aspect of \system, TS-W2V, and MATE on Bluetooth Headsets dataset. The following figure is segment vector t-SNE visualization of each model, where the different color of point represent different aspect.}
\label{fig:bt_pre_emb}
\end{figure*}

\begin{figure*}[htb]
\centering
\includegraphics[scale=0.35, trim={0 0 0 0}]{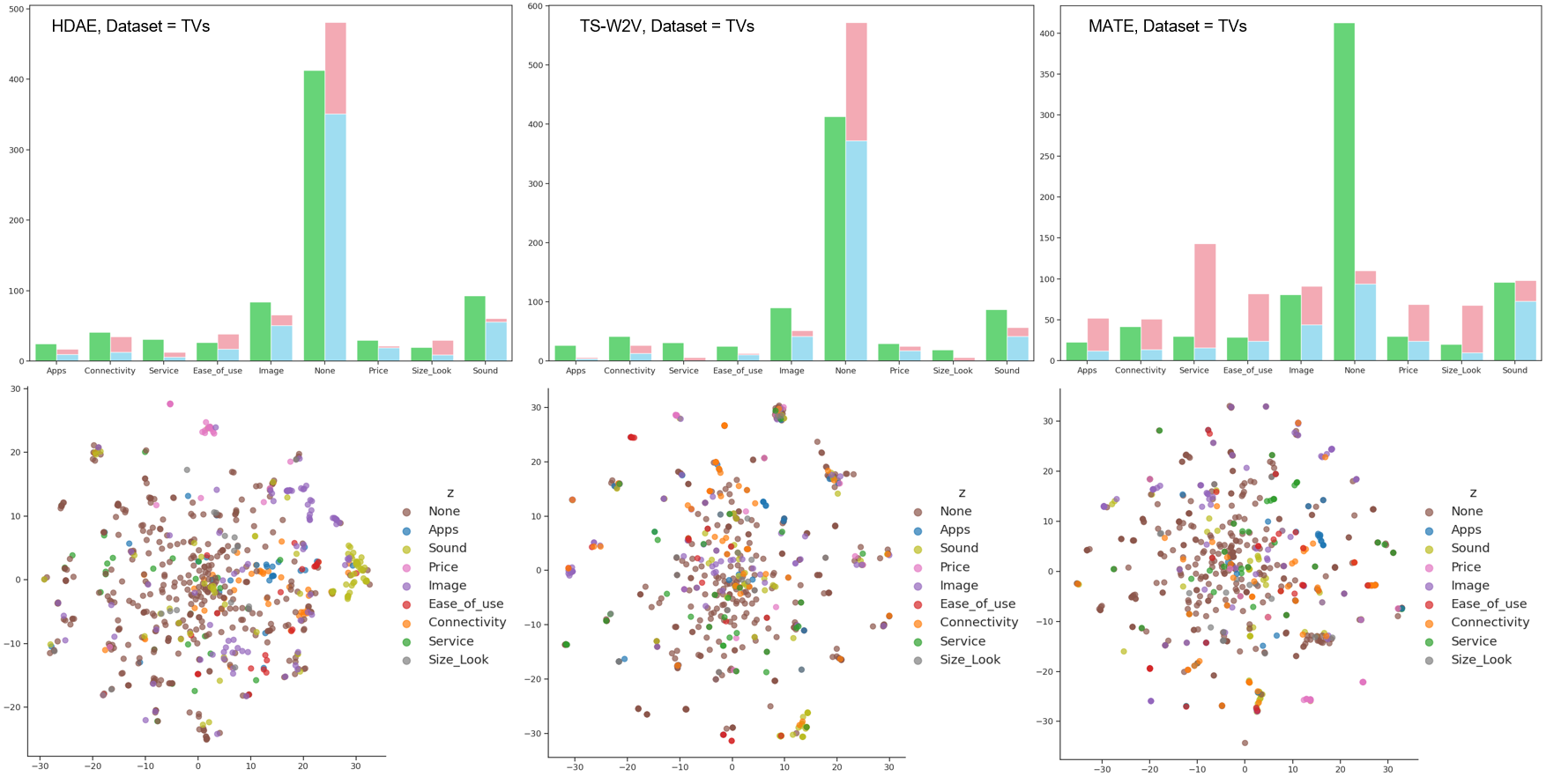}
\caption{Inference performance per aspect of \system, TS-W2V, and MATE on Televisions dataset. The following figure is segment vector t-SNE visualization of each model, where the different color of point represent different aspect.}
\label{fig:tv_pre_emb}
\end{figure*}

\begin{figure*}[htb]
\centering
\includegraphics[scale=0.31, trim={0 0 0 0}]{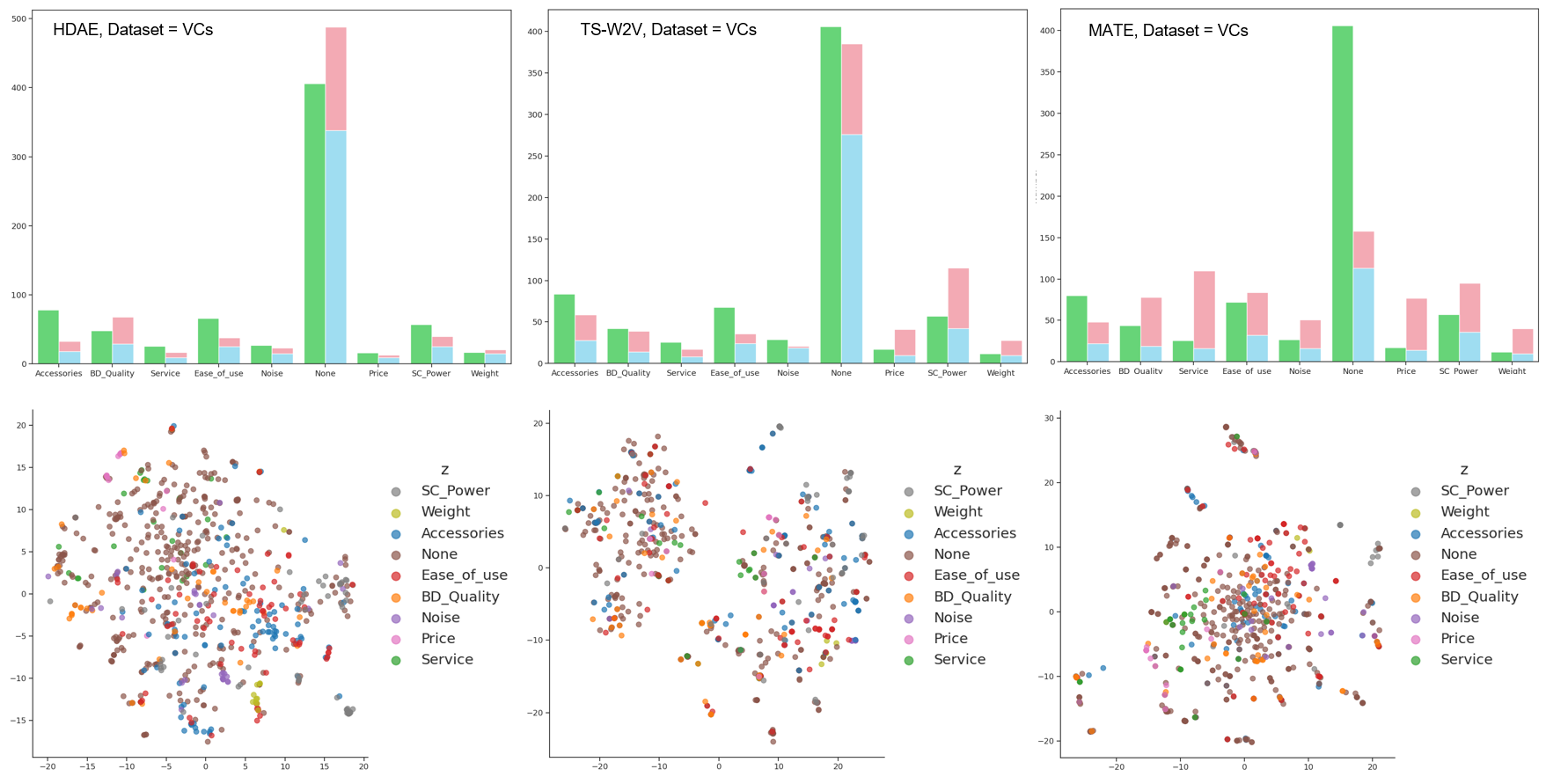}
\caption{Inference performance per aspect of \system, TS-W2V, and MATE on Vacuums dataset. The following figure is segment vector t-SNE visualization of each model, where the different color of point represent different aspect.}
\label{fig:vc_pre_emb}
\end{figure*}

Then, we provide more results for seed word based approaches' inference performance per aspect and their corresponding embedding visualization on Bags, Bluetooth Headsets, Boots, Keyboards, Televisions, and Vacuums (VCs) datasets, shown in Figure~\ref{fig:bag_pre_emb},~\ref{fig:kbs_pre_emb},~\ref{fig:boots_pre_emb},~\ref{fig:bt_pre_emb},~\ref{fig:tv_pre_emb},~\ref{fig:vc_pre_emb}, respectively.

\begin{table}[t]
\small
\centering
\vspace{-0.5 pc}
\scalebox{1.0}{
\begin{tabular}{p{2.5cm}|ccccccccc}
\hline \multicolumn{1}{c|}{\multirow{1}{*}{$\tau$}} & \multicolumn{1}{c}{{1e-1}} &
\multicolumn{1}{c}{{1e-2}} & \multicolumn{1}{c}{{1e-3}} & \multicolumn{1}{c}{{1e-4}} & \multicolumn{1}{c}{{1e-5}} & \multicolumn{1}{c}{{1e-6}} \\ \hline \multicolumn{1}{c|}{Bags} & \multicolumn{1}{c}{{67.1}} &
\multicolumn{1}{c}{{67.5}} & 
\multicolumn{1}{c}{{\textbf{68.8}}} &
\multicolumn{1}{c}{{67.8}} &
\multicolumn{1}{c}{{67.8}} & 
\multicolumn{1}{c}{{68.2}} 
\\ 
\multicolumn{1}{c|}{B/T} & \multicolumn{1}{c}{{68.4}} &
\multicolumn{1}{c}{{69.1}} & 
\multicolumn{1}{c}{{71.1}} &
\multicolumn{1}{c}{{\textbf{71.9}}} &
\multicolumn{1}{c}{{70.4}} & 
\multicolumn{1}{c}{{70.3}} 
\\ 
\multicolumn{1}{c|}{Boots} & \multicolumn{1}{c}{{61.8}} &
\multicolumn{1}{c}{{62.4}} & 
\multicolumn{1}{c}{{63.5}} &
\multicolumn{1}{c}{{63.7}} &
\multicolumn{1}{c}{{\textbf{64.0}}} & 
\multicolumn{1}{c}{{63.7}} 
\\ 
\multicolumn{1}{c|}{TVs} & \multicolumn{1}{c}{{69.8}} &
\multicolumn{1}{c}{{70.8}} & 
\multicolumn{1}{c}{{\textbf{71.2}}} &
\multicolumn{1}{c}{{70.4}} &
\multicolumn{1}{c}{{70.3}} & 
\multicolumn{1}{c}{{70.0}} \\ 
\hline
\end{tabular}}
\caption{Micro-averaged F1 of \system given grumbel-softmax temperature $\tau$.}
\label{tb:by_gml_soft_tem}
\end{table}

\begin{table}[t]
\small
\centering
\vspace{-0.5 pc}
\scalebox{1.0}{
\begin{tabular}{p{2.5cm}|ccccccccc}
\hline \multicolumn{1}{c|}{\multirow{1}{*}{$\lambda$}} & \multicolumn{1}{c}{{0}} & \multicolumn{1}{c}{{5}}
& \multicolumn{1}{c}{{10}} &
\multicolumn{1}{c}{{100}} & \multicolumn{1}{c}{{500}} & \multicolumn{1}{c}{{1000}} & \multicolumn{1}{c}{{3000}} \\ \hline \multicolumn{1}{c|}{Bags} & 
\multicolumn{1}{c}{{67.3}} &
\multicolumn{1}{c}{{$\textbf{68.8}$}} &
\multicolumn{1}{c}{{68.1}} &
\multicolumn{1}{c}{{67.5}} & 
\multicolumn{1}{c}{{67.0}} &
\multicolumn{1}{c}{{66.9}} &
\multicolumn{1}{c}{{66.0}} 
\\ 
\multicolumn{1}{c|}{B/T} &
\multicolumn{1}{c}{{70.1}} &
\multicolumn{1}{c}{{$\textbf{71.1}$}} &
\multicolumn{1}{c}{{70.8}} &
\multicolumn{1}{c}{{69.0}} & 
\multicolumn{1}{c}{{67.9}} &
\multicolumn{1}{c}{{67.2}} &
\multicolumn{1}{c}{{63.3}} 
\\ 
\multicolumn{1}{c|}{Boots} &
\multicolumn{1}{c}{{60.5}} &
\multicolumn{1}{c}{{62.4}} &
\multicolumn{1}{c}{{$\textbf{63.7}$}} &
\multicolumn{1}{c}{{62.8}} & 
\multicolumn{1}{c}{{62.4}} &
\multicolumn{1}{c}{{61.2}} &
\multicolumn{1}{c}{{60.7}} 
\\ 
\multicolumn{1}{c|}{TVs} &
\multicolumn{1}{c}{{61.1}} &
\multicolumn{1}{c}{{70.1}} &
\multicolumn{1}{c}{{$\textbf{71.2}$}} &
\multicolumn{1}{c}{{70.3}} & 
\multicolumn{1}{c}{{70.1}} &
\multicolumn{1}{c}{{69.8}} &
\multicolumn{1}{c}{{69.4}} \\ 
\hline
\end{tabular}} 
\caption{Micro-averaged F1 of \system given $\lambda$}
\label{tb:by_lambda}
\end{table}

\begin{table}[t]
\small
\centering
\vspace{-0.5 pc}
\scalebox{1.0}{
\begin{tabular}{p{2.5cm}|ccccccccc}
\hline \multicolumn{1}{c|}{\multirow{1}{*}{$\mathrm{I}$}} & \multicolumn{1}{c}{{2}} &
\multicolumn{1}{c}{{4}} & \multicolumn{1}{c}{{6}} & \multicolumn{1}{c}{{8}} \\ \hline \multicolumn{1}{c|}{Bags} & \multicolumn{1}{c}{{68.0}} &
\multicolumn{1}{c}{{68.3}} & 
\multicolumn{1}{c}{{\textbf{68.8}}} &
\multicolumn{1}{c}{{68.3}} \\ 
\multicolumn{1}{c|}{B/T} & \multicolumn{1}{c}{{68.5}} &
\multicolumn{1}{c}{{69.1}} & 
\multicolumn{1}{c}{{70.3}} &
\multicolumn{1}{c}{{\textbf{71.9}}} \\ 
\multicolumn{1}{c|}{Boots} & \multicolumn{1}{c}{{61.8}} &
\multicolumn{1}{c}{{62.4}} & 
\multicolumn{1}{c}{{\textbf{64.0}}} &
\multicolumn{1}{c}{{63.9}} \\ 
\multicolumn{1}{c|}{TVs} & \multicolumn{1}{c}{{69.3}} &
\multicolumn{1}{c}{{69.8}} & 
\multicolumn{1}{c}{{\textbf{71.2}}} &
\multicolumn{1}{c}{{70.5}} \\ 
\hline
\end{tabular}}
\caption{Micro-averaged F1 of \system given $\mathrm{I}$.}
\label{tb:by_disnum}
\end{table}

\begin{table}[t]
\small
\centering
\vspace{-0.5 pc}
\scalebox{1.0}{
\begin{tabular}{p{2.5cm}|ccccccccc}
\hline \multicolumn{1}{c|}{\multirow{1}{*}{$\beta$}} & \multicolumn{1}{c}{{0.005}} &
\multicolumn{1}{c}{{0.01}} & \multicolumn{1}{c}{{0.02}} & \multicolumn{1}{c}{{0.05}} \\ 
\hline 
\multicolumn{1}{c|}{Bags} & \multicolumn{1}{c}{{67.8}} &
\multicolumn{1}{c}{{\textbf{68.0}}} & 
\multicolumn{1}{c}{{67.7}} &
\multicolumn{1}{c}{{66.6}} \\ 
\multicolumn{1}{c|}{B/T} & \multicolumn{1}{c}{{70.5}} &
\multicolumn{1}{c}{{71.2}} & 
\multicolumn{1}{c}{{\textbf{71.9}}} & 
\multicolumn{1}{c}{{71.3}} \\ 
\multicolumn{1}{c|}{Boots} & \multicolumn{1}{c}{{63.1}} &
\multicolumn{1}{c}{{\textbf{64.0}}} & 
\multicolumn{1}{c}{{63.7}} & 
\multicolumn{1}{c}{{63.2}} \\ 
\multicolumn{1}{c|}{TVs} & \multicolumn{1}{c}{{68.1}} &
\multicolumn{1}{c}{{69.8}} & 
\multicolumn{1}{c}{{\textbf{70.5}}} & 
\multicolumn{1}{c}{{70.0}} \\ 
\hline
\end{tabular}}
\caption{Micro-averaged F1 of \system given $\beta$}
\label{tb:by_beta}
\end{table}

\begin{table}[t]
\small
\centering
\begin{tabular}{cccccccc}
\hline
\multicolumn{1}{|c|}{\multirow{1}{*}{}} 
& \multicolumn{5}{c|}{Restaurant review domain (En)}
\\
\multicolumn{1}{|l|}{\multirow{1}{*}{ratio r}} &
\multicolumn{1}{c}{10\%}&
\multicolumn{1}{c}{30\%}&
\multicolumn{1}{c}{50\%}&
\multicolumn{1}{c}{70\%}&
\multicolumn{1}{c|}{100\%}\\
\hline \multicolumn{1}{|l|}{{W2V-Gold}} & \multicolumn{1}{c}{{16.3}} &
\multicolumn{1}{c}{{33.0}} &
\multicolumn{1}{c}{{38.6}} &
\multicolumn{1}{c}{{46.7}} &
\multicolumn{1}{c|}{{58.8}}  \\
\multicolumn{1}{|l|}{{BERT-Gold}} & \multicolumn{1}{c}{{24.8}} &
\multicolumn{1}{c}{{36.5}} &
\multicolumn{1}{c}{{48.5}} &
\multicolumn{1}{c}{{55.9}} &
\multicolumn{1}{c|}{{63.1}}   \\
\hline 
\multicolumn{1}{|l|}{{MATE}} &
\multicolumn{1}{c}{{43.8}} &
\multicolumn{1}{c}{{46.7}} &
\multicolumn{1}{c}{{50.4}} &
\multicolumn{1}{c}{{54.3}}  &
\multicolumn{1}{c|}{{60.1}}  \\
\multicolumn{1}{|l|}{{TS-ATT}} & \multicolumn{1}{c}{48.5} &
\multicolumn{1}{c}{{50.6}} &
\multicolumn{1}{c}{{53.2}} &
\multicolumn{1}{c}{{57.7}} &
\multicolumn{1}{c|}{{61.1}}\\
\multicolumn{1}{|l|}{{TS-BT}} & \multicolumn{1}{c}{53.6} &
\multicolumn{1}{c}{{56.1}} &
\multicolumn{1}{c}{{58.4}} &
\multicolumn{1}{c}{{61.2}} &
\multicolumn{1}{c|}{{64.2}}\\
\multicolumn{1}{|l|}{{\system}} &
\multicolumn{1}{c}{\textbf{58.6}} &
\multicolumn{1}{c}{{\textbf{62.2}}} &
\multicolumn{1}{c}{{\textbf{64.1}}} &
\multicolumn{1}{c}{{\textbf{66.9}}} &
\multicolumn{1}{c|}{{\textbf{70.5}}} \\
\hline 
\end{tabular}
\caption{Micro-averaged F1 for 12-class sentence-level aspect detection in restaurant reviews in English with different ratios of training set r.}
\label{tb:by_training_set_eu}
\end{table}



\begin{table}[t]
\small
\centering
\begin{tabular}{cccccccc}
\hline
\multicolumn{1}{|c|}{\multirow{1}{*}{}} 
& \multicolumn{5}{c|}{Restaurant review domain (Sp)}
\\
\multicolumn{1}{|l|}{\multirow{1}{*}{ratio r}} &
\multicolumn{1}{c}{10\%}&
\multicolumn{1}{c}{30\%}&
\multicolumn{1}{c}{50\%}&
\multicolumn{1}{c}{70\%}&
\multicolumn{1}{c|}{100\%}\\
\hline \multicolumn{1}{|l|}{{W2V-Gold}} & \multicolumn{1}{c}{{21.3}} &
\multicolumn{1}{c}{{32.2}} &
\multicolumn{1}{c}{{35.5}} &
\multicolumn{1}{c}{{40.1}} &
\multicolumn{1}{c|}{{50.4}}  \\
\multicolumn{1}{|l|}{{BERT-Gold}} & \multicolumn{1}{c}{{29.2}} &
\multicolumn{1}{c}{{39.1}} &
\multicolumn{1}{c}{{42.2}} &
\multicolumn{1}{c}{{46.0}} &
\multicolumn{1}{c|}{{51.6}}  \\
\hline 
\multicolumn{1}{|l|}{{MATE}} &
\multicolumn{1}{c}{{32.5}} &
\multicolumn{1}{c}{{39.4}} &
\multicolumn{1}{c}{{45.1}} &
\multicolumn{1}{c}{{48.5}}&
\multicolumn{1}{c|}{{52.4}}  \\
\multicolumn{1}{|l|}{{TS-ATT}} & \multicolumn{1}{c}{{42.3}} &
\multicolumn{1}{c}{{46.8}} &
\multicolumn{1}{c}{{54.4}} &
\multicolumn{1}{c}{{58.2}} &
\multicolumn{1}{c|}{{62.3}}   \\
\multicolumn{1}{|l|}{{TS-BT}} & \multicolumn{1}{c}{{44.0}} &
\multicolumn{1}{c}{{46.5}} &
\multicolumn{1}{c}{{52.6}} &
\multicolumn{1}{c}{{60.7}} &
\multicolumn{1}{c|}{{66.6}}  \\
\multicolumn{1}{|l|}{{\system}} &
\multicolumn{1}{c}{{\textbf{66.8}}} &
\multicolumn{1}{c}{{\textbf{68.1}}} &
\multicolumn{1}{c}{{\textbf{70.0}}} &
\multicolumn{1}{c}{{\textbf{72.1}}} &
\multicolumn{1}{c|}{{\textbf{72.5}}} \\
\hline 
\end{tabular}
\caption{Micro-averaged F1 for 12-class sentence-level aspect detection in restaurant reviews in Spanish with different ratios of training set r.}
\label{tb:by_training_set_sp}
\end{table}



\begin{table}[t]
\small
\centering
\begin{tabular}{cccccccc}
\hline
\multicolumn{1}{|c|}{\multirow{1}{*}{}} 
& \multicolumn{5}{c|}{Restaurant review domain (Fr)}
\\
\multicolumn{1}{|l|}{\multirow{1}{*}{ratio r}} &
\multicolumn{1}{c}{10\%}&
\multicolumn{1}{c}{30\%}&
\multicolumn{1}{c}{50\%}&
\multicolumn{1}{c}{70\%}&
\multicolumn{1}{c|}{100\%}\\
\hline \multicolumn{1}{|l|}{{W2V-Gold}} & \multicolumn{1}{c}{{21.2}} &
\multicolumn{1}{c}{{29.9}} &
\multicolumn{1}{c}{{37.1}} &
\multicolumn{1}{c}{{43.1}} &
\multicolumn{1}{c|}{{50.4}}  \\
\multicolumn{1}{|l|}{{BERT-Gold}} & \multicolumn{1}{c}{{20.2}} &
\multicolumn{1}{c}{{24.8}} &
\multicolumn{1}{c}{{33.0}} &
\multicolumn{1}{c}{{40.9}} &
\multicolumn{1}{c|}{{50.6}}  \\
\hline 
\multicolumn{1}{|l|}{{MATE}} &
\multicolumn{1}{c}{{28.7}} &
\multicolumn{1}{c}{{34.8}} &
\multicolumn{1}{c}{{40.5}} &
\multicolumn{1}{c}{{45.2}} &
\multicolumn{1}{c|}{{48.1}}  \\
\multicolumn{1}{|l|}{{TS-ATT}} & \multicolumn{1}{c}{{32.8}} &
\multicolumn{1}{c}{{38.1}} &
\multicolumn{1}{c}{{44.1}} &
\multicolumn{1}{c}{{46.6}} &
\multicolumn{1}{c|}{{50.1}}   \\
\multicolumn{1}{|l|}{{TS-BT}} & \multicolumn{1}{c}{{43.0}} &
\multicolumn{1}{c}{{44.9}} &
\multicolumn{1}{c}{{46.5}} &
\multicolumn{1}{c}{{48.0}} &
\multicolumn{1}{c|}{{53.0}}  \\
\multicolumn{1}{|l|}{{\system}} &
\multicolumn{1}{c}{{\textbf{48.7}}} &
\multicolumn{1}{c}{{\textbf{51.8}}} &
\multicolumn{1}{c}{{\textbf{54.9}}} &
\multicolumn{1}{c}{{\textbf{60.8}}} &
\multicolumn{1}{c|}{{\textbf{65.4}}} \\
\hline 
\end{tabular}
\caption{Micro-averaged F1 for 12-class sentence-level aspect detection in restaurant reviews in French with different ratios of training set r.}
\label{tb:by_training_set_fr}
\end{table}


\begin{table}[t]
\small
\centering
\begin{tabular}{cccccccc}
\hline
\multicolumn{1}{|c|}{\multirow{1}{*}{}} 
& \multicolumn{5}{c|}{Restaurant review domain (Ru)}
\\
\multicolumn{1}{|l|}{\multirow{1}{*}{ratio r}} &
\multicolumn{1}{c}{10\%}&
\multicolumn{1}{c}{30\%}&
\multicolumn{1}{c}{50\%}&
\multicolumn{1}{c}{70\%}&
\multicolumn{1}{c|}{100\%}\\
\hline \multicolumn{1}{|l|}{{W2V-Gold}} & \multicolumn{1}{c}{{28.5}} &
\multicolumn{1}{c}{{32.5}} &
\multicolumn{1}{c}{{43.5}} &
\multicolumn{1}{c}{{50.0}} &
\multicolumn{1}{c|}{{55.7}}  \\
\multicolumn{1}{|l|}{{BERT-Gold}} & \multicolumn{1}{c}{{23.5}} &
\multicolumn{1}{c}{{31.5}} &
\multicolumn{1}{c}{{41.3}} &
\multicolumn{1}{c}{{47.9}} &
\multicolumn{1}{c|}{{55.3}}  \\
\hline 

\multicolumn{1}{|l|}{{MATE}} &
\multicolumn{1}{c}{{22.6}} &
\multicolumn{1}{c}{{30.3}} &
\multicolumn{1}{c}{{36.8}} &
\multicolumn{1}{c}{{44.3}} &
\multicolumn{1}{c|}{{51.2}}  \\
\multicolumn{1}{|l|}{{TS-ATT}} & \multicolumn{1}{c}{{58.8}} &
\multicolumn{1}{c}{{59.1}} &
\multicolumn{1}{c}{{59.8}} &
\multicolumn{1}{c}{{62.1}} &
\multicolumn{1}{c|}{{65.5}}   \\
\multicolumn{1}{|l|}{{TS-BT}} & \multicolumn{1}{c}{{59.5}} &
\multicolumn{1}{c}{{61.3}} &
\multicolumn{1}{c}{{62.1}} &
\multicolumn{1}{c}{{65.5}} &
\multicolumn{1}{c|}{{67.4}}  \\
\multicolumn{1}{|l|}{{\system}} &
\multicolumn{1}{c}{{\textbf{61.3}}} &
\multicolumn{1}{c}{{\textbf{65.0}}} &
\multicolumn{1}{c}{{\textbf{67.8}}} &
\multicolumn{1}{c}{{\textbf{71.5}}} &
\multicolumn{1}{c|}{{\textbf{76.8}}} \\
\hline 
\end{tabular}
\caption{Micro-averaged F1 for 12-class sentence-level aspect detection in restaurant reviews in Russian with different ratios of training set r.}
\label{tb:by_training_set_ru}
\end{table}



\begin{table}[t]
\small
\centering
\begin{tabular}{cccccccc}
\hline
\multicolumn{1}{|c|}{\multirow{1}{*}{}} 
& \multicolumn{5}{c|}{Restaurant review domain (Du)}
\\
\multicolumn{1}{|l|}{\multirow{1}{*}{ratio r}} &
\multicolumn{1}{c}{10\%}&
\multicolumn{1}{c}{30\%}&
\multicolumn{1}{c}{50\%}&
\multicolumn{1}{c}{70\%}&
\multicolumn{1}{c|}{100\%}\\
\hline \multicolumn{1}{|l|}{{W2V-Gold}} & \multicolumn{1}{c}{{24.4}} &
\multicolumn{1}{c}{{32.8}} &
\multicolumn{1}{c}{{42.4}} &
\multicolumn{1}{c}{{47.3}} &
\multicolumn{1}{c|}{{51.4}}  \\
\multicolumn{1}{|l|}{{BERT-Gold}} & \multicolumn{1}{c}{{28.1}} &
\multicolumn{1}{c}{{40.8}} &
\multicolumn{1}{c}{{47.5}} &
\multicolumn{1}{c}{{51.6}} &
\multicolumn{1}{c|}{{53.5}}  \\
\hline 
\multicolumn{1}{|l|}{{MATE}} &
\multicolumn{1}{c}{{38.8}} &
\multicolumn{1}{c}{{40.9}} &
\multicolumn{1}{c}{{46.2}} &
\multicolumn{1}{c}{{48.4}} &
\multicolumn{1}{c|}{{52.0}}  \\
\multicolumn{1}{|l|}{{TS-ATT}} & \multicolumn{1}{c}{{43.9}} &
\multicolumn{1}{c}{{45.8}} &
\multicolumn{1}{c}{{51.8}} &
\multicolumn{1}{c}{{53.5}} &
\multicolumn{1}{c|}{{55.4}}   \\
\multicolumn{1}{|l|}{{TS-BT}} & \multicolumn{1}{c}{{45.4}} &
\multicolumn{1}{c}{{47.1}} &
\multicolumn{1}{c}{{50.4}} &
\multicolumn{1}{c}{{53.6}} &
\multicolumn{1}{c|}{{57.6}}  \\
\multicolumn{1}{|l|}{{\system}} &
\multicolumn{1}{c}{{\textbf{58.5}}} &
\multicolumn{1}{c}{{\textbf{62.5}}} &
\multicolumn{1}{c}{{\textbf{68.3}}} &
\multicolumn{1}{c}{{\textbf{72.1}}} &
\multicolumn{1}{c|}{{\textbf{73.8}}} \\
\hline 
\end{tabular}
\caption{Micro-averaged F1 for 12-class sentence-level aspect detection in restaurant reviews in Dutch with different ratios of training set r.}
\label{tb:by_training_set_du}
\end{table}


\begin{table}[t]
\small
\centering
\begin{tabular}{cccccccc}
\hline
\multicolumn{1}{|c|}{\multirow{1}{*}{}} 
& \multicolumn{5}{c|}{Restaurant review domain (Tur)}
\\
\multicolumn{1}{|l|}{\multirow{1}{*}{ratio r}} &
\multicolumn{1}{c}{10\%}&
\multicolumn{1}{c}{30\%}&
\multicolumn{1}{c}{50\%}&
\multicolumn{1}{c}{70\%}&
\multicolumn{1}{c|}{100\%}\\
\hline \multicolumn{1}{|l|}{{W2V-Gold}} & \multicolumn{1}{c}{{28.6}} &
\multicolumn{1}{c}{{37.2}} &
\multicolumn{1}{c}{{42.3}} &
\multicolumn{1}{c}{{50.8}} &
\multicolumn{1}{c|}{{55.7}}  \\
\multicolumn{1}{|l|}{{BERT-Gold}} & \multicolumn{1}{c}{{31.5}} &
\multicolumn{1}{c}{{39.0}} &
\multicolumn{1}{c}{{45.6}} &
\multicolumn{1}{c}{{52.3}} &
\multicolumn{1}{c|}{{56.5}}  \\
\hline 
\multicolumn{1}{|l|}{{MATE}} &
\multicolumn{1}{c}{{41.3}} &
\multicolumn{1}{c}{{44.9}} &
\multicolumn{1}{c}{{47.1}} &
\multicolumn{1}{c}{{49.9}} &
\multicolumn{1}{c|}{{53.0}}  \\
\multicolumn{1}{|l|}{{TS-ATT}} & \multicolumn{1}{c}{{45.9}} &
\multicolumn{1}{c}{{47.5}} &
\multicolumn{1}{c}{{49.7}} &
\multicolumn{1}{c}{{52.8}} &
\multicolumn{1}{c|}{{55.5}}   \\
\multicolumn{1}{|l|}{{TS-BT}} & \multicolumn{1}{c}{{45.5}} &
\multicolumn{1}{c}{{48.7}} &
\multicolumn{1}{c}{{52.6}} &
\multicolumn{1}{c}{{54.3}} &
\multicolumn{1}{c|}{{57.6}}  \\
\multicolumn{1}{|l|}{{\system}} &
\multicolumn{1}{c}{{\textbf{49.8}}} &
\multicolumn{1}{c}{{\textbf{52.4}}} &
\multicolumn{1}{c}{{\textbf{56.9}}} &
\multicolumn{1}{c}{{\textbf{60.1}}} &
\multicolumn{1}{c|}{{\textbf{65.4}}} \\
\hline 
\end{tabular}
\caption{Micro-averaged F1 for 12-class sentence-level aspect detection in restaurant reviews in Turkish with different ratios of training set r.}
\label{tb:by_training_set_tur}
\end{table}



\begin{table}[t]
\small
\centering
\vspace{-0.5 pc}
\scalebox{1.0}{
\begin{tabular}{p{2.5cm}|ccccccccc}
\hline \multicolumn{1}{c|}{\multirow{1}{*}{\# of seed words}} & \multicolumn{1}{c}{{0}}
& \multicolumn{1}{c}{{5}} &
\multicolumn{1}{c}{{15}} & \multicolumn{1}{c}{{20}} & \multicolumn{1}{c}{{30}} \\ \hline \multicolumn{1}{c|}{Bags} & 
\multicolumn{1}{c}{{41.2}} &
\multicolumn{1}{c}{{61.7}} &
\multicolumn{1}{c}{{63.6}} & 
\multicolumn{1}{c}{{67.2}} &
\multicolumn{1}{c}{{68.8}} 
\\ 
\multicolumn{1}{c|}{KBs} &
\multicolumn{1}{c}{{33.2}} &
\multicolumn{1}{c}{{65.2}} &
\multicolumn{1}{c}{{68.2}} & 
\multicolumn{1}{c}{{69.4}} &
\multicolumn{1}{c}{{72.2}}
\\ 
\multicolumn{1}{c|}{B/T} &
\multicolumn{1}{c}{{42.3}} &
\multicolumn{1}{c}{{70.2}} &
\multicolumn{1}{c}{{70.5}} & 
\multicolumn{1}{c}{{71.9}} &
\multicolumn{1}{c}{{72.0}} 
\\ 
\multicolumn{1}{c|}{Boots} &
\multicolumn{1}{c}{{37.2}} &
\multicolumn{1}{c}{{61.0}} &
\multicolumn{1}{c}{{63.5}} & 
\multicolumn{1}{c}{{63.2}} &
\multicolumn{1}{c}{{64.0}}
\\ 
\multicolumn{1}{c|}{TV} &
\multicolumn{1}{c}{{45.1}} &
\multicolumn{1}{c}{{66.3}} &
\multicolumn{1}{c}{{65.7}} & 
\multicolumn{1}{c}{{68.2}} &
\multicolumn{1}{c}{{71.2}} 
\\
\multicolumn{1}{c|}{VCs} &
\multicolumn{1}{c}{{40.2}} &
\multicolumn{1}{c}{{59.6}} &
\multicolumn{1}{c}{{61.2}} & 
\multicolumn{1}{c}{{66.1}} &
\multicolumn{1}{c}{{66.9}} 
\\ 
\hline
\end{tabular}}
\caption{Micro-averaged F1 of \system given \# of seed words}
\label{tb:by_lambda}
\end{table}

\begin{table}[t]
\small
\centering
\vspace{-0.5 pc}
\scalebox{1.0}{
\begin{tabular}{p{2.5cm}|ccccccccc}
\hline \multicolumn{1}{c|}{\multirow{1}{*}{ratio of $J_{d_1}$}} & \multicolumn{1}{c}{{0.5}} & \multicolumn{1}{c}{{1}} &
\multicolumn{1}{c}{{5}} & \multicolumn{1}{c}{{10}} & \multicolumn{1}{c}{{100}} \\ \hline \multicolumn{1}{c|}{Bags} & 
\multicolumn{1}{c}{{67.9}} & 
\multicolumn{1}{c}{{68.8}}  &
\multicolumn{1}{c}{{68.1}} &
\multicolumn{1}{c}{{66.9}} & 
\multicolumn{1}{c}{{66.1}} 
\\ 
\multicolumn{1}{c|}{KBs} &
\multicolumn{1}{c}{{71.9}} & 
\multicolumn{1}{c}{{72.2}} &
\multicolumn{1}{c}{{71.3}} &
\multicolumn{1}{c}{{71.0}} & 
\multicolumn{1}{c}{{70.6}} 
\\ 
\multicolumn{1}{c|}{B/T} &
\multicolumn{1}{c}{{72.2}} &
\multicolumn{1}{c}{{72.0}} &
\multicolumn{1}{c}{{71.7}} & 
\multicolumn{1}{c}{{71.2}} & 
\multicolumn{1}{c}{{71.0}} 
\\ 
\multicolumn{1}{c|}{Boots} &
\multicolumn{1}{c}{{64.8}} & 
\multicolumn{1}{c}{{64.0}} &
\multicolumn{1}{c}{{64.1}} &
\multicolumn{1}{c}{{63.3}} & 
\multicolumn{1}{c}{{61.5}} 
\\ 
\multicolumn{1}{c|}{TV} &
\multicolumn{1}{c}{{69.2}} & 
\multicolumn{1}{c}{{71.2}} &
\multicolumn{1}{c}{{70.3}} &
\multicolumn{1}{c}{{69.5}} & 
\multicolumn{1}{c}{{68.9}} 
\\
\multicolumn{1}{c|}{VCs} &
\multicolumn{1}{c}{{65.0}} & 
\multicolumn{1}{c}{{66.9}} &
\multicolumn{1}{c}{{66.7}} &
\multicolumn{1}{c}{{64.2}} & 
\multicolumn{1}{c}{{63.9}}  
\\ 
\hline
\end{tabular}}
\caption{Micro-averaged F1 of \system given ratio of $J_{d_1}$}
\label{tb:by_lambdajd1}
\end{table}

\begin{table}[t]
\small
\centering
\vspace{-0.5 pc}
\scalebox{1.0}{
\begin{tabular}{p{2.5cm}|ccccccccc}
\hline \multicolumn{1}{c|}{\multirow{1}{*}{ratio of $J_{d_2}$}} & \multicolumn{1}{c}{{0.5}} & \multicolumn{1}{c}{{1}} &
\multicolumn{1}{c}{{5}} & \multicolumn{1}{c}{{10}} & \multicolumn{1}{c}{{100}} \\ \hline \multicolumn{1}{c|}{Bags} & 
\multicolumn{1}{c}{{67.9}} & 
\multicolumn{1}{c}{{68.8}} &
\multicolumn{1}{c}{{68.1}} &
\multicolumn{1}{c}{{67.8}} & 
\multicolumn{1}{c}{{67.2}} 
\\ 
\multicolumn{1}{c|}{KBs} &
\multicolumn{1}{c}{{71.0}} & 
\multicolumn{1}{c}{{72.2}} &
\multicolumn{1}{c}{{71.9}} &
\multicolumn{1}{c}{{71.0}} & 
\multicolumn{1}{c}{{71.3}} 
\\ 
\multicolumn{1}{c|}{B/T} &
\multicolumn{1}{c}{{72.2}}& 
\multicolumn{1}{c}{{72.0}}  &
\multicolumn{1}{c}{{72.0}} &
\multicolumn{1}{c}{{71.3}} & 
\multicolumn{1}{c}{{70.9}} 
\\ 
\multicolumn{1}{c|}{Boots} &
\multicolumn{1}{c}{{64.8}} & 
\multicolumn{1}{c}{{64.0}} &
\multicolumn{1}{c}{{63.5}} &
\multicolumn{1}{c}{{62.3}} & 
\multicolumn{1}{c}{{62.5}} 
\\ 
\multicolumn{1}{c|}{TV} &
\multicolumn{1}{c}{{69.2}} & 
\multicolumn{1}{c}{{71.2}} &
\multicolumn{1}{c}{{69.2}} &
\multicolumn{1}{c}{{68.5}} & 
\multicolumn{1}{c}{{68.7}} 
\\
\multicolumn{1}{c|}{VCs} &
\multicolumn{1}{c}{{65.0}} & 
\multicolumn{1}{c}{{66.9}} &
\multicolumn{1}{c}{{65.8}} &
\multicolumn{1}{c}{{65.3}} & 
\multicolumn{1}{c}{{64.1}}  
\\ 
\hline
\end{tabular}}
\caption{Micro-averaged F1 of \system given ratio of $J_{d_2}$}
\label{tb:by_lambdajd2}
\end{table}

\begin{table}[t]
\small
\centering
\vspace{-0.5 pc}
\scalebox{1.0}{
\begin{tabular}{p{2.5cm}|ccccccccc}
\hline \multicolumn{1}{c|}{\multirow{1}{*}{ratio of $J_{d_3}$}} & \multicolumn{1}{c}{{0.5}} & \multicolumn{1}{c}{{1}} &
\multicolumn{1}{c}{{5}} & \multicolumn{1}{c}{{10}} & \multicolumn{1}{c}{{100}} \\ \hline \multicolumn{1}{c|}{Bags} & 
\multicolumn{1}{c}{{66.9}} & 
\multicolumn{1}{c}{{68.8}} &
\multicolumn{1}{c}{{67.2}} &
\multicolumn{1}{c}{{67.8}} & 
\multicolumn{1}{c}{{67.1}} 
\\ 
\multicolumn{1}{c|}{KBs} &
\multicolumn{1}{c}{{71.0}} & 
\multicolumn{1}{c}{{72.2}} &
\multicolumn{1}{c}{{71.4}} &
\multicolumn{1}{c}{{71.0}} & 
\multicolumn{1}{c}{{71.4}} 
\\ 
\multicolumn{1}{c|}{B/T} &
\multicolumn{1}{c}{{71.9}} & 
\multicolumn{1}{c}{{72.0}} &
\multicolumn{1}{c}{{72.2}} &
\multicolumn{1}{c}{{71.8}} & 
\multicolumn{1}{c}{{71.2}} 
\\ 
\multicolumn{1}{c|}{Boots} &
\multicolumn{1}{c}{{63.5}} & 
\multicolumn{1}{c}{{64.0}} &
\multicolumn{1}{c}{{64.8}} &
\multicolumn{1}{c}{{64.1}} & 
\multicolumn{1}{c}{{63.2}} 
\\ 
\multicolumn{1}{c|}{TV} &
\multicolumn{1}{c}{{70.2}} & 
\multicolumn{1}{c}{{71.2}} &
\multicolumn{1}{c}{{69.2}} &
\multicolumn{1}{c}{{69.1}} & 
\multicolumn{1}{c}{{68.6}} 
\\
\multicolumn{1}{c|}{VCs} &
\multicolumn{1}{c}{{65.1}} & 
\multicolumn{1}{c}{{66.9}} &
\multicolumn{1}{c}{{65.0}} &
\multicolumn{1}{c}{{64.5}} & 
\multicolumn{1}{c}{{63.2}}  
\\ 
\hline
\end{tabular}}
\caption{Micro-averaged F1 of \system given ratio of $J_{d_3}$}
\label{tb:by_lambdajd3}
\end{table}

\subsection{Parameter Sensitivity Analysis} 

\label{sec:parameter_senetivity}

In this section, we provide more results for parameter sensitivity. Table~\ref{tb:by_gml_soft_tem} shows effects of the grumbel-softmax temperature $\tau$ on the performance of \system. We find that our model achieves the best results when small $\tau$ is set, suggesting that it is important to not to mix the latent semantic when predicting the segment's aspect. Table~\ref{tb:by_lambda} shows effects of the  distillation objective ratio $\lambda$ on the performance of \system. It is found that \system achieves the best performance when $\lambda$ is set to 5 or 10. From Table~\ref{tb:by_disnum}, we see effects of number of latent semantics $\mathrm{I}$ and \system achieves best result when large $\mathrm{I}$ is set. From Table~\ref{tb:by_beta}, we see effects of $\beta$ and \system achieves the best result when $\beta$ is set to 0.01 or 0.02.

Besides, we compare the \system performance when the number of seed words change. The result in Table\ref{tb:by_lambda} does show that seed words have impact on the performance. When the number of seed words increases, \system’s performance increases, while 20 seed words are satisfactory for \system to infer the segment’s aspects.

As to balances between different loss functions, we investigate HDAE’s performance according to the ratio of $J_{d_1}$, $J_{d_2}$, and $J_{d_3}$ objectives, as shown in Table~\ref{tb:by_lambdajd1}, \ref{tb:by_lambdajd2}, and \ref{tb:by_lambdajd3}. The result shows that HDAE achieves the best when small ratios ( $\leq$ 1)  are set, perhaps due to the strong regularization on each latent semantic meaning coming with the high ratios.


\subsection{Results in ground truth  aspect labels ratios} 

\label{sec:aspect_labels_ratios}

In this section, we provide results for baselines with (\system, MATE, TS-ATT, TS-BT) or without (W2V, BERT) leveraging seed word in different ground truth aspect labels ratios. We find the proposed model \system can achieve best performance in different ground truth aspect labels ratios, suggesting the effectiveness of purposed hyperbolic disentangled based method. Besides, we notice that in the low aspect labels data ratios, seed word based approaches (MATE, TS-ATT, TS-BT, and \system) can achieve better, demonstrating that seed words can give useful guidance and assist models to improve aspect inference ability.



\begin{table*}[t]
\fontsize{9}{10}\selectfont
\center
\small
\scalebox{1.0}{
\noindent\begin{tabular}{|c|c|c|c|c|}
\hline 
\multicolumn{2}{|l}{\multirow{1}{*}{Do not purchase.}} &
\multicolumn{1}{c|}{GT: Noise} 
\\
\multicolumn{3}{|l|}{\multirow{1}{*}{Seed Words: loud, noise, noisy, quiet, action, sound, quieter, know, make}} 
\\
\hline
\multicolumn{1}{|c|}{\system: General \color{red}\XSolidBrush} &
\multicolumn{1}{|c|}{{MATE: General \color{red}\XSolidBrush}} &
\multicolumn{1}{c|}{{TS-W2V: General \color{red}\XSolidBrush}} 
\\
\hline 
\multicolumn{2}{|l}{\multirow{1}{*}{Which died.}} &
\multicolumn{1}{c|}{GT: General}
\\
\multicolumn{3}{|l|}{\multirow{1}{*}{Seed Words: think, recommend, purchase, using, unit, star, microsoft, mouse}}
\\
\hline
\multicolumn{1}{|c|}{\system: Build Quality \color{red}\XSolidBrush} &
\multicolumn{1}{|c|}{{MATE: Build Quality \color{red}\XSolidBrush}} &
\multicolumn{1}{c|}{{TS-W2V: Build Quality \color{red}\XSolidBrush}} 
\\
\hline 
\multicolumn{2}{|l}{\multirow{1}{*}{Except the keyboard was one of those high keyed, clackety-clunkety types.}} &
\multicolumn{1}{c|}{GT: General} 
\\
\multicolumn{3}{|l|}{\multirow{1}{*}{Seed Words: think, recommend, purchase, using, unit, star, microsoft, mouse}}
\\
\hline
\multicolumn{1}{|c|}{\system: General \color{blue} \Checkmark} &
\multicolumn{1}{|c|}{{MATE: General \color{blue} \Checkmark}} &
\multicolumn{1}{c|}{{TS-W2V: Comfort \color{red}\XSolidBrush}} 
\\
\hline 
\multicolumn{2}{|l}{\multirow{1}{*}{I really liked the look of it.}} &
\multicolumn{1}{c|}{GT: Looks} 
\\
\multicolumn{3}{|l|}{\multirow{1}{*}{Seed Words: look, slim, original, appearance, little, attractive, beautiful}}
\\
\hline
\multicolumn{1}{|c|}{\system: Looks \color{blue} \Checkmark} &
\multicolumn{1}{|c|}{{MATE: General \color{red}\XSolidBrush}} &
\multicolumn{1}{c|}{{TS-W2V: Looks \color{blue} \Checkmark}} 
\\
\hline 
\multicolumn{2}{|l}{\multirow{1}{*}{I liked the feel of the keys.}} &
\multicolumn{1}{c|}{GT: Comfort} 
\\
\multicolumn{3}{|l|}{\multirow{1}{*}{Seed Words: feel, comfortable, mushy, key, like, keyboard, good, perfect}} 
\\
\hline
\multicolumn{1}{|c|}{\system: Comfort \color{blue} \Checkmark} &
\multicolumn{1}{|c|}{{MATE: General \color{red}\XSolidBrush}} &
\multicolumn{1}{c|}{{TS-W2V: Looks \color{red}\XSolidBrush}} 
\\
\hline 
\multicolumn{2}{|l}{\multirow{1}{*}{But it has all the buttons to interface with my iMac.}} &
\multicolumn{1}{c|}{GT: Extra functionality} 
\\
\multicolumn{3}{|l|}{\multirow{1}{*}{Seed Words: buttons, light, pencil, volume, power, feature, bright, mute, handy, low, dark}} 
\\
\hline
\multicolumn{1}{|c|}{\system: Extra functionality \color{blue} \Checkmark} &
\multicolumn{1}{|c|}{{MATE: Extra functionality \color{blue} \Checkmark}} &
\multicolumn{1}{c|}{{TS-W2V: Extra functionality \color{blue} \Checkmark}} 
\\
\hline 
\multicolumn{2}{|l}{\multirow{1}{*}{It is quiet.}} &
\multicolumn{1}{c|}{GT: Noise} 
\\
\multicolumn{3}{|l|}{\multirow{1}{*}{Seed Words: loud, noise, noisy, red, action, sound, quieter, know, make}} 
\\
\hline
\multicolumn{1}{|c|}{\system: Noise \color{blue} \Checkmark} &
\multicolumn{1}{|c|}{{MATE: Noise \color{blue} \Checkmark}} &
\multicolumn{1}{c|}{{TS-W2V: Noise \color{blue} \Checkmark}} 
\\
\hline 
\multicolumn{2}{|l}{\multirow{1}{*}{The layout of the keys makes it difficult for me to use, with keys like the backspace.}} &
\multicolumn{1}{c|}{GT: Layout} 
\\
\multicolumn{3}{|l|}{\multirow{1}{*}{Seed Words: key, delete, backspace, size, layout, end, insert, home, bar, perfect, space}} 
\\
\hline
\multicolumn{1}{|c|}{\system: Layout \color{blue} \Checkmark} &
\multicolumn{1}{|c|}{{MATE: General \color{red}\XSolidBrush}} &
\multicolumn{1}{c|}{{TS-W2V: Layout \color{blue} \Checkmark}} 
\\
\hline 
\multicolumn{2}{|l}{\multirow{1}{*}{And doesn't {\color{red}depress} at times}} &
\multicolumn{1}{c|}{GT: Build Quality} 
\\
\multicolumn{3}{|l|}{\multirow{1}{*}{Seed Words: {\color{red}working}, build, {\color{red}stopped}, quality, month, spacebar, {\color{red}stuck}, left, plastic, kind, died}} 
\\
\hline
\multicolumn{1}{|c|}{\system: Build Quality \color{blue} \Checkmark} &
\multicolumn{1}{|c|}{{MATE: Build Quality \color{blue} \Checkmark}} &
\multicolumn{1}{c|}{{TS-W2V: General \color{red}\XSolidBrush}} 
\\
\hline 
\multicolumn{2}{|l}{\multirow{1}{*}{And the key for the `` t '' is already {\color{red}broken}.}} &
\multicolumn{1}{c|}{GT: Build Quality} 
\\
\multicolumn{3}{|l|}{\multirow{1}{*}{Seed Words: {\color{red}working}, build, {\color{red}stopped}, quality, month, spacebar, stuck, left, plastic, kind, died}} 
\\
\hline
\multicolumn{1}{|c|}{\system: Build Quality \color{blue} \Checkmark} &
\multicolumn{1}{|c|}{{MATE: Layout \color{red}\XSolidBrush}} &
\multicolumn{1}{c|}{{TS-W2V: Comfort \color{red}\XSolidBrush}} 
\\
\hline 
\multicolumn{2}{|l}{\multirow{1}{*}{Has a top row of quick {\color{red}link}}} &
\multicolumn{1}{c|}{GT: Extra functionality} 
\\
\multicolumn{3}{|l|}{\multirow{1}{*}{Seed Words: {\color{red}button}, light, pencil, volume, power, feature, bright, mute, handy, low, dark}} 
\\
\hline
\multicolumn{1}{|c|}{\system: Extra functionality \color{blue} \Checkmark} &
\multicolumn{1}{|c|}{{MATE: Extra functionality \color{blue} \Checkmark}} &
\multicolumn{1}{c|}{{TS-W2V: Comfort \color{red}\XSolidBrush}} 
\\
\hline 
\multicolumn{2}{|l}{\multirow{1}{*}{The keyboard is {\color{red}sleek} and visually  {\color{red}appealing}.}} &
\multicolumn{1}{c|}{GT: Looks} 
\\
\multicolumn{3}{|l|}{\multirow{1}{*}{Seed Words: look, slim, original, {\color{red}appearance}, little, attractive, {\color{red}beautiful}}}
\\
\hline
\multicolumn{1}{|c|}{\system: Looks \color{blue} \Checkmark} &
\multicolumn{1}{|c|}{{MATE: General \color{red}\XSolidBrush}} &
\multicolumn{1}{c|}{{TS-W2V: General \color{red}\XSolidBrush}}
\\
\hline 
\multicolumn{2}{|l}{\multirow{1}{*}{That had {\color{red}popped off}.}} &
\multicolumn{1}{c|}{GT: Build Quality} 
\\
\multicolumn{3}{|l|}{\multirow{1}{*}{Seed Words: working, build, {\color{red}stopped}, quality, month, spacebar, stuck, left, plastic, kind, {\color{red}died}}} 
\\
\hline
\multicolumn{1}{|c|}{\system: Build Quality \color{blue} \Checkmark} &
\multicolumn{1}{|c|}{{MATE: Build Quality \color{blue} \Checkmark}} &
\multicolumn{1}{c|}{{TS-W2V: General \color{red}\XSolidBrush}} 
\\
\hline 
\multicolumn{2}{|l}{\multirow{1}{*}{It is very {\color{red}responsive}.}} &
\multicolumn{1}{c|}{GT: Comfort} 
\\
\multicolumn{3}{|l|}{\multirow{1}{*}{Seed Words: {\color{red}feel}, comfortable, mushy, key, like, keyboard, {\color{red}good}, perfect, press, wrist, {\color{red}action}, shallow, smooth}} 
\\
\hline
\multicolumn{1}{|c|}{\system: Comfort \color{blue} \Checkmark} &
\multicolumn{1}{|c|}{{MATE: Connectivity \color{red}\XSolidBrush}} &
\multicolumn{1}{c|}{{TS-W2V: General \color{red}\XSolidBrush}} 
\\
\hline 
\end{tabular}}
\caption{Comparison of predictions on sample Keyboards product review segments between \system, MATE, and TS-W2V. For each review segment, the ground true (GT) aspect and its corresponding seed words are provided.}
\vspace{-1.5 pc}
\label{tb:case_st_kb}
\end{table*}


\begin{table*}[t]
\fontsize{9}{10}
\selectfont
\center
\small
\scalebox{1.0}{
\noindent\begin{tabular}{|c|c|c|c|c|}
\hline 
\multicolumn{2}{|l}{\multirow{1}{*}{But may be unsightly.}} &
\multicolumn{1}{c|}{GT: Look} 
\\
\multicolumn{3}{|l|}{\multirow{1}{*}{Seed Words: cute, look, looked, great, fringe, great, fringe, style, color, love, design, going, unattractive, attractive}}
\\
\hline
\multicolumn{1}{|c|}{\system: Durability \color{red}\XSolidBrush} &
\multicolumn{1}{|c|}{{MATE: Size \color{red}\XSolidBrush}} &
\multicolumn{1}{c|}{{TS-W2V: General \color{red}\XSolidBrush}} 
\\
\hline 
\multicolumn{2}{|l}{\multirow{1}{*}{However, I bought them in February.}} &
\multicolumn{1}{c|}{GT: General} 
\\
\multicolumn{3}{|l|}{\multirow{1}{*}{Seed Words: return, pair, thought, year, boot, high, kind, merrell, excited, wore, disappointed, buy, problem}}
\\
\hline
\multicolumn{1}{|c|}{\system: General \color{blue} \Checkmark} &
\multicolumn{1}{|c|}{{MATE: Price \color{red}\XSolidBrush}} &
\multicolumn{1}{c|}{{TS-W2V: General \color{blue} \Checkmark}} 
\\
\hline 
\multicolumn{2}{|l}{\multirow{1}{*}{I ordered these a half-size up}} &
\multicolumn{1}{c|}{GT: Size} 
\\
\multicolumn{3}{|l|}{\multirow{1}{*}{Seed Words: size, ordered, half, order, big, little, bigger, usually, narrow, normally, width}}
\\
\hline
\multicolumn{1}{|c|}{\system: Size \color{blue} \Checkmark} &
\multicolumn{1}{|c|}{{MATE: General \color{red}\XSolidBrush}} &
\multicolumn{1}{c|}{{TS-W2V: Size \color{blue} \Checkmark}} 
\\
\hline 
\multicolumn{2}{|l}{\multirow{1}{*}{Nice looking shoes.}} &
\multicolumn{1}{c|}{GT: Look} 
\\
\multicolumn{3}{|l|}{\multirow{1}{*}{Seed Words: cute, look, looked, great, fringe, great, fringe, style, color, love, design, going, unattractive, attractive}}
\\
\hline
\multicolumn{1}{|c|}{\system: Look \color{blue} \Checkmark} &
\multicolumn{1}{|c|}{{MATE: Look \color{blue} \Checkmark}} &
\multicolumn{1}{c|}{{TS-W2V: Color \color{red}\XSolidBrush}} 
\\
\hline 
\multicolumn{2}{|l}{\multirow{1}{*}{Hurt after 30 minutes}} &
\multicolumn{1}{c|}{GT: Comfort} 
\\
\multicolumn{3}{|l|}{\multirow{1}{*}{Seed Words: comfortable, fit, foot, hurt, ankle, comfy, blister, comfort, break, wear, able, extreme, walking, sock}}
\\
\hline
\multicolumn{1}{|c|}{\system: Comfort \color{blue} \Checkmark} &
\multicolumn{1}{|c|}{{MATE: Comfort \color{blue} \Checkmark}} &
\multicolumn{1}{c|}{{TS-W2V: Comfort \color{blue} \Checkmark}} 
\\
\hline 
\multicolumn{2}{|l}{\multirow{1}{*}{The material is very thin and flimsy}} &
\multicolumn{1}{c|}{GT: Materials} 
\\
\multicolumn{3}{|l|}{\multirow{1}{*}{Seed Words: leather, inside, fringe, material, heel, cheaper, soft, sole, buckle, suede, true, rubber, advertised}}
\\
\hline
\multicolumn{1}{|c|}{\system: Materials \color{blue} \Checkmark} &
\multicolumn{1}{|c|}{{MATE: Materials \color{blue} \Checkmark}} &
\multicolumn{1}{c|}{{TS-W2V: Materials \color{blue} \Checkmark}} 
\\
\hline 
\multicolumn{2}{|l}{\multirow{1}{*}{Which helps in the Maine {\color{red}winter}.}} &
\multicolumn{1}{c|}{GT: Weather resistance} 
\\
\multicolumn{3}{|l|}{\multirow{1}{*}{Seed Words: dry, waterproof, rain, wet, water, foot, soaked, {\color{red}snow}, {\color{red}cold}, puddle, warm, got, chicago, protect, suppose}}
\\
\hline
\multicolumn{1}{|c|}{\system: Weather resistance \color{blue} \Checkmark} &
\multicolumn{1}{|c|}{{MATE: General \color{red}\XSolidBrush}} &
\multicolumn{1}{c|}{{TS-W2V: General \color{red}\XSolidBrush}} 
\\
\hline 
\multicolumn{2}{|l}{\multirow{1}{*}{It was a washed {\color{red}purplish} but not a true {\color{red}purple}}} &
\multicolumn{1}{c|}{GT: Color} 
\\
\multicolumn{3}{|l|}{\multirow{1}{*}{Seed Words: {\color{red}color}, love, {\color{red}style}, unbelievably, gorgeous, blue, favorite, taupe, wonderful, red, exactly, design}}
\\
\hline
\multicolumn{1}{|c|}{\system: Color \color{blue} \Checkmark} &
\multicolumn{1}{|c|}{{MATE: Size \color{red}\XSolidBrush}} &
\multicolumn{1}{c|}{{TS-W2V: General \color{red}\XSolidBrush}} 
\\
\hline 
\multicolumn{2}{|l}{\multirow{1}{*}{I wanted a true {\color{red}purple} boot.}} &
\multicolumn{1}{c|}{GT: Color} 
\\
\multicolumn{3}{|l|}{\multirow{1}{*}{Seed Words: {\color{red}color}, love, style, unbelievably, gorgeous, blue, favorite, taupe, wonderful, red, exactly, design}}
\\
\hline
\multicolumn{1}{|c|}{\system: Color \color{blue} \Checkmark} &
\multicolumn{1}{|c|}{{MATE: General \color{red}\XSolidBrush}} &
\multicolumn{1}{c|}{{TS-W2V: Price \color{red}\XSolidBrush}} 
\\
\hline 
\multicolumn{2}{|l}{\multirow{1}{*}{That I can easily {\color{red}slip on} and not {\color{red}struggle} to put it on.}} &
\multicolumn{1}{c|}{GT: Comfort} 
\\
\multicolumn{3}{|l|}{\multirow{1}{*}{Seed Words: comfortable, {\color{red}fit}, foot, hurt, ankle, comfy, blister, {\color{red}comfort}, break, {\color{red}wear}, able, extreme, walking, sock}}
\\
\hline
\multicolumn{1}{|c|}{$\system$: Comfort \color{blue} \Checkmark} &
\multicolumn{1}{|c|}{{MATE: Comfort \color{blue} \Checkmark}} &
\multicolumn{1}{c|}{{TS-W2V: General \color{red} \XSolidBrush}} 
\\
\hline 
\multicolumn{2}{|l}{\multirow{1}{*}{They seem pretty {\color{red}solid} to me.}} &
\multicolumn{1}{c|}{GT: Durability} 
\\
\multicolumn{3}{|l|}{\multirow{1}{*}{Seed Words: rubber, quality, use, buckle, appear, heel, {\color{red}long}, time, {\color{red}breaking}, enjoy, provide}}
\\
\hline
\multicolumn{1}{|c|}{\system: Durability \color{blue} \Checkmark} &
\multicolumn{1}{|c|}{{MATE: Durability \color{blue} \Checkmark}} &
\multicolumn{1}{c|}{{TS-W2V: General \color{red}\XSolidBrush}} 
\\
\hline 
\multicolumn{2}{|l}{\multirow{1}{*}{And {\color{red}scratch up} your fingers getting your foot into the things though.}} &
\multicolumn{1}{c|}{GT: Comfort} 
\\
\multicolumn{3}{|l|}{\multirow{1}{*}{Seed Words: {\color{red}comfortable}, fit, foot, hurt, ankle, comfy, blister, comfort, break, wear, able, extreme, walking, sock}}
\\
\hline
\multicolumn{1}{|c|}{\system: Comfort \color{blue} \Checkmark} &
\multicolumn{1}{|c|}{{MATE: General \color{red}\XSolidBrush}} &
\multicolumn{1}{c|}{{TS-W2V: Weather Resistance \color{red}\XSolidBrush}} 
\\
\hline 
\multicolumn{2}{|l}{\multirow{1}{*}{It would {\color{red}stay tied} as well.}} &
\multicolumn{1}{c|}{GT: Durability} 
\\
\multicolumn{3}{|l|}{\multirow{1}{*}{Seed Words: rubber, quality, use, buckle, appear, heel, {\color{red}long}, time, breaking, enjoy, provide, {\color{red}protection}, clearly, cheaply}}
\\
\hline
\multicolumn{1}{|c|}{\system: Durability \color{blue} \Checkmark} &
\multicolumn{1}{|c|}{{MATE: Look \color{red}\XSolidBrush}} &
\multicolumn{1}{c|}{{TS-W2V: General \color{red}\XSolidBrush}} 
\\
\hline 
\multicolumn{2}{|l}{\multirow{1}{*}{You need to design these with a {\color{red}zipper}.}} &
\multicolumn{1}{c|}{GT: Materials} 
\\
\multicolumn{3}{|l|}{\multirow{1}{*}{Seed Words: leather, inside, fringe, {\color{red}material}, heel, cheaper, soft, sole, buckle, suede, true, rubber, advertised}}
\\
\hline
\multicolumn{1}{|c|}{\system: Materials \color{blue} \Checkmark} &
\multicolumn{1}{|c|}{{MATE: Materials \color{blue} \Checkmark}} &
\multicolumn{1}{c|}{{TS-W2V: General \color{red}\XSolidBrush}} 
\\
\hline 
\end{tabular}
}
\caption{Comparison of predictions on sample Boots product review segments between $\system$, MATE, and TS-W2V. For each review segment, the ground true (GT) aspect and its corresponding seed words are provided.}
\label{tb:case_st_kb}
\end{table*}

\begin{table*}[t]
\fontsize{9}{10}\selectfont
\center
\small
\scalebox{1.0}{
\noindent\begin{tabular}{|c|c|c|c|c|}
\hline 
\multicolumn{2}{|l}{\multirow{1}{*}{Plus, a lot of the more expensive cases aren't as bright as this one.}} &
\multicolumn{1}{c|}{GT: Looks} 
\\
\multicolumn{3}{|l|}{\multirow{1}{*}{Seed Words: look, color, pink, stylish, looked, pretty, lime, green, fashionable, picture, awesome, good, great}}
\\
\hline
\multicolumn{1}{|c|}{\system: General \color{red}\XSolidBrush} &
\multicolumn{1}{|c|}{{MATE: Price \color{red}\XSolidBrush}} &
\multicolumn{1}{c|}{{TS-W2V: General \color{red}\XSolidBrush}} 
\\
\hline 
\multicolumn{2}{|l}{\multirow{1}{*}{And we bought two of this sleeve.}} &
\multicolumn{1}{c|}{GT: General} 
\\
\multicolumn{3}{|l|}{\multirow{1}{*}{Seed Words: recommend, bought, used, hard, bag, work, using, pleased, say, backpack, weight, case, new, purchased}}
\\
\hline
\multicolumn{1}{|c|}{\system: General \color{blue} \Checkmark} &
\multicolumn{1}{|c|}{{MATE: Size fit \color{red}\XSolidBrush}} &
\multicolumn{1}{c|}{{TS-W2V: General \color{blue} \Checkmark}} 
\\
\hline 
\multicolumn{2}{|l}{\multirow{1}{*}{My only complaint is the interior color}} &
\multicolumn{1}{c|}{GT: Looks} 
\\
\multicolumn{3}{|l|}{\multirow{1}{*}{Seed Words: look, color, pink, stylish, looked, pretty, lime, green, fashionable, picture, awesome, good, great}}
\\
\hline
\multicolumn{1}{|c|}{\system: Looks \color{blue} \Checkmark} &
\multicolumn{1}{|c|}{{MATE: Looks \color{blue} \Checkmark}} &
\multicolumn{1}{c|}{{TS-W2V: Looks \color{blue} \Checkmark}} 
\\
\hline 
\multicolumn{2}{|l}{\multirow{1}{*}{About what size laptop fits in it.}} &
\multicolumn{1}{c|}{GT: Size fit} 
\\
\multicolumn{3}{|l|}{\multirow{1}{*}{Seed Words: fit, size, macbook, big, space, air, slightly, lot, bulk, perfect, inch, laptop, perfectly, 13, cable, folder, notebook}}
\\
\hline
\multicolumn{1}{|c|}{\system: Size fit \color{blue} \Checkmark} &
\multicolumn{1}{|c|}{{MATE: Looks \color{red}\XSolidBrush}} &
\multicolumn{1}{c|}{{TS-W2V: Size fit \color{blue} \Checkmark}} 
\\
\hline 
\multicolumn{2}{|l}{\multirow{1}{*}{I can barely sling it over my shoulder.}} &
\multicolumn{1}{c|}{GT: Handles} 
\\
\multicolumn{3}{|l|}{\multirow{1}{*}{Seed Words: strap, handle, shoulder, broke, later, month, hand, comfortable, plastic, wear, tear, ripped}}
\\
\hline
\multicolumn{1}{|c|}{\system: Handles \color{blue} \Checkmark} &
\multicolumn{1}{|c|}{{MATE: Handles \color{blue} \Checkmark}} &
\multicolumn{1}{c|}{{TS-W2V: Handles \color{blue} \Checkmark}} 
\\
\hline 
\multicolumn{2}{|l}{\multirow{1}{*}{It just looks kinda funny.}} &
\multicolumn{1}{c|}{GT: Looks} 
\\
\multicolumn{3}{|l|}{\multirow{1}{*}{Seed Words: look, color, pink, stylish, looked, pretty, lime, green, fashionable, picture, awesome, good, great, design, fre, cell, styling}}
\\
\hline
\multicolumn{1}{|c|}{\system: Looks \color{blue} \Checkmark} &
\multicolumn{1}{|c|}{{MATE: Looks \color{blue} \Checkmark}} &
\multicolumn{1}{c|}{{TS-W2V: Looks \color{blue} \Checkmark}} 
\\
\hline 
\multicolumn{2}{|l}{\multirow{1}{*}{Starting to {\color{red}fall apart}.}} &
\multicolumn{1}{c|}{GT: Quality} 
\\
\multicolumn{3}{|l|}{\multirow{1}{*}{Seed Words: quality, material, handle, {\color{red}poor}, {\color{red}broke}, durable, month, later, wear, tear, zipper, chemical, started, excellent, terrible}}
\\
\hline
\multicolumn{1}{|c|}{\system: Quality \color{blue} \Checkmark} &
\multicolumn{1}{|c|}{{MATE: Looks \color{red}\XSolidBrush}} &
\multicolumn{1}{c|}{{TS-W2V: General \color{red}\XSolidBrush}} 
\\
\hline 
\multicolumn{2}{|l}{\multirow{1}{*}{And part of it {\color{red}chipped off}}} &
\multicolumn{1}{c|}{GT: Quality} 
\\
\multicolumn{3}{|l|}{\multirow{1}{*}{Seed Words: quality, material, handle, {\color{red}poor}, {\color{red}broke}, durable, month, later, wear, tear, zipper, chemical, started, excellent, terrible}}
\\
\hline
\multicolumn{1}{|c|}{\system: Quality \color{blue} \Checkmark} &
\multicolumn{1}{|c|}{{MATE: Handles \color{red}\XSolidBrush}} &
\multicolumn{1}{c|}{{TS-W2V: General \color{red}\XSolidBrush}} 
\\
\hline 
\multicolumn{2}{|l}{\multirow{1}{*}{{Very fast {\color{red}shipping}.}}} &
\multicolumn{1}{c|}{GT: Customer service} 
\\
\multicolumn{3}{|l|}{\multirow{1}{*}{Seed Words: hassle, {\color{red}return}, unable, Christmas, difficult, free, replaced,  {\color{red}arrived}, promptly, gift, easy, getting, returned, week, amazon}}
\\
\hline
\multicolumn{1}{|c|}{\system: Customer service \color{blue} \Checkmark} &
\multicolumn{1}{|c|}{{MATE: General service \color{red}\XSolidBrush}} &
\multicolumn{1}{c|}{{TS-W2V: General \color{red}\XSolidBrush}}
\\
\hline 
\multicolumn{2}{|l}{\multirow{1}{*}{The case is so {\color{red}small}}} &
\multicolumn{1}{c|}{GT: Size fit} 
\\
\multicolumn{3}{|l|}{\multirow{1}{*}{Seed Words: fit, {\color{red}size}, macbook, big, {\color{red}space}, air, slightly, lot, bulk, perfect, inch, laptop, perfectly, 13, cable, folder}}
\\
\hline
\multicolumn{1}{|c|}{\system: Size fit \color{blue} \Checkmark} &
\multicolumn{1}{|c|}{{MATE: Protection \color{red}\XSolidBrush}} &
\multicolumn{1}{c|}{{TS-W2V: General \color{red}\XSolidBrush}}
\\
\hline 
\multicolumn{2}{|l}{\multirow{1}{*}{Which is a {\color{red}good deal}}} &
\multicolumn{1}{c|}{GT: Price} 
\\
\multicolumn{3}{|l|}{\multirow{1}{*}{Seed Words: price, {\color{red}worth}, pay, good, great, requested, believe, send, 100, suppose, buck, based, reasonable, inexpensive}}
\\
\hline
\multicolumn{1}{|c|}{\system: Price \color{blue} \Checkmark} &
\multicolumn{1}{|c|}{{MATE: Price \color{blue} \Checkmark}} &
\multicolumn{1}{c|}{{TS-W2V: Looks \color{red}\XSolidBrush}} 
\\
\hline 
\multicolumn{2}{|l}{\multirow{1}{*}{On the plus side it does have a lot of nooks and crannies to  {\color{red}put stuff in}.}} &
\multicolumn{1}{c|}{GT: Compartments} 
\\
\multicolumn{3}{|l|}{\multirow{1}{*}{Seed Words: pocket, cable,  {\color{red}compartment}, outside, lot, wish, wallet, connector, space, power, pen, folder, charger, flap, mouse, nice}}
\\
\hline
\multicolumn{1}{|c|}{\system: Compartments \color{blue} \Checkmark} &
\multicolumn{1}{|c|}{{MATE: Size Fit \color{red}\XSolidBrush}} &
\multicolumn{1}{c|}{{TS-W2V: Size Fit \color{red}\XSolidBrush}} 
\\
\hline 
\end{tabular}}
\caption{Comparison of predictions on sample Bags product review segments between \system, MATE, and TS-W2V. For each review segment, the ground true (GT) aspect and its corresponding seed words are provided.}
\vspace{-1.5 pc}
\label{tb:case_st_kb}
\end{table*}

\begin{table*}[t]
\fontsize{9}{10}\selectfont
\center
\small
\scalebox{1.0}{
\noindent\begin{tabular}{|c|c|c|c|c|}
\hline 
\multicolumn{2}{|l}{\multirow{1}{*}{If you don't need all the features.}} &
\multicolumn{1}{c|}{GT: Apps Interface} 
\\
\multicolumn{3}{|l|}{\multirow{1}{*}{Seed Words: netflix, user, file, hulu, apps, watch, flash, internet, smart, video}}
\\
\hline
\multicolumn{1}{|c|}{\system: Ease of use \color{red}\XSolidBrush} &
\multicolumn{1}{|c|}{{MATE: Size Look \color{red}\XSolidBrush}} &
\multicolumn{1}{c|}{{TS-W2V: General \color{red}\XSolidBrush}} 
\\
\hline 
\multicolumn{2}{|l}{\multirow{1}{*}{The Yahoo! widgets do not work}} &
\multicolumn{1}{c|}{GT: Apps Interface}
\\
\multicolumn{3}{|l|}{\multirow{1}{*}{Seed Words: netflix, user, file, hulu, apps, watch, flash, internet, smart, video}}
\\
\hline
\multicolumn{1}{|c|}{\system: Apps Interface \color{blue} \Checkmark} &
\multicolumn{1}{|c|}{{MATE: Customer Service \color{red}\XSolidBrush}} &
\multicolumn{1}{c|}{{TS-W2V: Apps Interface \color{blue} \Checkmark}} 
\\
\hline 
\multicolumn{2}{|l}{\multirow{1}{*}{The picture quality is very sharp and crisp}} &
\multicolumn{1}{c|}{GT: Image} 
\\
\multicolumn{3}{|l|}{\multirow{1}{*}{Seed Words: picture, color, quality, back, bright, nice, clear, look, excellent, crisp, screen, right, dead, pixel, trace, beautiful}}
\\
\hline
\multicolumn{1}{|c|}{\system: Image \color{blue} \Checkmark} &
\multicolumn{1}{|c|}{{MATE: Price \color{red}\XSolidBrush}} &
\multicolumn{1}{c|}{{TS-W2V: Image \color{blue} \Checkmark}} 
\\
\hline 
\multicolumn{2}{|l}{\multirow{1}{*}{The price is enticing.}} &
\multicolumn{1}{c|}{GT: Price} 
\\
\multicolumn{3}{|l|}{\multirow{1}{*}{Seed Words: price, value, money, 1500, worth, 300, paid, solid, used, pretty, great}}
\\
\hline
\multicolumn{1}{|c|}{\system: Price \color{blue} \Checkmark} &
\multicolumn{1}{|c|}{{MATE: Image \color{red}\XSolidBrush}} &
\multicolumn{1}{c|}{{TS-W2V: Price \color{blue} \Checkmark}} 
\\
\hline 
\multicolumn{2}{|l}{\multirow{1}{*}{But bright colors generally looked}} &
\multicolumn{1}{c|}{GT: Image} 
\\
\multicolumn{3}{|l|}{\multirow{1}{*}{Seed Words: picture, color, quality, back, bright, nice, clear, look, excellent, crisp, screen, right, dead, pixel, trace, beautiful}}
\\
\hline
\multicolumn{1}{|c|}{\system: Image \color{blue} \Checkmark} &
\multicolumn{1}{|c|}{{MATE: Image \color{blue} \Checkmark}} &
\multicolumn{1}{c|}{{TS-W2V: Image \color{blue} \Checkmark}} 
\\
\hline 
\multicolumn{2}{|l}{\multirow{1}{*}{The sound from the TV itself is very tiny.}} &
\multicolumn{1}{c|}{GT: Sound} 
\\
\multicolumn{3}{|l|}{\multirow{1}{*}{Seed Words: sound, speaker, good, quality, loud, tinny, bass, hear, horrible, treble, change}}
\\
\hline
\multicolumn{1}{|c|}{\system: Sound \color{blue} \Checkmark} &
\multicolumn{1}{|c|}{{MATE: General \color{red}\XSolidBrush}} &
\multicolumn{1}{c|}{{TS-W2V: Sound \color{blue} \Checkmark}} 
\\
\hline 
\multicolumn{2}{|l}{\multirow{1}{*}{The cable connection port}} &
\multicolumn{1}{c|}{GT: Connectivity} 
\\
\multicolumn{3}{|l|}{\multirow{1}{*}{Seed Words: hdmi, port, computer, input, component, usb, internet, connection, output, server, connect, audio, play, stopped}}
\\
\hline
\multicolumn{1}{|c|}{\system: Connectivity \color{blue} \Checkmark} &
\multicolumn{1}{|c|}{{MATE: Connectivity \color{blue} \Checkmark}} &
\multicolumn{1}{c|}{{TS-W2V: Connectivity \color{blue} \Checkmark}} 
\\
\hline 
\multicolumn{2}{|l}{\multirow{1}{*}{Picture and sound are both acceptable.}} &
\multicolumn{1}{c|}{GT: Sound} 
\\
\multicolumn{3}{|l|}{\multirow{1}{*}{Seed Words: sound, speaker, good, quality, loud, tinny, bass, hear, horrible, treble, change}}
\\
\hline
\multicolumn{1}{|c|}{\system: Sound \color{blue} \Checkmark} &
\multicolumn{1}{|c|}{{MATE: Sound \color{blue} \Checkmark}} &
\multicolumn{1}{c|}{{TS-W2V: Sound \color{blue} \Checkmark}} 
\\
\hline 
\multicolumn{2}{|l}{\multirow{1}{*}{Are difficult to use and setup}} &
\multicolumn{1}{c|}{GT: Ease of Use} 
\\
\multicolumn{3}{|l|}{\multirow{1}{*}{Seed Words: easy, channel, remote, setup, feature, user, menu, use, setting, using, turn, set, input, cause, shuts, cycle, signal}}
\\
\hline
\multicolumn{1}{|c|}{\system: Ease of Use \color{blue} \Checkmark} &
\multicolumn{1}{|c|}{{MATE: Ease of Use \color{blue} \Checkmark}} &
\multicolumn{1}{c|}{{TS-W2V: Ease of Use \color{blue} \Checkmark}} 
\\
\hline 
\multicolumn{2}{|l}{\multirow{1}{*}{Fast moving objects were incredibly {\color{red}pixelated} and blotchy.}} &
\multicolumn{1}{c|}{GT: Image} 
\\
\multicolumn{3}{|l|}{\multirow{1}{*}{Seed Words: {\color{red}picture}, color, quality, back, bright, nice, clear, look, excellent, crisp, screen, right, dead, {\color{red}pixel}, trace, beautiful}}
\\
\hline
\multicolumn{1}{|c|}{\system: Image \color{blue} \Checkmark} &
\multicolumn{1}{|c|}{{MATE: Image \color{blue} \Checkmark}} &
\multicolumn{1}{c|}{{TS-W2V: General \color{red}\XSolidBrush}} 
\\
\hline 
\multicolumn{2}{|l}{\multirow{1}{*}{Lacks a net {\color{red}browser} and the only thing.}} &
\multicolumn{1}{c|}{GT: Apps Interface} 
\\
\multicolumn{3}{|l|}{\multirow{1}{*}{Seed Words: netflix, user, file, hulu, apps, watch, flash, {\color{red}internet}, smart, video}}
\\
\hline
\multicolumn{1}{|c|}{\system: Apps Interface \color{blue} \Checkmark} &
\multicolumn{1}{|c|}{{MATE: Ease of Use \color{red}\XSolidBrush}} &
\multicolumn{1}{c|}{{TS-W2V: Sound \color{red}\XSolidBrush}} 
\\
\hline 
\multicolumn{2}{|l}{\multirow{1}{*}{Is {\color{red}pandora}.}} &
\multicolumn{1}{c|}{GT: Apps Interface} 
\\
\multicolumn{3}{|l|}{\multirow{1}{*}{Seed Words: netflix, user, file, hulu, {\color{red}apps}, watch, flash, internet, smart, video}}
\\
\hline
\multicolumn{1}{|c|}{\system: Apps Interface \color{blue} \Checkmark} &
\multicolumn{1}{|c|}{{MATE: General \color{red}\XSolidBrush}} &
\multicolumn{1}{c|}{{TS-W2V: General \color{red}\XSolidBrush}} 
\\
\hline 
\multicolumn{2}{|l}{\multirow{1}{*}{Washed out but {\color{red}720p} movies.}} &
\multicolumn{1}{c|}{GT: Image} 
\\
\multicolumn{3}{|l|}{\multirow{1}{*}{Seed Words: {\color{red}picture}, color, quality, back, bright, nice, clear, look, excellent, crisp, screen, right, dead, {\color{red}pixel}, trace, beautiful}}
\\
\hline
\multicolumn{1}{|c|}{\system: Image \color{blue} \Checkmark} &
\multicolumn{1}{|c|}{{MATE: General \color{red}\XSolidBrush}} &
\multicolumn{1}{c|}{{TS-W2V: Connectivity \color{red}\XSolidBrush}} 
\\
\hline 
\end{tabular}}
\caption{Comparison of predictions on sample Televisions product review segments between \system, MATE, and TS-W2V. For each review segment, the ground true (GT) aspect and its corresponding seed words are provided.}
\label{tb:case_st_kb}
\end{table*}

\begin{table*}[t]
\fontsize{9}{10}\selectfont
\center
\small
\scalebox{1.0}{
\noindent\begin{tabular}{|c|c|c|c|c|}
\hline 
\multicolumn{2}{|l}{\multirow{1}{*}{To roll the windows up.}} &
\multicolumn{1}{c|}{GT: Sound} 
\\
\multicolumn{3}{|l|}{\multirow{1}{*}{Seed Words: sound, quality, hear, noise, volume, end, audio, good, voice, conversation, people, told, clearly, background}}
\\
\hline
\multicolumn{1}{|c|}{\system: General \color{red}\XSolidBrush} &
\multicolumn{1}{|c|}{{MATE: General \color{red}\XSolidBrush}} &
\multicolumn{1}{c|}{{TS-W2V: General \color{red}\XSolidBrush}} 
\\
\hline 
\multicolumn{2}{|l}{\multirow{1}{*}{The volume is GREAT!}} &
\multicolumn{1}{c|}{GT: Sound} 
\\
\multicolumn{3}{|l|}{\multirow{1}{*}{Seed Words: sound, quality, hear, noise, volume, end, audio, good, voice, conversation, people, told, clearly, background}}
\\
\hline
\multicolumn{1}{|c|}{\system: Sound \color{blue} \Checkmark} &
\multicolumn{1}{|c|}{{MATE: Price \color{red}\XSolidBrush}} &
\multicolumn{1}{c|}{{TS-W2V: Look \color{red}\XSolidBrush}} 
\\
\hline 
\multicolumn{2}{|l}{\multirow{1}{*}{Very easy to put on.}} &
\multicolumn{1}{c|}{GT: Ease Of Use} 
\\
\multicolumn{3}{|l|}{\multirow{1}{*}{Seed Words: easy, control, button, simple, setup, mic, phone, colored, flashing, confusing, make, convenient}}
\\
\hline
\multicolumn{1}{|c|}{\system: Ease of use \color{blue} \Checkmark} &
\multicolumn{1}{|c|}{{MATE: Durability \color{red}\XSolidBrush}} &
\multicolumn{1}{c|}{{TS-W2V: Ease Of Use \color{blue} \Checkmark}} 
\\
\hline 
\multicolumn{2}{|l}{\multirow{1}{*}{I have had mine for about a week}} &
\multicolumn{1}{c|}{GT: Durability} 
\\
\multicolumn{3}{|l|}{\multirow{1}{*}{Seed Words: week, long, month, durable, sweat, charging, turn, second, expected, holding, standard, comment}}
\\
\hline
\multicolumn{1}{|c|}{\system: Durability \color{blue} \Checkmark} &
\multicolumn{1}{|c|}{{MATE: General \color{red}\XSolidBrush}} &
\multicolumn{1}{c|}{{TS-W2V: Durability \color{blue} \Checkmark}} 
\\
\hline 
\multicolumn{2}{|l}{\multirow{1}{*}{They hurts your ears.}} &
\multicolumn{1}{c|}{GT: Comfort} 
\\
\multicolumn{3}{|l|}{\multirow{1}{*}{Seed Words: ear, fit, comfortable, bud, feel, jaybird, comfortably, lightweight, wear, readjusting, hard}}
\\
\hline
\multicolumn{1}{|c|}{\system: Comfort \color{blue} \Checkmark} &
\multicolumn{1}{|c|}{{MATE: Comfort \color{blue} \Checkmark}} &
\multicolumn{1}{c|}{{TS-W2V: Comfort \color{blue} \Checkmark}} 
\\
\hline 
\multicolumn{2}{|l}{\multirow{1}{*}{Attached to a non bluetooh phone, a Razr, and a Blackberry with excelent results in every case}} &
\multicolumn{1}{c|}{GT: Connectivity} 
\\
\multicolumn{3}{|l|}{\multirow{1}{*}{Seed Words: phone, range, easy, easily, galaxy, foot, connection, optimal, zone, cut, paired, link, connects}}
\\
\hline
\multicolumn{1}{|c|}{\system: Connectivity \color{blue} \Checkmark} &
\multicolumn{1}{|c|}{{MATE: Durability \color{red}\XSolidBrush}} &
\multicolumn{1}{c|}{{TS-W2V: Connectivity \color{blue} \Checkmark}} 
\\
\hline 
\multicolumn{2}{|l}{\multirow{1}{*}{As for the ear pad.}} &
\multicolumn{1}{c|}{GT: Comfort} 
\\
\multicolumn{3}{|l|}{\multirow{1}{*}{Seed Words: ear, fit, comfortable, bud, feel, jaybird, comfortably, lightweight, wear, readjusting, hard}}
\\
\hline
\multicolumn{1}{|c|}{\system: Comfort \color{blue} \Checkmark} &
\multicolumn{1}{|c|}{{MATE: Comfort \color{blue} \Checkmark}} &
\multicolumn{1}{c|}{{TS-W2V: Comfort \color{blue} \Checkmark}} 
\\
\hline 
\multicolumn{2}{|l}{\multirow{1}{*}{Strengths : Sweat and water resistant, lighter.}} &
\multicolumn{1}{c|}{GT: Durability} 
\\
\multicolumn{3}{|l|}{\multirow{1}{*}{Seed Words: week, long, month, durable, sweat, charging, turn, second, expected, holding, standard, comment}}
\\
\hline
\multicolumn{1}{|c|}{\system: Durability \color{blue} \Checkmark} &
\multicolumn{1}{|c|}{{MATE: Durability \color{blue} \Checkmark}} &
\multicolumn{1}{c|}{{TS-W2V: Durability \color{blue} \Checkmark}} 
\\
\hline 
\multicolumn{2}{|l}{\multirow{1}{*}{It is portable and sweat proof.}} &
\multicolumn{1}{c|}{GT: Durability} 
\\
\multicolumn{3}{|l|}{\multirow{1}{*}{Seed Words: week, long, month, durable, sweat, charging, turn, second, expected, holding, standard, comment}}
\\
\hline
\multicolumn{1}{|c|}{\system: Durability \color{blue} \Checkmark} &
\multicolumn{1}{|c|}{{MATE: Durability \color{blue} \Checkmark}} &
\multicolumn{1}{c|}{{TS-W2V: Durability \color{blue} \Checkmark}} 
\\
\hline 
\multicolumn{2}{|l}{\multirow{1}{*}{I do like the {\color{red}design} of it and the fact.}} &
\multicolumn{1}{c|}{GT: Look} 
\\
\multicolumn{3}{|l|}{\multirow{1}{*}{Seed Words: {\color{red}look}, {\color{red}appearance}, hs850, nice, earpiece, h700, smaller, feel, fine, bluetooth, great, good, headset}}
\\
\hline
\multicolumn{1}{|c|}{\system: Look \color{blue} \Checkmark} &
\multicolumn{1}{|c|}{{MATE: Durability \color{red}\XSolidBrush}} &
\multicolumn{1}{c|}{{TS-W2V: General \color{red}\XSolidBrush}} 
\\
\hline 
\multicolumn{2}{|l}{\multirow{1}{*}{To {\color{red}roll} the windows up}} &
\multicolumn{1}{c|}{GT: Sound} 
\\
\multicolumn{3}{|l|}{\multirow{1}{*}{Seed Words: {\color{red}sound}, quality, hear, noise, volume, end, audio, good, voice, conversation, people, told, clearly, background}}
\\
\hline
\multicolumn{1}{|c|}{\system: Sound \color{blue} \Checkmark} &
\multicolumn{1}{|c|}{{MATE: Comfort \color{red}\XSolidBrush}} &
\multicolumn{1}{c|}{{TS-W2V: General \color{red}\XSolidBrush}} 
\\
\hline 
\multicolumn{2}{|l}{\multirow{1}{*}{The {\color{red}style} is very {\color{red}cool} and the unit feels top quality.}} &
\multicolumn{1}{c|}{GT: Look} 
\\
\multicolumn{3}{|l|}{\multirow{1}{*}{Seed Words: {\color{red}look}, {\color{red}appearance}, hs850, nice, earpiece, h700, smaller, feel, fine, bluetooth, great, {\color{red}good}, headset}}
\\
\hline
\multicolumn{1}{|c|}{\system: Look \color{blue} \Checkmark} &
\multicolumn{1}{|c|}{{MATE: General \color{red}\XSolidBrush}} &
\multicolumn{1}{c|}{{TS-W2V: Look \color{blue} \Checkmark}} 
\\
\hline 
\multicolumn{2}{|l}{\multirow{1}{*}{On the plus side, the {\color{red}call} quality is good.}} &
\multicolumn{1}{c|}{GT: Sound} 
\\
\multicolumn{3}{|l|}{\multirow{1}{*}{Seed Words: {\color{red}sound}, quality, hear, noise, volume, end, audio, good, {\color{red}voice}, conversation, people, told, clearly, background}}
\\
\hline
\multicolumn{1}{|c|}{\system: Sound \color{blue} \Checkmark} &
\multicolumn{1}{|c|}{{MATE: Price \color{red}\XSolidBrush}} &
\multicolumn{1}{c|}{{TS-W2V: Look \color{red}\XSolidBrush}} 
\\
\hline 
\multicolumn{2}{|l}{\multirow{1}{*}{It starting to {\color{red}hurt} me.}} &
\multicolumn{1}{c|}{GT: Comfort} 
\\
\multicolumn{3}{|l|}{\multirow{1}{*}{Seed Words: ear, fit, comfortable, bud, {\color{red}feel}, jaybird, comfortably, lightweight, wear, readjusting, hard}}
\\
\hline
\multicolumn{1}{|c|}{\system: Comfort \color{blue} \Checkmark} &
\multicolumn{1}{|c|}{{MATE: Comfort \color{blue} \Checkmark}} &
\multicolumn{1}{c|}{{TS-W2V: General \color{red}\XSolidBrush}} 
\\
\hline 
\end{tabular}}
\caption{Comparison of predictions on sample Bluetooth Headsets product review segments between \system, MATE, and TS-W2V. For each review segment, the ground true (GT) aspect and its corresponding seed words are provided.}
\vspace{-1.5 pc}
\label{tb:case_st_kb}
\end{table*}



\begin{table*}[t]
\fontsize{9}{10}\selectfont
\center
\small
\scalebox{1.0}{
\noindent\begin{tabular}{|c|c|c|c|c|}

\hline 
\multicolumn{2}{|l}{\multirow{1}{*}{Eventually I started listening to my iPod.}} &
\multicolumn{1}{c|}{GT: General} 
\\
\multicolumn{3}{|l|}{\multirow{1}{*}{Seed Words: vac, cleaner, vacuum, buy, bought, new, better, year, recommend, product, owned, review, gave, away, kenmore, dyson}}
\\
\hline
\multicolumn{1}{|c|}{\system: Build Quality \color{red}\XSolidBrush} &
\multicolumn{1}{|c|}{{MATE: Build Quality \color{red}\XSolidBrush}} &
\multicolumn{1}{c|}{{TS-W2V: General \color{blue} \Checkmark}} 
\\
\hline 
\multicolumn{2}{|l}{\multirow{1}{*}{It is easy to move because of the adjustable wheels on the side of the brush}} &
\multicolumn{1}{c|}{GT: Ease of use} 
\\
\multicolumn{3}{|l|}{\multirow{1}{*}{Seed Words: easy, cord, push, corner, vacuuming, pile, maneuver, nozzle, awkward, crevice, constantly, bog, impossible, short}}
\\
\hline
\multicolumn{1}{|c|}{\system: Ease of use \color{blue} \Checkmark} &
\multicolumn{1}{|c|}{{MATE: Accessories \color{red}\XSolidBrush}} &
\multicolumn{1}{c|}{{TS-W2V:  Ease of use \color{blue} \Checkmark}} 
\\
\hline 
\multicolumn{2}{|l}{\multirow{1}{*}{Too bulky, cord is unusually stiff and tangles are impossible to remove.}} &
\multicolumn{1}{c|}{GT: Ease of use} 
\\
\multicolumn{3}{|l|}{\multirow{1}{*}{Seed Words: easy, cord, push, corner, vacuuming, pile, maneuver, nozzle, awkward, crevice, constantly, bog, impossible, short}}
\\
\hline
\multicolumn{1}{|c|}{\system: Ease of use \color{blue} \Checkmark} &
\multicolumn{1}{|c|}{{MATE: Build Quality \color{red}\XSolidBrush}} &
\multicolumn{1}{c|}{{TS-W2V: Ease of use \color{blue} \Checkmark}} 
\\
\hline 
\multicolumn{2}{|l}{\multirow{1}{*}{It is so easy to maneuver.}} &
\multicolumn{1}{c|}{GT: Ease of Use} 
\\
\multicolumn{3}{|l|}{\multirow{1}{*}{Seed Words: easy, cord, push, corner, vacuuming, pile, maneuver, nozzle, awkward, crevice, constantly, bog, impossible, short}}
\\
\hline
\multicolumn{1}{|c|}{\system: Ease of use \color{blue} \Checkmark} &
\multicolumn{1}{|c|}{{MATE: Weight \color{red}\XSolidBrush}} &
\multicolumn{1}{c|}{{TS-W2V: Ease of Use \color{blue} \Checkmark}} 
\\
\hline 
\multicolumn{2}{|l}{\multirow{1}{*}{Because it was so loud.}} &
\multicolumn{1}{c|}{GT: Noise}
\\
\multicolumn{3}{|l|}{\multirow{1}{*}{Seed Words: quiet, noisy, loud, powerful, noise, louder, ear, loudest, light, incredibly, deafening, seriously, actually}}
\\
\hline
\multicolumn{1}{|c|}{\system: Noise \color{blue} \Checkmark} &
\multicolumn{1}{|c|}{{MATE: Noise \color{blue} \Checkmark}} &
\multicolumn{1}{c|}{{TS-W2V: Noise \color{blue} \Checkmark}} 
\\
\hline 
\multicolumn{2}{|l}{\multirow{1}{*}{And have powerful suction.}} &
\multicolumn{1}{c|}{GT: Suction Power} 
\\
\multicolumn{3}{|l|}{\multirow{1}{*}{Seed Words: suction, pick, powerful, power, good, hair, carpet, such, quiet, really, performs, dirt, tile, ok}}
\\
\hline
\multicolumn{1}{|c|}{\system: Suction Power \color{blue} \Checkmark} &
\multicolumn{1}{|c|}{{MATE: Suction Power \color{blue} \Checkmark}} &
\multicolumn{1}{c|}{{TS-W2V: Suction Power \color{blue} \Checkmark}} 
\\
\hline 
\multicolumn{2}{|l}{\multirow{1}{*}{While the suction is very good.}} &
\multicolumn{1}{c|}{GT: Suction Power} 
\\
\multicolumn{3}{|l|}{\multirow{1}{*}{Seed Words: suction, pick, powerful, power, good, hair, carpet, such, quiet, really, performs, dirt, tile, ok}}
\\
\hline
\multicolumn{1}{|c|}{\system: Suction Power \color{blue} \Checkmark} &
\multicolumn{1}{|c|}{{MATE: Noise \color{red}\XSolidBrush}} &
\multicolumn{1}{c|}{{TS-W2V: Suction Power \color{blue} \Checkmark}} 
\\
\hline 
\multicolumn{2}{|l}{\multirow{1}{*}{Which {\color{red}prevents} {\color{red}wearing} and {\color{red}tearing} the plastic parts on the brush}} &
\multicolumn{1}{c|}{GT: Build Quality} 
\\
\multicolumn{3}{|l|}{\multirow{1}{*}{Seed Words: belt, {\color{red}broke}, turn, working, burning, electrical, built, {\color{red}stop}, month, roller, time, minute}}
\\
\hline
\multicolumn{1}{|c|}{\system: Build Quality \color{blue} \Checkmark} &
\multicolumn{1}{|c|}{{MATE: Accessories \color{red}\XSolidBrush}} &
\multicolumn{1}{c|}{{TS-W2V: Build Quality \color{blue} \Checkmark}} 
\\
\hline 
\multicolumn{2}{|l}{\multirow{1}{*}{The sides {\color{red}wrapped} with {\color{red}protective} rubber like rim}} &
\multicolumn{1}{c|}{GT: Build Quality} 
\\
\multicolumn{3}{|l|}{\multirow{1}{*}{Seed Words: belt, broke, turn, working, burning, electrical, {\color{red}built}, {\color{red}stop}, month, roller, time, minute, {\color{red}problem}}}
\\
\hline
\multicolumn{1}{|c|}{\system: Build Quality \color{blue} \Checkmark} &
\multicolumn{1}{|c|}{{MATE: Noise \color{red}\XSolidBrush}} &
\multicolumn{1}{c|}{{TS-W2V: General \color{red}\XSolidBrush}} 
\\
\hline 
\multicolumn{2}{|l}{\multirow{1}{*}{Then the engine completely {\color{red}stopped} vacuuming.}} &
\multicolumn{1}{c|}{GT: Build Quality} 
\\
\multicolumn{3}{|l|}{\multirow{1}{*}{Seed Words: belt, broke, turn, {\color{red}working}, burning, electrical, built, stop, month, roller, time, minute, {\color{red}problem}, brush, design}}
\\
\hline
\multicolumn{1}{|c|}{\system: Build Quality \color{blue} \Checkmark} &
\multicolumn{1}{|c|}{{MATE: Suction Power \color{red}\XSolidBrush}} &
\multicolumn{1}{c|}{{TS-W2V: Ease of Use \color{red}\XSolidBrush}} 
\\
\hline 
\multicolumn{2}{|l}{\multirow{1}{*}{A {\color{red}small}, light-weight appliance that can do a big job.}} &
\multicolumn{1}{c|}{GT: Weight} 
\\
\multicolumn{3}{|l|}{\multirow{1}{*}{Seed Words: light, weight, lightweight, heavy, {\color{red}size}, compact, maneuver, guess, quiet, quite, probably}}
\\
\hline
\multicolumn{1}{|c|}{\system: Weight \color{blue} \Checkmark} &
\multicolumn{1}{|c|}{{MATE: Price \color{red}\XSolidBrush}} &
\multicolumn{1}{c|}{{TS-W2V: Suction Power \color{red}\XSolidBrush}} 
\\
\hline 
\multicolumn{2}{|l}{\multirow{1}{*}{You have to hold it at a very {\color{red}uncomfortable} angle}} &
\multicolumn{1}{c|}{GT: Ease of Use} 
\\
\multicolumn{3}{|l|}{\multirow{1}{*}{Seed Words: {\color{red}easy}, cord, push, corner, vacuuming, pile, {\color{red}maneuver}, nozzle, awkward, crevice, constantly, bog, impossible, short}}
\\
\hline
\multicolumn{1}{|c|}{\system: Ease of Use \color{blue} \Checkmark} &
\multicolumn{1}{|c|}{{MATE: Ease of Use \color{blue} \Checkmark}} &
\multicolumn{1}{c|}{{TS-W2V: General \color{red}\XSolidBrush}} 
\\
\hline 
\multicolumn{2}{|l}{\multirow{1}{*}{The {\color{red}tools} were previously stored inside the canister}} &
\multicolumn{1}{c|}{GT: Accessories} 
\\
\multicolumn{3}{|l|}{\multirow{1}{*}{Seed Words: {\color{red}filter}, {\color{red}brush}, {\color{red}attachment}, roll, turbo, easily, expensive, wide, turn, bag, replacing, typical, hepa}}
\\
\hline
\multicolumn{1}{|c|}{\system: Accessories \color{blue} \Checkmark} &
\multicolumn{1}{|c|}{{MATE: Build Quality \color{red}\XSolidBrush}} &
\multicolumn{1}{c|}{{TS-W2V: General \color{red}\XSolidBrush}} 
\\
\hline 
\end{tabular}}
\caption{Comparison of predictions on sample Vacuums product review segments between \system, MATE, and TS-W2V. For each review segment, the ground true (GT) aspect and its corresponding seed words are provided.}
\vspace{-1.5 pc}
\label{tb:case_st_kb}
\end{table*}

\end{document}